# Fusions of Description Logics and Abstract Description Systems

**Franz Baader**                                        BAADER@CS.RWTH-AACHEN.DE
**Carsten Lutz**                                        LUTZ@CS.RWTH-AACHEN.DE
*Teaching and Research Area for Theoretical Computer Science,*
*RWTH Aachen, Ahornstraße 55, 52074 Aachen, Germany*

**Holger Sturm**                                        HOLGER.STURM@UNI-KONSTANZ.DE
*Fachbereich Philosophie, Universität Konstanz,*
*78457 Konstanz, Germany*

**Frank Wolter**                                        WOLTER@INFORMATIK.UNI-LEIPZIG.DE
*Institut für Informatik, Universität Leipzig,*
*Augustus-Platz 10-11, 04109 Leipzig, Germany*

## Abstract

Fusions are a simple way of combining logics. For normal modal logics, fusions have been investigated in detail. In particular, it is known that, under certain conditions, decidability transfers from the component logics to their fusion. Though description logics are closely related to modal logics, they are not necessarily normal. In addition, ABox reasoning in description logics is not covered by the results from modal logics.

In this paper, we extend the decidability transfer results from normal modal logics to a large class of description logics. To cover different description logics in a uniform way, we introduce abstract description systems, which can be seen as a common generalization of description and modal logics, and show the transfer results in this general setting.

## 1. Introduction

Knowledge representation systems based on description logics (DL) can be used to represent the knowledge of an application domain in a structured and formally well-understood way (Brachman & Schmolze, 1985; Baader & Hollunder, 1991; Brachman, McGuinness, Patel-Schneider, Alperin Resnick, & Borgida, 1991; Woods & Schmolze, 1992; Borgida, 1995; Horrocks, 1998). In such systems, the important notions of the domain can be described by *concept descriptions*, i.e., expressions that are built from atomic concepts (unary predicates) and atomic roles (binary predicates) using the concept constructors provided by the description logic employed by the system. The atomic concepts and the concept descriptions represent sets of individuals, whereas roles represent binary relations between individuals. For example, using the atomic concepts Woman and Human, and the atomic role child, the concept of all *women having only daughters* (i.e., women such that all their children are again women) can be represented by the description Woman $\sqcap$ $\forall$child.Woman, and the concept of all *mothers* by the description Woman $\sqcap$ $\exists$child.Human. In this example, we have used the constructors concept conjunction ($\sqcap$), value restriction ($\forall R.C$), and existential restriction ($\exists R.C$). In the DL literature, also various other constructors have been considered. A prominent example are so-called number restrictions, which are available in almost all DL systems. For example, using number restrictions the concept of all *women* *having exactly two children* can be represented by the concept description

$$\mathsf{Woman} \sqcap (\leq 2\mathsf{child}) \sqcap (\geq 2\mathsf{child}).$$

The knowledge base of a DL system consists of a *terminological component* (TBox) and an *assertional component* (ABox). In its simplest form, the TBox consists of concept definitions, which assign names (abbreviations) to complex descriptions. More general TBox formalisms allow for so-called *general concept inclusion* axioms (GCIs) between complex descriptions. For example, the concept inclusion

$$\mathsf{Human} \sqcap (\geq 3\mathsf{child}) \sqsubseteq \exists\mathsf{entitled.Taxbreak}$$

states that people having at least three children are entitled to a tax break. The ABox formalism consists of concept assertions (stating that an individual belongs to a concept) and role assertions (stating that two individuals are related by a role). For example, the assertions $\mathsf{Woman}(\mathsf{MARY}), \mathsf{child}(\mathsf{MARY}, \mathsf{TOM}), \mathsf{Human}(\mathsf{TOM})$ state that Mary is a woman, who has a child, Tom, who is a human.

DL systems provide their users with various inference capabilities that allow them to deduce implicit knowledge from the explicitly represented knowledge. For instance, the *subsumption* problem is concerned with subconcept-superconcept relationships: $C$ is subsumed by $D$ ($C \sqsubseteq D$) if, and only if, all instances of $C$ are also instances of $D$, i.e., the first description is always interpreted as a subset of the second description. For example, the concept description $\mathsf{Woman}$ obviously subsumes the concept description $\mathsf{Woman} \sqcap \forall\mathsf{child.Woman}$. The concept description $C$ is *satisfiable* iff it is non-contradictory, i.e., it can be interpreted by a nonempty set. In DLs allowing for conjunction and negation of concepts, subsumption can be reduced to (un)satisfiability: $C \sqsubseteq D$ iff $C \sqcap \neg D$ is unsatisfiable. The *instance* checking problem consists of deciding whether a given individual is an instance of a given concept. For example, w.r.t. the assertions from above, $\mathsf{MARY}$ is an instance of the concept description $\mathsf{Woman} \sqcap \exists\mathsf{child.Human}$. The ABox $\mathcal{A}$ is *consistent* iff it is non-contradictory, i.e., it has a model. In DLs allowing for negation of concepts, the instance problem can be reduced to (in)consistency of ABoxes: $i$ is an instance of $C$ w.r.t. the ABox $\mathcal{A}$ iff $\mathcal{A} \cup \{\neg C(i)\}$ is inconsistent.

In order to ensure a reasonable and predictable behavior of a DL system, reasoning in the DL employed by the system should at least be decidable, and preferably of low complexity. Consequently, the expressive power of the DL in question must be restricted in an appropriate way. If the imposed restrictions are too severe, however, then the important notions of the application domain can no longer be expressed. Investigating this trade-off between the expressivity of DLs and the complexity of their inference problems has thus been one of the most important issues in DL research (see, e.g., Levesque & Brachman, 1987; Nebel, 1988; Schmidt-Schauß, 1989; Schmidt-Schauß & Smolka, 1991; Nebel, 1990; Donini, Lenzerini, Nardi, & Nutt, 1991, 1997; Donini, Hollunder, Lenzerini, Spaccamela, Nardi, & Nutt, 1992; Schaerf, 1993; Donini, Lenzerini, Nardi, & Schaerf, 1994; De Giacomo & Lenzerini, 1994a, 1994b, 1995; Calvanese, De Giacomo, & Lenzerini, 1999; Lutz, 1999; Horrocks, Sattler, & Tobies, 2000).

This paper investigates an approach for extending the expressivity of DLs that (in many cases) guarantees that reasoning remains decidable: the fusion of DLs. In order to explain the difference between the usual union and the fusion of DLs, let us consider a simple example. Assume that the DL $\mathcal{D}_1$ is $\mathcal{ALC}$, i.e., it provides for the Boolean operators $\sqcap$, $\sqcup$, $\neg$ and the additional concept constructors value restriction $\forall R.C$ and existential restriction $\exists R.C$, and that the DL $\mathcal{D}_2$ provides for the Boolean operators and number restrictions $(\leq nR)$ and $(\geq nR)$. If an application requires concept constructors from both DLs for expressing its relevant concepts, then one would usually consider the *union* $\mathcal{D}_1 \cup \mathcal{D}_2$ of $\mathcal{D}_1$ and $\mathcal{D}_2$, which allows for the unrestricted use of all constructors. For example, the concept description $C_1 := (\exists R.A) \sqcap (\exists R.\neg A) \sqcap (\leq 1R)$ is a legal $\mathcal{D}_1 \cup \mathcal{D}_2$ description. Note that this description is unsatisfiable, due to the interaction between constructors of $\mathcal{D}_1$ and $\mathcal{D}_2$. The *fusion* $\mathcal{D}_1 \otimes \mathcal{D}_2$ of $\mathcal{D}_1$ and $\mathcal{D}_2$ prevents such interactions by imposing the following restriction: one assumes that the set of all role names is partitioned into two sets, one that can be used in constructors of $\mathcal{D}_1$, and another one that can be used in constructors of $\mathcal{D}_2$. Thus, the description $C_1$ from above is not a legal $\mathcal{D}_1 \otimes \mathcal{D}_2$ description since it uses the same role $R$ both in the existential restriction (which are $\mathcal{D}_1$-constructors) and in the number restriction (which is a $\mathcal{D}_2$-constructor). In contrast, the descriptions $(\exists R_1.A) \sqcap (\exists R_1.\neg A) \sqcap (\leq 1R_2)$ and $(\exists R_1.(\leq 1R_2))$ are admissible in $\mathcal{D}_1 \otimes \mathcal{D}_2$ since they employ different roles in the $\mathcal{D}_1$- and $\mathcal{D}_2$-constructors. If the concepts that must be expressed are such that they require both constructors from $\mathcal{D}_1$ and $\mathcal{D}_2$, but the ones from $\mathcal{D}_1$ for other roles than the ones from $\mathcal{D}_2$, then one does not really need the union of $\mathcal{D}_1$ and $\mathcal{D}_2$; the fusion would be sufficient.

What is the advantage of taking the fusion instead of the union? Basically, for the union of two DLs one must design new reasoning methods, whereas reasoning in the fusion can be reduced to reasoning in the component DLs. Indeed, reasoning in the union may even be undecidable whereas reasoning in the fusion is still decidable. As an example, we consider the DLs (i) $\mathcal{ALCF}$, which extends the basic DL $\mathcal{ALC}$ by functional roles (features) and the same-as constructor (agreement) on chains of functional roles (Hollunder & Nutt, 1990; Baader, Bürckert, Nebel, Nutt, & Smolka, 1993); and (ii) $\mathcal{ALC}^{+,\circ,\sqcup}$, which extends $\mathcal{ALC}$ by transitive closure, composition, and union of roles (Baader, 1991; Schild, 1991). For both DLs, subsumption of concept descriptions is known to be decidable (Hollunder & Nutt, 1990; Schild, 1991; Baader, 1991). However, their *union* $\mathcal{ALCF}^{+,\circ,\sqcup}$ has an undecidable subsumption problem (Baader et al., 1993). This undecidability result depends on the fact that, in $\mathcal{ALCF}^{+,\circ,\sqcup}$, the role constructors transitive closure, composition, and union can be applied to functional roles that also appear within the same-as constructor. This is not allowed in the fusion $\mathcal{ALCF} \otimes \mathcal{ALC}^{+,\circ,\sqcup}$. Of course, failure of a certain undecidability proof does not make the fusion decidable.

Why do we know that the fusion of decidable DLs is again decidable? Actually, in general we don't, and this was our main reason for writing this paper. The notion "fusion" was introduced and investigated in modal logic, basically to transfer results like finite axiomatizability, decidability, finite model property, etc. from uni-modal logics (with one pair of box and diamond operators) to multi-modal logics (with several such pairs, possibly satisfying different axioms). This has led to rather general transfer results (see, e.g., Wolter, 1998; Kracht & Wolter, 1991; Fine & Schurz, 1996; Spaan, 1993; Gabbay, 1999 for results that concern decidability), which are sometimes restricted to so-called *normal* modal logics (Chellas, 1980). Since there is a close relationship between modal logics and DLs (Schild, 1991), it is clear that these transfer results also apply to some DLs. The question is, however, to which DLs exactly and to which inference problems. First, some DLs allow for constructors that are not considered in modal logics (e.g., the same-as constructor mentioned above). Second, some DL constructors that have been considered in modal logics, such as qualified number restrictions ($\leq nR.C$), ($\geq nR.C$) (Hollunder & Baader, 1991), which correspond to graded modalities (Van der Hoek & de Rijke, 1995), can easily be shown to be non-normal. Third, the transfer results for decidability are concerned with the satisfiability problem (with or without general inclusion axioms). ABoxes and the related inference problems are not considered. ABoxes can be simulated in modal logics allowing for so-called nominals, i.e., names for individuals, within formulae (Prior, 1967; Gargov & Goranko, 1993; Areces, Blackburn, & Marx, 2000). However, as we will see below, the general transfer results do not apply to modal logics with nominals.

The purpose of this paper is to clarify for which DLs decidability of the component DLs transfers to their fusion. To this purpose, we introduce so-called *abstract description systems* (ADSs), which can be seen as a common generalization of description and modal logics. We define the fusion of ADSs, and state four theorems that say under which conditions decidability transfers from the component ADSs to their fusion. Two of these theorems are concerned with inference w.r.t. general concept inclusion axioms and two with inference without TBox axioms. In both cases, we first formulate and prove the results for the consistency problem of ABoxes (more precisely, the corresponding problem for ADSs) and then establish analogous results for the satisfiability problem of concepts.

From the DL point of view, the four theorems shown in this paper are concerned with the following four *decision problems*:

(i) decidability of consistency of ABoxes w.r.t. TBox axioms (Theorem 17);

(ii) decidability of satisfiability of concepts w.r.t. TBox axioms; (Corollary 22);

(iii) decidability of consistency of ABoxes without TBox axioms (Theorem 29); and

(iv) decidability of satisfiability of concepts without TBox axioms (Corollary 34).

These theorems imply that decidability of the consistency problem and the satisfiability problem transfers to the fusion for most DLs considered in the literature. The *main exceptions* (which do not satisfy the prerequisites of the theorems) are

(a) DLs that are not propositionally closed, i.e, do not contain all Boolean connectives;

(b) DLs allowing for individuals (called nominals in modal logic) in concept descriptions; and

(c) DLs explicitly allowing for the universal role or for negation of roles.

Results from modal logic for problem (iv) usually require the component modal logics to be normal. Our Theorem 29 is less restrictive, and thus also applies to DLs allowing for constructors like qualified number restrictions.

## 2. Description logics

Before defining abstract description systems in the next section, we introduce the main features of DLs that must be covered by this definition. To this purpose, we first introduce $\mathcal{ALC}$, the basic DL containing all Boolean connectives, and the relevant inference problems. Then, we consider different possibilities for extending $\mathcal{ALC}$ to more expressive DLs.

**Definition 1 ($\mathcal{ALC}$ Syntax).** Let $N_C$, $N_R$, and $N_I$ be countable and pairwise disjoint sets of concept, role, and individual names, respectively. The set of $\mathcal{ALC}$ concept descriptions is the smallest set such that

1. every concept name is a concept description,

2. if $C$ and $D$ are concept descriptions and $R$ is a role name, then the following expressions are also concept descriptions:

   - $\neg C$ (negation), $C \sqcap D$ (conjunction), $C \sqcup D$ (disjunction),
   - $\exists R.C$ (existential restriction), and $\forall R.C$ (value restriction).

We use $\top$ as an abbreviation of $A \sqcup \neg A$ and $\bot$ as an abbreviation for $A \sqcap \neg A$ (where $A$ is an arbitrary concept name).

Let $C$ and $D$ be concept descriptions. Then $C \sqsubseteq D$ is a *general concept inclusion axiom* (GCI). A finite set of such axioms is called a *TBox*.

Let $C$ be a concept description, $R$ a role name, and $i, j$ individual names. Then $C(i)$ is a *concept assertion* and $R(i, j)$ a *role assertion*. A finite set of such assertions is called an ABox.

The meaning of $\mathcal{ALC}$-concept descriptions, TBoxes, and ABoxes can be defined with the help of a set-theoretic semantics.

**Definition 2 ($\mathcal{ALC}$ Semantics).** An $\mathcal{ALC}$-interpretation $\mathcal{I}$ is a pair $(\Delta^{\mathcal{I}}, \cdot^{\mathcal{I}})$, where $\Delta^{\mathcal{I}}$ is a nonempty set, the *domain* of the interpretation, and $\cdot^{\mathcal{I}}$ is the *interpretation function*. The interpretation function maps

- each concept name $A$ to a subset $A^{\mathcal{I}}$ of $\Delta^{\mathcal{I}}$,

- each role name $R$ to a subset $R^{\mathcal{I}}$ of $\Delta^{\mathcal{I}} \times \Delta^{\mathcal{I}}$,

- each individual name $i$ to an element $i^{\mathcal{I}}$ of $\Delta^{\mathcal{I}}$ such that different names are mapped to different elements (unique name assumption).

For a role name $R$ and an element $a \in \Delta^{\mathcal{I}}$ we define $R^{\mathcal{I}}(a) := \{b \mid (a, b) \in R^{\mathcal{I}}\}$. The interpretation function can inductively be extended to complex concepts as follows:

$$(\neg C)^{\mathcal{I}} := \Delta^{\mathcal{I}} \setminus C^{\mathcal{I}}$$
$$(C \sqcap D)^{\mathcal{I}} := C^{\mathcal{I}} \cap D^{\mathcal{I}}$$
$$(C \sqcup D)^{\mathcal{I}} := C^{\mathcal{I}} \cup D^{\mathcal{I}}$$
$$(\exists R.C)^{\mathcal{I}} := \{a \in \Delta^{\mathcal{I}} \mid R^{\mathcal{I}}(a) \cap C^{\mathcal{I}} \neq \emptyset\}$$
$$(\forall R.C)^{\mathcal{I}} := \{a \in \Delta^{\mathcal{I}} \mid R^{\mathcal{I}}(a) \subseteq C^{\mathcal{I}}\}$$

An interpretation $\mathcal{I}$ is a *model of the TBox* $\mathcal{T}$ iff it satisfies $C^{\mathcal{I}} \subseteq D^{\mathcal{I}}$ for all GCIs $C \sqsubseteq D$ in $\mathcal{T}$. It is a *model of the ABox* $\mathcal{A}$ iff it satisfies $i^{\mathcal{I}} \in C^{\mathcal{I}}$ for all concept assertions $C(i) \in \mathcal{A}$ and $(i^{\mathcal{I}}, j^{\mathcal{I}}) \in R^{\mathcal{I}}$ for all role assertions $R(i, j) \in \mathcal{A}$. Finally, $\mathcal{I}$ is a *model of an ABox relative to a TBox* iff it is a model of both the ABox and the TBox.

Given this semantics, we can now formally define the relevant inference problems.

**Definition 3 (Inferences).** Let $C$ and $D$ be concept descriptions, $i$ an individual name, $\mathcal{T}$ a TBox, and $\mathcal{A}$ an ABox. We say that $C$ *subsumes* $D$ *relative to the TBox* $\mathcal{T}$ ($D \sqsubseteq_{\mathcal{T}} C$) iff $D^{\mathcal{I}} \subseteq C^{\mathcal{I}}$ for all models $\mathcal{I}$ of $\mathcal{T}$. The concept description $C$ is *satisfiable relative to the TBox* $\mathcal{T}$ iff there exists a model $\mathcal{I}$ of $\mathcal{T}$ such that $C^{\mathcal{I}} \neq \emptyset$. The individual $i$ is an *instance* of $C$ in the ABox $\mathcal{A}$ *relative to the TBox* $\mathcal{T}$ iff $i^{\mathcal{I}} \in C^{\mathcal{I}}$ for all models of $\mathcal{A}$ relative to $\mathcal{T}$. The ABox $\mathcal{A}$ is *consistent relative to the TBox* $\mathcal{T}$ iff there exists a model of $\mathcal{A}$ relative to $\mathcal{T}$.

These three inferences can also be considered without reference to a TBox: $C$ *subsumes* $D$ ($C$ is *satisfiable*) iff $C$ subsumes $D$ ($C$ is satisfiable) relative to the empty TBox, and $i$ is an *instance* of $C$ in $\mathcal{A}$ ($\mathcal{A}$ is *consistent*) iff $i$ is an instance of $C$ in $\mathcal{A}$ ($\mathcal{A}$ is consistent) relative to the empty TBox.

We restrict our attention to DLs that are propositionally closed (i.e., allow for the Boolean operators conjunction, disjunction, and negation). Consequently, subsumption can be reduced to (un)satisfiability since $C \sqsubseteq_{\mathcal{T}} D$ iff $C \sqcap \neg D$ is unsatisfiable relative to $\mathcal{T}$. Conversely, (un)satisfiability can be reduced to subsumption since $C$ is unsatisfiable relative to $\mathcal{T}$ iff $C \sqsubseteq_{\mathcal{T}} \bot$. For this reason, it is irrelevant whether we consider the subsumption or the satisfiability problem in our results concerning the transfer of decidability of these problems from component DLs to their fusion (informally called *transfer results* in the following).

Similarly, the instance problem can be reduced to the (in)consistency problem and vice versa: $i$ is an instance of $C$ in $\mathcal{A}$ relative to $\mathcal{T}$ iff $\mathcal{A} \cup \{\neg C(i)\}$ is inconsistent relative to $\mathcal{T}$; and $\mathcal{A}$ is inconsistent relative to $\mathcal{T}$ iff $i$ is an instance of $\bot$ in $\mathcal{A}$ relative to $\mathcal{T}$, where $i$ is an arbitrary individual name. Consequently, it is irrelevant whether we consider the instance problem or the consistency problem in our transfer results.

Finally, the satisfiability problem can be reduced to the consistency problem: $C$ is satisfiable relative to $\mathcal{T}$ iff the ABox $\{C(i)\}$ is consistent relative to $\mathcal{T}$, where $i$ is an arbitrary individual name. However, the converse need not be true. It should be obvious that this implies that a transfer result for the satisfiability problem does not yield the corresponding transfer result for the consistency problem: from decidability of the consistency problem for the component DLs we can only deduce decidability of the satisfiability problem in their fusion. What might be less obvious is that a transfer result for the consistency problem need not imply the corresponding transfer result for the satisfiability problem: if the satisfiability problems in the component DLs are decidable, then the transfer result for the consistency problem can just not be applied (since the prerequisite of this transfer result, namely, decidability of the *consistency* problem in the component DLs, need not be satisfied). However, we will show that the method used to show the transfer result for the consistency problem also applies to the satisfiability problem.

## 2.1 More expressive DLs

There are several possibilities for extending $\mathcal{ALC}$ in order to obtain a more expressive DL. The three most prominent are adding additional concept constructors, adding role constructors, and formulating restrictions on role interpretations. In addition to giving examples for such extensions, we also introduce a naming scheme for the obtained DLs. Additional concept constructors are indicated by appending caligraphic letters to the language name, role constructors by symbols in superscript, and restrictions on roles by letters in subscript.

We start with introducing restrictions on role interpretations, since we need to refer to such restrictions when defining certain concept constructors.

### 2.1.1 RESTRICTIONS ON ROLE INTERPRETATIONS

These restrictions enforce the interpretations of roles to satisfy certain properties, such as functionality, transitivity, etc. We consider three prominent examples:

1. **Functional roles.** Here one considers a subset $N_F$ of the set of role names $N_R$, whose elements are called *features*. An interpretation must map features $f \in N_F$ to functional binary relations $f^\mathcal{I} \subseteq \Delta^\mathcal{I} \times \Delta^\mathcal{I}$, i.e., relations satisfying $\forall a, b, c. f^\mathcal{I}(a, b) \wedge f^\mathcal{I}(a, c) \to b = c$. We will sometimes treat functional relations as partial functions, and write $f^\mathcal{I}(a) = b$ rather than $f^\mathcal{I}(a, b)$. $\mathcal{ALC}$ extended with features is denoted by $\mathcal{ALC}_f$.

2. **Transitive roles.** Here one considers a subset $N_{R^+}$ of $N_R$. Role names $R \in N_{R^+}$ are called *transitive roles*. An interpretation must map transitive roles $R \in N_{R^+}$ to transitive binary relations $R^\mathcal{I} \subseteq \Delta^\mathcal{I} \times \Delta^\mathcal{I}$. $\mathcal{ALC}$ extended with transitive roles is denoted by $\mathcal{ALC}_{R^+}$.

3. **Role hierarchies.** A *role inclusion axiom* is an expression of the form $R \sqsubseteq S$ with $R, S \in N_R$. A finite set $H$ of role inclusion axioms is called a *role hierarchy*. An interpretation must satisfy $R^\mathcal{I} \subseteq S^\mathcal{I}$ for all $R \sqsubseteq S \in H$. $\mathcal{ALC}$ extended with a role hierarchy $H$ is denoted by $\mathcal{ALC}_{\mathcal{H}(H)}$. If $H$ is clear from the context or irrelevant, we write $\mathcal{ALC}_\mathcal{H}$ instead of $\mathcal{ALC}_{\mathcal{H}(H)}$.

The above restrictions can also be combined with each other. For example, $\mathcal{ALC}_{\mathcal{H}R^+}$ is $\mathcal{ALC}$ with a role hierarchy and transitive roles.

Transitive roles in DLs were first investigated by Sattler (1996). Features were introduced in DLs by Hollunder and Nutt (1990) and (under the name "attributes") in the CLASSIC system (Brachman et al., 1991), in both cases in conjunction with feature agreements and disagreements (see concept constructors below). Features without agreements and disagreements are, e.g., used in the DL $\mathcal{SHIF}$ (Horrocks & Sattler, 1999), albeit in a more expressive "local" way, where functionality can be asserted to hold at certain individuals, but not necessarily on the whole model. According to our naming scheme, we indicate the presence of features in a DL by the letter $f$ in subscript.[1]

A remark on role hierarchies is also in order: in our definition, if $H_1$ and $H_2$ are different role hierarchies, then $\mathcal{ALC}_{\mathcal{H}(H_1)}$ and $\mathcal{ALC}_{\mathcal{H}(H_2)}$ are different DLs. In the DL literature, usually only one logic $\mathcal{ALC}_\mathcal{H}$ is defined and role hierarchies are treated like TBoxes, i.e., satisfiability and subsumption are defined relative to TBoxes *and* role hierarchies (see, e.g., Horrocks, 1998). For our purposes, however, it is more convenient to define one DL per role hierarchy since distinct role hierarchies impose distinct restrictions on the interpretation of roles. The advantages of this approach will become clear later on when frames and abstract description systems are introduced.

---

1. Note that some authors (e.g., Horrocks & Sattler, 1999) use an appended $\mathcal{F}$ to denote local features. Following Hollunder and Nutt (1990), we will use $\mathcal{F}$ to denote a DL that allows for feature agreements (see below).

| Name | Syntax | Semantics | Symbol |
|------|--------|-----------|--------|
| Unqualified | $\geq nR$ | $\{a \in \Delta^{\mathcal{I}} \mid |R^{\mathcal{I}}(a)| \geq n\}$ | $\mathcal{N}$ |
| number restrictions | $\leq nR$ | $\{a \in \Delta^{\mathcal{I}} \mid |R^{\mathcal{I}}(a)| \leq n\}$ | |
| Qualified | $\geq nR.C$ | $\{a \in \Delta^{\mathcal{I}} \mid |R^{\mathcal{I}}(a) \cap C^{\mathcal{I}}| \geq n\}$ | $\mathcal{Q}$ |
| number restrictions | $\leq nR.C$ | $\{a \in \Delta^{\mathcal{I}} \mid |R^{\mathcal{I}}(a) \cap C^{\mathcal{I}}| \leq n\}$ | |
| Nominals | $I$ | $I^{\mathcal{I}} \subseteq \Delta^{\mathcal{I}}$ with $|I^{\mathcal{I}}| = 1$ | $\mathcal{O}$ |
| Feature agreement | $u_1 \downarrow u_2$ | $\{a \in \Delta^{\mathcal{I}} \mid \exists b \in \Delta^{\mathcal{I}}.\ u_1^{\mathcal{I}}(a) = b = u_2^{\mathcal{I}}(a)\}$ | $\mathcal{F}$ |
| and disagreement | $u_1 \uparrow u_2$ | $\{a \in \Delta^{\mathcal{I}} \mid \exists b_1, b_2 \in \Delta^{\mathcal{I}}.$ | |
| | | $\qquad u_1^{\mathcal{I}}(a) = b_1 \neq b_2 = u_2^{\mathcal{I}}(b_1)\}$ | |

Figure 1: Some description logic concept constructors.

### 2.1.2 CONCEPT CONSTRUCTORS

Concept constructors take concept and/or role descriptions and transform them into more complex concept descriptions. In addition to the constructors available in $\mathcal{ALC}$, various other concept constructors are considered in the DL literature. A small collection of such constructors can be found in Figure 1, where $|S|$ denotes the cardinality of a set $S$. The symbols in the rightmost column indicate the naming scheme for the resulting DL. As mentioned above the name modifiers for concept constructors are not written in subscript, they are appended to the language name. For example, $\mathcal{ALC}_{\mathcal{HR}^+}$ extended with qualified number restrictions is called $\mathcal{ALCQ}_{\mathcal{HR}^+}$. The syntax of the extended DLs is as expected, i.e., the constructors may be arbitrarily combined. The semantics is obtained by augmenting the semantics of $\mathcal{ALC}$ with the appropriate conditions, which can be found in the third column in Figure 1. Nominals and feature (dis)agreements need some more explanation:

- **Nominals.** We consider a set $N_O$ of (names for) *nominals*, which is pairwise disjoint to the sets $N_C$, $N_R$, and $N_I$. Elements from $N_O$ are often denoted by $I$ (possibly with index). An interpretation must map nominals to singleton subsets of $\Delta^{\mathcal{I}}$. The intention underlying nominals is that they stand for elements of $\Delta$, just like individual names. However, since we want to use the nominal $I \in N_O$ as a (nullary) concept constructor, $\mathcal{I}$ must interpret them by a set, namely the singleton set consisting of the individual that $I$ denotes.

- **Feature (dis)agreements.** $\mathcal{ALCF}$ is the extension of $\mathcal{ALC}_f$ with feature agreements and disagreements. Beside the additional concept constructors, $\mathcal{ALCF}$ uses feature chains as part of the (dis)agreement constructor. A *feature chain* is an expression of the form $u = f_1 \circ \cdots \circ f_n$. The interpretation $u^{\mathcal{I}}$ of such a feature chain is just the composition of the partial functions $f_1^{\mathcal{I}}, \ldots, f_n^{\mathcal{I}}$, where composition is to be read from left to right.

DLs including nominals or feature (dis)agreements and additional concept constructors or restrictions on role interpretations are defined (and named) in the obvious way.

Number restriction are available in almost all DL systems. The DL $\mathcal{ALCN}$ (i.e., $\mathcal{ALC}$ extended with number restrictions) was first treated by Hollunder and Nutt (1990), as was $\mathcal{ALCF}$. The DL $\mathcal{ALCQ}$ was first investigated by Hollunder and Baader (1991), and $\mathcal{ALCO}$ by Schaerf (1994).

| Name | Syntax | Semantics | Symbol |
|------|--------|-----------|--------|
| Role composition | $R_1 \circ R_2$ | $\{(a,b) \in \Delta^{\mathcal{I}} \times \Delta^{\mathcal{I}} \mid$ $\exists c \in \Delta^{\mathcal{I}}.\ (a,c) \in R_1^{\mathcal{I}} \wedge (c,b) \in R_2^{\mathcal{I}}\}$ | $\circ$ |
| Role complement | $\overline{R}$ | $\{(a,b) \in \Delta^{\mathcal{I}} \times \Delta^{\mathcal{I}} \mid (a,b) \notin R^{\mathcal{I}}\}$ | $\neg$ |
| Role conjunction | $R_1 \sqcap R_2$ | $\{(a,b) \in \Delta^{\mathcal{I}} \times \Delta^{\mathcal{I}} \mid (a,b) \in R_1^{\mathcal{I}} \wedge (a,b) \in R_2^{\mathcal{I}}\}$ | $\sqcap$ |
| Role disjunction | $R_1 \sqcup R_2$ | $\{(a,b) \in \Delta^{\mathcal{I}} \times \Delta^{\mathcal{I}} \mid (a,b) \in R_1^{\mathcal{I}} \vee (a,b) \in R_2^{\mathcal{I}}\}$ | $\sqcup$ |
| Inverse roles | $R^{-1}$ | $\{(a,b) \in \Delta^{\mathcal{I}} \times \Delta^{\mathcal{I}} \mid (b,a) \in R^{\mathcal{I}}\}$ | $-1$ |
| Transitive closure | $R^+$ | $\{(a,b) \in \Delta^{\mathcal{I}} \times \Delta^{\mathcal{I}} \mid (a,b) \in (R^{\mathcal{I}})^+\}$ | $+$ |
| Universal role | $U$ | $\Delta^{\mathcal{I}} \times \Delta^{\mathcal{I}}$ | $U$ |

For a binary relation $R$, $R^+$ denotes the transitive closure of $R$.

Figure 2: Some description logic role constructors.

### 2.1.3 ROLE CONSTRUCTORS

Role constructors allow us to build complex role descriptions. A collection of role constructors can be found in Figure 2. Again, the rightmost column indicates the naming scheme, where name modifiers for role constructors are written in superscript and separated by commas. For example, $\mathcal{ALCQ}$ with inverse roles and transitive closure is called $\mathcal{ALCQ}^{+,-1}$. In DLs admitting role constructors, the set of role descriptions is defined inductively, analogously to the set of concept descriptions. The semantics of role constructors is given in the third column of Figure 2. As with concept descriptions, it can be used to extend the interpretation function from role names to role descriptions.

In a DL with role constructors, role descriptions can be used wherever role names may be used in the corresponding DLs without role constructors. For example,

$$\exists (R_1 \sqcap R_3).C \sqcap \forall (R_2 \sqcup \overline{R_2}).\neg C$$

is an $\mathcal{ALC}^{\neg,\sqcap,\sqcup}$-concept description. This concept description is unsatisfiable since $R_2 \sqcup \overline{R_2}$ is equivalent to the universal role. Note that role descriptions can also be used within role assertions in an ABox.

The DL $\mathcal{ALC}^{\circ,\sqcup,+}$ was first treated by Baader (1991) (under the name $\mathcal{ALC}_{trans}$); Schild (1991) has shown that this DL is a notational variant of propositional dynamic logic (PDL). DLs with Boolean operators on roles were investigated by Lutz and Sattler (2000). The inverse operator was available in the system CRACK (Bresciani, Franconi, & Tessaris, 1995), and reasoning in DLs with inverse roles was, for example, investigated by Calvanese et al. (1998) and Horrocks et al. (2000). The universal role can be expressed using DLs with Boolean operators on roles (see the above example), and it can in turn be used to simulate general concept inclusion axioms within concept descriptions.

## 2.2 Restricting the syntax

Until now, constructors could be combined arbitrarily. Sometimes it makes sense to restrict the interaction between constructors since reasoning in the restricted DL may be easier than reasoning in the unrestricted DL. We will consider DLs imposing certain restrictions on

1. which roles may be used inside certain concept constructors,

2. which roles may be used inside certain role constructors,

3. the combination of role constructors, and

4. the role constructors that may be used inside certain concept constructors.

As an example for the first case, consider the fragment of $\mathcal{ALCQ}_{R^+}$ in which transitive roles may be used in existential and universal restrictions, but not in number restrictions (see, e.g., Horrocks et al., 2000).

As the result of taking the fusion of two DLs, we will obtain DLs whose set of roles $N_R$ is partitioned. For example, the fusion of $\mathcal{ALCQ}$ with $\mathcal{ALC}^{-1}$ yields a fragment of $\mathcal{ALCQ}^{-1}$ where $N_R$ is partitioned in two sets, say $N_{R_1}$ and $N_{R_2}$. In this fragment, the inverse role constructor and roles from $N_{R_2}$ may not be used within qualified number restrictions, while roles from $N_{R_1}$ may not be used inside the inverse role constructor.[2] Thus, this DL is an example for the first, the second, and the fourth case.

Now consider the DL $\mathcal{ALCF}$ introduced above, which does not only extend $\mathcal{ALC}_f$ with feature (dis)agreement as a concept constructor, but also provides the role composition constructor. However, the role chains built using composition have to be comprised exclusively of features and non-functional roles may not appear inside feature (dis)agreement. Hence, $\mathcal{ALCF}$ is also an example for the first, second, and fourth case.

As an example for the third case, the fragment of $\mathcal{ALC}^{\neg,\sqcap}$ in which role conjunction may not be used inside the role complement constructor is considered by Lutz and Sattler (2000).

For these restricted DLs, we do not introduce an explicit naming scheme. Note that, in this paper, we do *not* deal with DLs in which the combinability of concept constructors with each other is restricted since these DLs would not fit into the framework of abstract description systems introduced in the next section. An example of such a DL would be one with atomic negation of concepts, i.e., where negation may only be applied to concept names (e.g., the DL $\mathcal{AL}$ discussed by Donini et al., 1997).

## 3. Abstract description systems

In order to define the fusion of DLs and prove general results for fusions of DLs, one needs a formal definition of what are "description logics". Since there exists a wide variety of DLs with very different characteristics, we introduce a very general formalization, which should cover all of the DLs considered in the literature, but also includes logics that would usually not be subsumed under the name DL.

### 3.1 Syntax and semantics

The *syntax* of an abstract description system is given by its *abstract description language*, which determines a set of terms, term assertions, and object assertions. In this setting, concept descriptions are represented by terms that are built using the abstract description language. General inclusion axioms in DLs are represented by term assertions and ABox assertions in DLs are represented by object assertions.

---

2. This will become clearer once we have given a formal definition of the fusion.

**Definition 4 (Abstract description language).** An *abstract description language* (ADL) is determined by a countably infinite set $V$ of set variables, a countably infinite set $\mathcal{X}$ of object variables, a (possibly infinite) countable set $\mathcal{R}$ of relation symbols of arity two,[3] and a (possibly infinite) countable set $\mathcal{F}$ of functions symbols $f$, which are equipped with arities $n_f$. All these sets have to be pairwise disjoint.

The *terms* $t_j$ of this ADL are built using the follow syntax rules:

$$t_j \quad \longrightarrow \quad x,\ \neg t_1,\ t_1 \wedge t_2,\ t_1 \vee t_2,\ f(t_1, \ldots, t_{n_f}),$$

where $x \in V$, $f \in \mathcal{F}$, and the Boolean operators $\neg, \wedge, \vee$ are different from all function symbols in $\mathcal{F}$. For a term $t$, we denote by $var(t)$ the set of set variables used in $t$. The symbol $\top$ is used as an abbreviation of $x \vee \neg x$ and $\bot$ as an abbreviation for $x \wedge \neg x$ (where $x$ is a set variable).

The *term assertions* of this ADL are

- $t_1 \sqsubseteq t_2$, for all terms $t_1, t_2$,

and the *object assertions* are

- $R(a, b)$, for $a, b \in \mathcal{X}$ and $R \in \mathcal{R}$;

- $(a : t)$, for $a \in \mathcal{X}$ and $t$ a term.

The sets of term and object assertions together form the set of *assertions* of the ADL.

From the DL point of view, the set variables correspond to concept names, object variables to individual names, relation symbols to roles, and the Boolean operators as well as the function symbols correspond to concept constructors. Thus, terms correspond to concept descriptions. As an example, let us view concept descriptions of the DL $\mathcal{ALCN}^{\sqcap}$, i.e., $\mathcal{ALC}$ extended with number restrictions and conjunction of roles, as terms of an ADL. Value restrictions and existential restrictions can be seen as unary function symbols: for each role description $R$, we have the function symbols $f_{\forall R}$ and $f_{\exists R}$, which take a term $t_C$ (corresponding to the concept description $C$) and transform it into the more complex terms $f_{\forall R}(t_C)$ and $f_{\exists R}(t_C)$ (corresponding to the concept descriptions $\forall R.C$ and $\exists R.C$). Similarly, number restrictions can be seen as nullary function symbols: for each role description $R$ and each $n \in \mathbb{N}$, we have the function symbols $f_{\geq nR}$ and $f_{\leq nR}$. Hence, the $\mathcal{ALCN}^{\sqcap}$-concept description $A \sqcap \forall (R_1 \sqcap R_2).\neg (B \sqcap (\geq 2R_1))$ corresponds to the term $x_A \wedge f_{\forall (R_1 \sqcap R_2)}(\neg (x_B \wedge f_{(\geq 2R_1)}))$. We will analyze the connection between ADLs and DLs more formally later on.

The *semantics* of abstract description systems is defined based on *abstract description models*. These models are the general semantic structures in which the terms of the ADL are interpreted. It should already be noted here, however, that an abstract description system usually does not take into account all abstract description models available for the language: it allows only for a *selected subclass* of these models. This subclass determines the semantics of the system.

---

3. To keep things simpler, we restrict our attention to the case of binary predicates, i.e., roles in DL. However, the results can easily be extended to $n$-ary predicates.

**Definition 5.** Let $L$ be an ADL as in Definition 4. An *abstract description model* (ADM) for $L$ is of the form

$$\mathfrak{W} = \left\langle W, \mathcal{F}^{\mathfrak{W}} = \{f^{\mathfrak{W}} \mid f \in \mathcal{F}\}, \mathcal{R}^{\mathfrak{W}} = \{R^{\mathfrak{W}} \mid R \in \mathcal{R}\} \right\rangle,$$

where $W$ is a nonempty set, the $f^{\mathfrak{W}}$ are functions mapping every sequence $\langle X_1, \ldots, X_{n_f} \rangle$ of subsets of $W$ to a subset of $W$, and the $R^{\mathfrak{W}}$ are binary relations on $W$.

Since ADMs do not interpret variables, we need an assignment that assigns a subset of $W$ to each set variable, before we can evaluate terms in an ADM. To evaluate object assertions, we need an additional assignment that assigns an element of $W$ to each object variable.

**Definition 6.** Let $\mathcal{L}$ be an ADL and $\mathfrak{W} = \langle W, \mathcal{F}^{\mathfrak{W}}, \mathcal{R}^{\mathfrak{W}} \rangle$ be an ADM for $\mathcal{L}$. An *assignment* for $\mathfrak{W}$ is a pair $\mathcal{A} = (\mathcal{A}_1, \mathcal{A}_2)$ such that $\mathcal{A}_1$ is a mapping from the set of set variables $V$ into $2^W$, and $\mathcal{A}_2$ is an injective[4] mapping from the set of object variables $\mathcal{X}$ into $W$. Let $\mathfrak{W}$ be an ADM and $\mathcal{A} = (\mathcal{A}_1, \mathcal{A}_2)$ be an assignment for $\mathfrak{W}$. With each $\mathcal{L}$-term $t$, we inductively associate a value $t^{\mathfrak{W},\mathcal{A}}$ in $2^W$ as follows:

- $x^{\mathfrak{W},\mathcal{A}} := \mathcal{A}_1(x)$ for all variables $x \in V$,

- $(\neg t)^{\mathfrak{W},\mathcal{A}} := W \setminus (t)^{\mathfrak{W},\mathcal{A}}$, $(t_1 \wedge t_2)^{\mathfrak{W},\mathcal{A}} := t_1^{\mathfrak{W},\mathcal{A}} \cap t_2^{\mathfrak{W},\mathcal{A}}$, $(t_1 \vee t_2)^{\mathfrak{W},\mathcal{A}} := t_1^{\mathfrak{W},\mathcal{A}} \cup t_2^{\mathfrak{W},\mathcal{A}}$,

- $f(t_1, \ldots, t_{n_f})^{\mathfrak{W},\mathcal{A}} := f^{\mathfrak{W}}(t_1^{\mathfrak{W},\mathcal{A}}, \ldots, t_{n_f}^{\mathfrak{W},\mathcal{A}})$.

If $x_1, \ldots, x_n$ are the set variables occurring in $t$, then we often write $t^{\mathfrak{W}}(X_1, \ldots, X_n)$ as shorthand for $t^{\mathfrak{W},\mathcal{A}}$, where $\mathcal{A}$ is an assignment with $x_i^{\mathcal{A}} = X_i$ for $1 \leq i \leq n$.

The truth-relation $\models$ between $\langle \mathfrak{W}, \mathcal{A} \rangle$ and assertions is defined as follows:

- $\langle \mathfrak{W}, \mathcal{A} \rangle \models R(a, b)$ iff $\mathcal{A}_2(a) R^{\mathfrak{W}} \mathcal{A}_2(b)$,

- $\langle \mathfrak{W}, \mathcal{A} \rangle \models a : t$ iff $\mathcal{A}_2(a) \in t^{\mathfrak{W},\mathcal{A}}$,

- $\langle \mathfrak{W}, \mathcal{A} \rangle \models t_1 \sqsubseteq t_2$ iff $t_1^{\mathfrak{W},\mathcal{A}} \subseteq t_2^{\mathfrak{W},\mathcal{A}}$.

In this case we say that the assertion is satisfied in $\langle \mathfrak{W}, \mathcal{A} \rangle$. If, for an ADM $\mathfrak{W}$ and a set of assertions $\Gamma$, there exists an assignment $\mathcal{A}$ for $\mathfrak{W}$ such that each assertion in $\Gamma$ is satisfied in $\langle \mathfrak{W}, \mathcal{A} \rangle$, then $\mathfrak{W}$ is a *model for* $\Gamma$.

There are two differences between ADMs and DL interpretations. First, in a DL interpretation, the interpretation of the role names fixes the interpretation of the function symbols corresponding to concept constructors that involve roles (like value restrictions, number restrictions, etc.). The interpretation of the concept names corresponds to an assignment. Thus, a DL model is an ADM together with an assignment, whereas an ADM alone corresponds to what is called frame in modal logics. Second, in DL the roles used in concept constructors may, of course, also occur in role assertions. In contrast, the definition of ADMs per se does not enforce any connection between the interpretation of the function symbols and the interpretation of the relation symbols. Such connections can, however, be enforced by restricting the attention to a subclass of all possible ADMs for the ADL.

---

4. This corresponds to the unique name assumption.

**Definition 7.** An *abstract description system* (ADS) is a pair $(\mathcal{L}, \mathcal{M})$, where $\mathcal{L}$ is an ADL and $\mathcal{M}$ is a class of ADMs for $\mathcal{L}$ that is closed under isomorphic copies.[5]

From the DL point of view, the choice of the class $\mathcal{M}$ defines the semantics of the concept and role constructors, and it allows us, e.g., to incorporate restrictions on role interpretations. In this sense, the ADS can be viewed as determining a (description) logic.

To be more concrete, in a DL interpretation the interpretation of the function symbols is determined by the interpretation of the role names. Thus one can, for example, restrict the class of models to ADMs that interpret a certain role as a transitive relation or as the composition of two other roles. Another restriction that can be realized by the choice of $\mathcal{M}$ is that nominals (corresponding to nullary function symbols) must be interpreted as singleton sets.

Let us now define reasoning problems for abstract description systems. We will introduce satisfiability of sets of assertions (with or without term assertions), which corresponds to consistency of ABoxes (with or without GCIs), and satisfiability of terms (with or without term assertions), which corresponds to satisfiability of concept descriptions (with or without GCIs).

**Definition 8.** Given an ADS $(\mathcal{L}, \mathcal{M})$, a finite set of assertions $\Gamma$ is called *satisfiable* in $(\mathcal{L}, \mathcal{M})$ iff there exists an ADM $\mathfrak{W} \in \mathcal{M}$ and an assignment $\mathcal{A}$ for $\mathfrak{W}$ such that $\langle \mathfrak{W}, \mathcal{A} \rangle$ satisfies all assertions in $\Gamma$. The *term* $t$ is called *satisfiable* in $(\mathcal{L}, \mathcal{M})$ iff $\{a : t\}$ is satisfiable in $(\mathcal{L}, \mathcal{M})$, where $a$ is an arbitrary object variable.

- The *satisfiability problem* for $(\mathcal{L}, \mathcal{M})$ is concerned with the following question: given a finite set of *object* assertions $\Gamma$ of $\mathcal{L}$, is $\Gamma$ satisfiable in $(\mathcal{L}, \mathcal{M})$.

- The *relativized satisfiability problem* for $(\mathcal{L}, \mathcal{M})$ is concerned with the following question: given a finite set of assertions $\Gamma$ of $\mathcal{L}$, is $\Gamma$ satisfiable in $(\mathcal{L}, \mathcal{M})$.

- The *term satisfiability problem* for $(\mathcal{L}, \mathcal{M})$ is concerned with the following question: given a term $t$ of $\mathcal{L}$, is $t$ satisfiable in $(\mathcal{L}, \mathcal{M})$.

- The *relativized term satisfiability problem* for $(\mathcal{L}, \mathcal{M})$ is concerned with the following question: given a term $t$ and a set of *term* assertions $\Gamma$ of $\mathcal{L}$, is $\{a : t\} \cup \Gamma$ satisfiable in $(\mathcal{L}, \mathcal{M})$.

In the next section, we will define the fusion of two ADSs, and show that (relativized) satisfiability is decidable in the fusion if (relativized) satisfiability in the component ADSs is decidable. For these transfer results to hold, we must restrict ourselves to so-called local ADSs.

**Definition 9.** Given a family $(\mathfrak{W}_p)_{p \in P}$ of ADMs $\mathfrak{W}_p = \langle W_p, \mathcal{F}^{\mathfrak{W}_p}, \mathcal{R}^{\mathfrak{W}_p} \rangle$ over pairwise disjoint domains $W_p$, we say that $\mathfrak{W} = \langle W, \mathcal{F}^{\mathfrak{W}}, \mathcal{R}^{\mathfrak{W}} \rangle$ is the *disjoint union* of $(\mathfrak{W}_p)_{p \in P}$ iff

- $W = \bigcup_{p \in P} W_p$,

---

5. Intuitively, this means that, if an ADM $\mathfrak{W}$ belongs to $\mathcal{M}$, then all ADMs that differ from it only w.r.t. the names of the elements in its domain $W$ also belong to $\mathcal{M}$.

- $f^{\mathfrak{W}}(X_1, \ldots, X_{n_f}) = \bigcup_{p \in P} f^{\mathfrak{W}_p}(X_1 \cap W_p, \ldots, X_{n_f} \cap W_p)$ for all $f \in \mathcal{F}$ and $X_1, \ldots, X_{n_f} \subseteq W$,

- $R^{\mathfrak{W}} = \bigcup_{p \in P} R^{\mathfrak{W}_p}$ for all $R \in \mathcal{R}$.

An ADS $S = (\mathcal{L}, \mathcal{M})$ is called *local* iff $\mathcal{M}$ is closed under disjoint unions.

In the remainder of this section, we first analyze the connection between ADSs and DLs in more detail, and then comment on the relationship to modal logics.

### 3.2 Correspondence to description logics

We will show that the DLs introduced in Section 2 correspond to ADSs. In order to do this, we first need to introduce frames, a notion well-known from modal logic. Let $L$ be one of the DLs introduced in Section 2.

**Definition 10 (Frames).** An $L$-frame $\mathfrak{F}$ is a pair $(\Delta^{\mathfrak{F}}, \cdot^{\mathfrak{F}})$, where $\Delta^{\mathfrak{F}}$ is a nonempty set, called the *domain* of $\mathfrak{F}$, and $\cdot^{\mathfrak{F}}$ is the *interpretation function*, which maps

- each nominal $I$ to a singleton subset $I^{\mathfrak{F}}$ of $\Delta^{\mathfrak{F}}$, and

- each role name $R$ to a subset $R^{\mathfrak{F}}$ of $\Delta^{\mathfrak{F}} \times \Delta^{\mathfrak{F}}$ such that the restrictions for role interpretations in $L$ are satisfied. For example, in $\mathcal{ALC}_{R^+}$, each $R \in N_{R^+}$ is mapped to a transitive binary relation.

The interpretation function $\cdot^{\mathfrak{F}}$ can inductively be extended to complex roles in the obvious way, i.e., by interpreting the role constructors in $L$ according to their semantics as given in Figure 2.

An interpretation $\mathcal{I}$ is *based* on a frame $\mathfrak{F}$ iff $\Delta^{\mathcal{I}} = \Delta^{\mathfrak{F}}$, $R^{\mathcal{I}} = R^{\mathfrak{F}}$ for all roles $R \in N_R$, and $I^{\mathcal{I}} = I^{\mathfrak{F}}$ for all nominals $I \in N_O$.

A frame can be viewed as an interpretation that is partial in the sense that the interpretation of individual and concept names is not fixed. Note that (in contrast to the case of concept and individual names) the interpretation of nominals is already fixed in the frame. The reason for this is that, if we do not interpret nominals in the frame, then we have to treat them as set variables on the ADS side. These would, however, have to be variables to which only singleton sets may be assigned. Since such a restriction is not possible in the framework of ADSs as defined above, we interpret nominals in the frame. The consequence is that they correspond to functions of arity 0 on the ADS side.

Now, we define the abstract description system $S = (\mathcal{L}, \mathcal{M})$ corresponding to a DL $L$. It is straightforward to translate the syntax of $L$ into an abstract description language $\mathcal{L}$.

**Definition 11 (Corresponding ADL).** Let $L$ be a DL with concept and role constructors as well as restrictions on role interpretations as introduced in Section 2. The *corresponding abstract description language* $\mathcal{L}$ is defined as follows. For every concept name $A$ in $L$, there exists a set variable $x_A$ in $\mathcal{L}$, and for every individual name $i$ in $L$ there exists an object variable $a_i$ in $\mathcal{L}$. Let $\mathcal{R}$ be the set of (possibly complex) role descriptions of $L$. The set of relation symbols of $\mathcal{L}$ is $\mathcal{R}$, and the set of function symbols of $\mathcal{L}$ is the smallest set containing

1. for every role description $R \in \mathcal{R}$, unary function symbols $f_{\exists R}$ and $f_{\forall R}$,

2. if $L$ provides unqualified number restrictions, then, for every $n \in \mathbb{N}$ and every role description $R \in \mathcal{R}$, function symbols $f_{\geq nR}$ and $f_{\leq nR}$ of arity 0,

3. if $L$ provides qualified number restrictions, then, for every $n \in \mathbb{N}$ and every role $R \in \mathcal{R}$, unary function symbols $f_{\geq nR}$ and $f_{\leq nR}$,

4. if $L$ provides nominals, then, for every $I \in N_O$, a function symbol $f_I$ of arity 0,

5. if $L$ provides feature agreement and disagreement, then, for every pair of feature chains $(u_1, u_2)$, two function symbols $f_{u_1 \downarrow u_2}$ and $f_{u_1 \uparrow u_2}$ of arity 0.

For an $L$-concept description $C$, let $t_C$ denote the representation of $C$ as an $\mathcal{L}$-term, which is defined in the obvious way: concept names $A$ are translated into set variables $x_A$, the concept constructors $\neg$, $\sqcap$, and $\sqcup$ are mapped to $\neg$, $\wedge$, and $\vee$, respectively, and all other concept constructors are translated to the corresponding function symbols. Obviously, both the sets of function and relation symbols of $\mathcal{L}$ may be infinite.

An example of the translation of concept descriptions into terms of an ADL was already given above: the $\mathcal{ALCN}^{\sqcap}$-concept description $A \sqcap \forall (R_1 \sqcap R_2).\neg(B \sqcap (\geq 2R_1))$ corresponds to the term $x_A \wedge f_{\forall (R_1 \sqcap R_2)}(\neg(x_B \wedge f_{(\geq 2R_1)}))$.

We now define the set of abstract description models $\mathcal{M}$ corresponding to the DL $L$. For every $L$-frame, $\mathcal{M}$ contains a corresponding ADM.

**Definition 12 (Corresponding ADM).** Let $\mathfrak{F} = (\Delta^{\mathfrak{F}}, \cdot^{\mathfrak{F}})$ be a frame. The *corresponding abstract description model* $\mathfrak{W} = \langle W, \mathcal{F}^{\mathfrak{W}}, \mathcal{R}^{\mathfrak{W}} \rangle$ has domain $W := \Delta^{\mathfrak{F}}$. The relation symbols of $\mathcal{L}$ are just the role descriptions of $L$, and thus they are interpreted in the frame $\mathfrak{F}$. For each relation symbol $R \in \mathcal{R}$ we can hence define $R^{\mathfrak{W}} := R^{\mathfrak{F}}$.

To define $\mathcal{F}^{\mathfrak{W}}$, we need to define $f^{\mathfrak{W}}$ for every nullary function symbol $f$ in $\mathcal{L}$, and $f^{\mathfrak{W}}(X)$ for every unary function symbol $f$ in $\mathcal{L}$ and every $X \subseteq \Delta^{\mathcal{I}}$. Let $A$ be an arbitrary concept name. For each $X \subseteq \Delta^{\mathfrak{F}}$, let $\mathcal{I}_X$ be the interpretation based on $\mathfrak{F}$ mapping the concept name $A$ to $X$ and every other concept name to $\emptyset$.[6] To define $f^{\mathfrak{W}}$, we make a case distinction according to the type of $f$:

1. $f_{\exists R}^{\mathfrak{W}}(X) := (\exists R.A)^{\mathcal{I}_X}, \quad f_{\forall R}^{\mathfrak{W}}(X) := (\forall R.A)^{\mathcal{I}_X},$

2. $f_{\geq nR}^{\mathfrak{W}} := (\geq nR)^{\mathcal{I}_\emptyset}, \quad f_{\leq nR}^{\mathfrak{W}} := (\leq nR)^{\mathcal{I}_\emptyset},$

3. $f_{\geq nR}^{\mathfrak{W}}(X) := (\geq nR.A)^{\mathcal{I}_X}, \quad f_{\leq nR}^{\mathfrak{W}}(X) := (\leq nR.A)^{\mathcal{I}_X},$

4. $f_I^{\mathfrak{W}} := I^{\mathcal{I}_\emptyset},$

5. $f_{u_1 \downarrow u_2}^{\mathfrak{W}} = (u_1 \downarrow u_2)^{\mathcal{I}_\emptyset}, \quad f_{u_1 \uparrow u_2}^{\mathfrak{W}} = (u_1 \uparrow u_2)^{\mathcal{I}_\emptyset}.$

The class of ADMs $\mathcal{M}$ thus obtained from a DL $L$ is obviously closed under isomorphic copies since this also holds for the set of $L$-frames (independently of which DL $L$ we consider). Hence, the tuple $S = (\mathcal{L}, \mathcal{M})$ corresponding to a DL $L$ is indeed an ADS.

As an example, let us view the DL $\mathcal{ALCN}^{\sqcap}$ as an ADS. The ADL $\mathcal{L}$ corresponding to $\mathcal{ALCN}^{\sqcap}$ has already been discussed. Thus, we concentrate on the class of ADMs $\mathcal{M}$ induced by the frames of $\mathcal{ALCN}^{\sqcap}$. Assume that $\mathfrak{F}$ is such a frame, i.e., $\mathfrak{F}$ consists of a nonempty domain and interpretations $R^{\mathfrak{F}}$ of the role names $R$. The ADM $\mathfrak{W} = \langle W, \mathcal{F}^{\mathfrak{W}}, \mathcal{R}^{\mathfrak{W}} \rangle$ induced by $\mathfrak{F}$ is defined as follows. The set $W$ is identical to the domain of $\mathfrak{F}$. Each role description yields a relation symbol, which is interpreted in $\mathfrak{W}$ just as in the frame. For example, $(R_1 \sqcap R_2)^{\mathfrak{W}} = R_1^{\mathfrak{F}} \cap R_2^{\mathfrak{F}}$. It remains to define the interpretation of the function symbols. We illustrate this on two examples. First, consider the (unary) function symbol $f_{\forall(R_1 \sqcap R_2)}$. Given a subset $X$ of $W$, the function $f_{\forall(R_1 \sqcap R_2)}^{\mathfrak{W}}$ maps $X$ to

$$f_{\forall(R_1 \sqcap R_2)}^{\mathfrak{W}}(X) := \{w \in W \mid v \in X \text{ for all } v \text{ with } (w,v) \in R_1^{\mathfrak{F}} \cap R_2^{\mathfrak{F}}\},$$

i.e., the interpretation of the concept description $\forall(R_1 \sqcap R_2).A$ in the interpretations based on $\mathfrak{F}$ interpreting $A$ by $X$. Accordingly, the value of the constant symbol $f_{(\geq 2R)}$ in $\mathfrak{W}$ is given by the interpretation of $(\geq 2R)$ in the interpretations based on $\mathfrak{F}$.

---

6. Taking the empty set here is arbitrary.

It is easy to show that the interpretation of concept descriptions in $L$ coincides with the interpretation of the corresponding terms in $S = (\mathcal{L}, \mathcal{M})$.

**Lemma 13.** *Let $\mathfrak{F}$ be a frame, $\mathfrak{W} = \langle W, \mathcal{F}^{\mathfrak{W}}, \mathcal{R}^{\mathfrak{W}} \rangle$ be the ADM corresponding to $\mathfrak{F}$, $\mathcal{A} = (\mathcal{A}_1, \mathcal{A}_2)$ be an assignment for $\mathfrak{W}$, $C$ be a concept description, and let the concept names used in $C$ be among $A_1, \ldots, A_k$. For all interpretations $\mathcal{I}$ based on $\mathfrak{F}$ with $A_i^{\mathcal{I}} = \mathcal{A}_1(x_{A_i})$ for all $1 \leq i \leq k$, we have that*

$$C^{\mathcal{I}} = t_C^{\mathfrak{W}, \mathcal{A}}.$$

As an easy consequence of this lemma, there is a close connection between reasoning in a DL $L$ and reasoning in the corresponding ADS. Given a TBox $\mathcal{T}$ and an ABox $\mathcal{A}$ of the DL $L$, we define the corresponding set $S(\mathcal{T}, \mathcal{A})$ of assertions of the corresponding ADL $(\mathcal{L}, \mathcal{M})$ in the obvious way, i.e., each GCI $C \sqsubseteq D$ in $\mathcal{T}$ yields a term assertion $t_C \sqsubseteq t_D$, each role assertion $R(i, j)$ in $\mathcal{A}$ yields an object assertion $R(a_i, a_j)$, and each concept assertion $C(i)$ yields an object assertion $a_i : t_C$.

**Proposition 14.** *The ABox $\mathcal{A}$ is consistent relative to the TBox $\mathcal{T}$ in $L$ iff $S(\mathcal{T}, \mathcal{A})$ is satisfiable in the corresponding ADS.*

We do not treat non-relativized consistency explicitly since it is the special case of relativized consistency where the TBox is empty.

As already mentioned above, our transfer results require the component ADSs to be local. We call a DL $L$ *local* iff the ADS $(\mathcal{L}, \mathcal{M})$ corresponding to $L$ is local. It turns out that not all DLs introduced in Section 2 are local.

**Proposition 15.** *Let $L$ be one of the DLs introduced in Section 2. Then, $L$ is local iff $L$ does not include any of the following constructors: nominals, role complement, universal role.*

**Proof.** We start with the "only if" direction, which is more interesting since it shows *why* ADSs corresponding to DLs with nominals, role complement, or the universal role are not local. We make a case distinction according to which of these constructors $L$ contains.

- **Nominals.** Consider the disjoint union $\mathfrak{W}$ of the ADMs $\mathfrak{W}_1$ and $\mathfrak{W}_2$, and assume that $\mathfrak{W}_1$ and $\mathfrak{W}_2$ correspond to frames of a DL with nominals. By definition of the disjoint union, we know that $\Delta^{\mathfrak{W}_1} \cap \Delta^{\mathfrak{W}_2} = \emptyset$. If $I \in N_O$ is a nominal, then the definition of the disjoint union implies that $f_I^{\mathfrak{W}} = f_I^{\mathfrak{W}_1} \cup f_I^{\mathfrak{W}_2}$. Since nominals are interpreted by singleton sets in $\mathfrak{W}_1$ and $\mathfrak{W}_2$, and since the domains of $\mathfrak{W}_1$ and $\mathfrak{W}_2$ are disjoint, this implies that $f_I^{\mathfrak{W}}$ is a set of cardinality 2. Consequently, $\mathfrak{W}$ cannot correspond to an ADM induced by a frame for a DL with nominals, since such frames interpret nominals by singleton sets.

- **Universal role.** Again, consider the disjoint union $\mathfrak{W}$ of the ADMs $\mathfrak{W}_1$ and $\mathfrak{W}_2$, and assume that $\mathfrak{W}_1$ and $\mathfrak{W}_2$ correspond to frames of a DL with the universal role. Let $U$ denote the universal role, i.e., a role name for which the interpretation is restricted to the binary relation relating each pair of individuals of the domain. By the definition of the disjoint union, we have $U^{\mathfrak{W}} = U^{\mathfrak{W}_1} \cup U^{\mathfrak{W}_2} = \Delta^{\mathfrak{W}_1} \times \Delta^{\mathfrak{W}_1} \cup \Delta^{\mathfrak{W}_2} \times \Delta^{\mathfrak{W}_2} \neq \Delta^{\mathfrak{W}} \times \Delta^{\mathfrak{W}}$. Consequently, $\mathfrak{W}$ cannot correspond to an ADM induced by a frame for a DL with universal role, since such a frame would interpret $U$ by $\Delta^{\mathfrak{W}} \times \Delta^{\mathfrak{W}}$.

- **Role complement.** Again, consider the disjoint union $\mathfrak{W}$ of the ADMs $\mathfrak{W}_1$ and $\mathfrak{W}_2$, and assume that $\mathfrak{W}_1$ and $\mathfrak{W}_2$ correspond to frames of a DL with role negation. For an arbitrary role name $R$, we have $\overline{R}^{\mathfrak{W}} = \overline{R}^{\mathfrak{W}_1} \cup \overline{R}^{\mathfrak{W}_2} = (\Delta^{\mathfrak{W}_1} \times \Delta^{\mathfrak{W}_1} \setminus R^{\mathfrak{W}_1}) \cup (\Delta^{\mathfrak{W}_2} \times \Delta^{\mathfrak{W}_2} \setminus R^{\mathfrak{W}_2}) \neq (\Delta^{\mathfrak{W}_1} \cup \Delta^{\mathfrak{W}_2}) \setminus (R^{\mathfrak{W}_1} \cup R^{\mathfrak{W}_2}) = \Delta^{\mathfrak{W}} \setminus R^{\mathfrak{W}}$.

It remains to prove the "if" direction. Assume that $L$ is one of the DLs introduced in Section 2 that does not allow for nominals, role complements, and the universal role. Let $(\mathfrak{F}_p)_{p \in P}$ be a family of $L$-frames $\mathfrak{F}_p = (\Delta^{\mathfrak{F}_p}, \cdot^{\mathfrak{F}_p})$ and let $\mathfrak{W}_p = \langle W_p, \mathcal{F}^{\mathfrak{W}_p}, \mathcal{R}^{\mathfrak{W}_p} \rangle$ be the ADMs corresponding to them. By definition, $\Delta^{\mathfrak{F}_p} = W_p$ for all $p \in P$. Assume that the domains $(W_p)_{p \in P}$ are pairwise disjoint. We must show that the disjoint union of $(\mathfrak{W}_p)_{p \in P}$ also corresponds to an $L$-frame. To this purpose, we define the frame $\mathfrak{F} = (\Delta^{\mathfrak{F}}, \cdot^{\mathfrak{F}})$ as follows:

- $\Delta^{\mathfrak{F}} := \bigcup_{p \in P} \Delta^{\mathfrak{F}_p}$ and

- $R^{\mathfrak{F}} := \bigcup_{p \in P} R^{\mathfrak{F}_p}$ for all $R \in N_R$.

Let $\mathfrak{W} = \langle W, \mathcal{F}^{\mathfrak{W}}, \mathcal{R}^{\mathfrak{W}} \rangle \in \mathcal{M}$ be the ADM corresponding to $\mathfrak{F}$. By Definition 12 (corresponding ADM), we have $W = \bigcup_{p \in P} W_p$ and $R^{\mathfrak{W}} = \bigcup_{p \in P} R^{\mathfrak{W}_p}$ for all $R \in N_R$. By induction on the structure of complex roles, it is easy to show that this also holds for all $R \in \mathcal{R}$, i.e., all complex role descriptions. For example, consider the role description $R_1 \circ R_2$. By induction, we know that $R_1^{\mathfrak{W}} = \bigcup_{p \in P} R_1^{\mathfrak{W}_p}$ and $R_2^{\mathfrak{W}} = \bigcup_{p \in P} R_2^{\mathfrak{W}_p}$. Since the sets $(W_p)_{p \in P}$ are pairwise disjoint,

$$(R_1 \circ R_2)^{\mathfrak{W}} = R_1^{\mathfrak{W}} \circ R_2^{\mathfrak{W}} = \bigcup_{p \in P} R_1^{\mathfrak{W}_p} \circ \bigcup_{p \in P} R_2^{\mathfrak{W}_p} = \bigcup_{p \in P} R_1^{\mathfrak{W}_p} \circ R_2^{\mathfrak{W}_p} = \bigcup_{p \in P} (R_1 \circ R_2)^{\mathfrak{W}_p}.$$

Since $R^{\mathfrak{W}_p} = R^{\mathfrak{F}_p}$ for all $R \in \mathcal{R}$ and $p \in P$, we obtain the following fact:

$(*)$ For all $p \in P$, $a \in \Delta^{\mathfrak{F}_p}$, and role descriptions $R \in \mathcal{R}$, the following holds: $R^{\mathfrak{F}}(a) = R^{\mathfrak{F}_p}(a)$; in particular, $R^{\mathfrak{F}}(a) \subseteq \Delta^{\mathfrak{F}_p}$.

It remains to show that, for all $n \geq 0$, all $X_1, \ldots, X_n \subseteq W$, and all function symbols $f$ of arity $n$, we have

$$f^{\mathfrak{W}}(X_1, \ldots, X_n) = \bigcup_{p \in P} f^{\mathfrak{W}_p}(X_1 \cap W_p, \ldots, X_n \cap W_p).$$

This can be proved by making a case distinction according to the type of $f$. We treat two cases exemplarily.

- $f = f_{u_1 \downarrow u_2}$. Since $W = \bigcup_{p \in P} W_p$ and the sets $W_p$ are pairwise disjoint, $f^{\mathfrak{W}}_{u_1 \downarrow u_2}$ is the disjoint union of the sets $f^{\mathfrak{W}}_{u_1 \downarrow u_2} \cap W_p$ for $p \in P$. It remains to show that $f^{\mathfrak{W}}_{u_1 \downarrow u_2} \cap W_p = f^{\mathfrak{W}_p}_{u_1 \downarrow u_2}$ $(p \in P)$. By definition of $f^{\mathfrak{W}_p}_{u_1 \downarrow u_2}$, we know that $a \in f^{\mathfrak{W}_p}_{u_1 \downarrow u_2}$ iff $a \in \Delta^{\mathfrak{I}_p}$, both $u_1^{\mathfrak{I}_p}(a)$ and $u_2^{\mathfrak{I}_p}(a)$ are defined, and $u_1^{\mathfrak{I}_p}(a) = u_2^{\mathfrak{I}_p}(a)$. By $(*)$, this is the case iff $a \in \Delta^{\mathfrak{I}_p}$, both $u_1^{\mathfrak{I}}(a)$ and $u_2^{\mathfrak{I}}(a)$ are defined and $u_1^{\mathfrak{I}}(a) = u_2^{\mathfrak{I}}(a)$, which is equivalent to $a \in f^{\mathfrak{W}}_{u_1 \downarrow u_2} \cap W_p$.

- $f = f_{\geq n R}$. Since $W = \bigcup_{p \in P} W_p$ and the sets $W_p$ are pairwise disjoint, $f^{\mathfrak{W}}_{\geq n R}(X)$ is the disjoint union of the sets $f^{\mathfrak{W}}_{\geq n R}(X) \cap W_p$ for $p \in P$. It remains to show that $f^{\mathfrak{W}}_{\geq n R}(X) \cap W_p = f^{\mathfrak{W}_p}_{\geq n R}(X \cap W_p)$ $(p \in P)$. By definition of $f^{\mathfrak{W}_p}_{\geq n R}$, we know that $a \in f^{\mathfrak{W}_p}_{\geq n R}(X \cap W_p)$ iff $a \in \Delta^{\mathfrak{I}_p}$ and $|R^{\mathfrak{I}_p}(a) \cap (X \cap W_p)| \geq n$. By $(*)$, this is the case iff $|R^{\mathfrak{I}}(a) \cap (X \cap W_p)| \geq n$ iff $|R^{\mathfrak{I}}(a) \cap X| \geq n$, and hence iff $a \in f^{\mathfrak{W}}_{\geq n R}(X) \cap W_p$. ❑

It should be noted that arguments similar to the ones used in the proof of the "only if" direction show that, in the presence of the universal role or of role negation, function symbols (e.g., $f_{\forall U}$) may also violate the locality condition.

The transfer results for decidability that are developed in this paper only apply to fusions of local ADSs. Hence, the "only if" direction of the proposition implies that our results are not applicable to fusions of ADSs corresponding to DLs that incorporate nominals, role complement, or the universal role.

## 3.3 Correspondence to modal logics

In this paper our concern are fusions of description logics and not modal logics. Nevertheless, it is useful to have a brief look at the relationship between ADSs and modal logic. Standard modal languages can be regarded as ADLs without relation symbols and object variables (just identify propositional formulas with terms). Given such an ADL $\mathcal{L}$, a set $L$ of $\mathcal{L}$-terms is called a *classical modal logic* iff is contains all tautologies of classical propositional logic and is closed under modus ponens, substitutions, and the regularity rule

$$\frac{x_1 \leftrightarrow y_1, \ldots, x_{n_f} \leftrightarrow y_{n_f}}{f(x_1, \ldots, x_{n_f}) \leftrightarrow f(y_1, \ldots, y_{n_f})}$$

for all function symbols $f$ of $\mathcal{L}$. The minimal classical modal logic in the language with one unary function symbol is known as the logic $\mathbf{E}$ (see Chellas, 1980).

Any ADS $(\mathcal{L}, \mathcal{M})$ based on $\mathcal{L}$ determines a classical modal logic $L$ by taking the valid terms, i.e., by defining

$$t \in L \quad \text{iff} \quad t^{\mathfrak{W}, \mathcal{A}} = W \text{ for all } \mathfrak{W} \in \mathcal{M} \text{ and assignments } \mathcal{A} \text{ in } \mathfrak{W}.$$

The logic $\mathbf{E}$ is determined by the ADS with precisely one unary operator whose class of ADMs consists of all models. Chellas formulates this completeness result (Theorem 9.8 in Chellas, 1980) for so-called minimal models (alias neighborhood-frames), which are, however, just a notational variant of abstract description models with one unary operator (Došen, 1988). If the classical modal logic $L$ is determined by an ADS with decidable term satisfiability problem, then $L$ is decidable since $t \in L$ iff $\neg t$ is unsatisfiable.

A classical modal logic $L$ is called *normal* iff it additionally contains

$$f(x_1, \ldots, x_{j-1}, x_j \wedge y_j, x_{j+1}, \ldots, x_{n_f}) \quad \leftrightarrow \quad f(x_1, \ldots, x_{j-1}, x_j, x_{j+1}, \ldots, x_{n_f}) \wedge$$
$$f(x_1, \ldots, x_{j-1}, y_j, x_{j+1}, \ldots, x_{n_f})$$

and

$$f(\top, \bot, \ldots, \bot), \;\; f(\bot, \top, \bot, \ldots, \bot), \;\; \ldots, f(\bot, \ldots, \bot, \top),$$

for all function symbols $f$ and all $j$ with $1 \leq j \leq n_f$ (Jónsson & Tarski, 1951; Jónsson & Tarski, 1952; Goldblatt, 1989). This definition of normal modal logics assumes that the formulas (terms) are built using only necessity (box) operators.[7] We will work here only with necessity operators; the corresponding possibility-operators are definable by putting

$$f^{\diamond}(x_1, \ldots, x_{n_f}) = \neg f(\neg x_1, \ldots, \neg x_{n_f}).$$

The minimal normal modal logic in the language with one unary operator is known as $\mathbf{K}$ (Chellas, 1980).

We call a function $F : W^n \to W$ *normal* iff for all $1 \leq j \leq n$ and $X_1, \ldots, X_n, Y_j \subseteq W$

$$F(X_1, \ldots, X_{j-1}, X_j \cap Y_j, X_{j+1}, \ldots, X_n) \quad = \quad F(X_1, \ldots, X_{j-1}, X_j, X_{j+1}, \ldots, X_n) \cap$$
$$F(X_1, \ldots, X_{j-1}, Y_j, X_{j+1}, \ldots, X_n))$$

and

$$F(W, \emptyset, \ldots, \emptyset) = F(\emptyset, W, \emptyset, \ldots, \emptyset) = \cdots = F(\emptyset, \ldots, \emptyset, W) = W.$$

Note that a unary function $F$ is normal iff $F(W) = W$ and $F(X \cap Y) = F(X) \cap F(Y)$, for any $X, Y \subseteq W$. A function symbol $f$ is called *normal* in an ADS $(\mathcal{L}, \mathcal{M})$ iff the functions $f^{\mathfrak{W}}$ are normal for all $\mathfrak{W} \in \mathcal{M}$.

For any role $R$ of some DL, the function symbol $f_{\forall R}$ is normal in the corresponding ADS. To the contrary, it is readily checked that neither $f_{\geq nR}$ and $f_{\leq nR}$ nor their duals $f^{\diamond}_{\geq nR}$ and $f^{\diamond}_{\leq nR}$ are normal.

Obviously, an ADS $(\mathcal{L}, \mathcal{M})$ determines a normal modal logic iff all function symbols of $\mathcal{L}$ are normal in $(\mathcal{L}, \mathcal{M})$. Completeness of $\mathbf{K}$ with respect to Kripke semantics (Chellas, 1980) implies that the logic $\mathbf{K}$ is determined by the ADS with one unary operator whose class of ADMs consists of all models interpreting this operator by a normal function.

---

7. Note that some authors define normal modal logics using possibility (diamond) operators, in which case the definitions are the duals of what we have introduced and thus at first sight look quite different.

## 4. Fusions of abstract description systems

In this section, we define the fusion of abstract description systems and prove two transfer theorems for decidability, one concerning satisfiability and the other one concerning relativized satisfiability.

**Definition 16.** The *fusion* $S_1 \otimes S_2 = (\mathcal{L}_1 \otimes \mathcal{L}_2, \mathcal{M}_1 \otimes \mathcal{M}_2)$ of two abstract description systems $S_1 = (\mathcal{L}_1, \mathcal{M}_1)$ and $S_2 = (\mathcal{L}_2, \mathcal{M}_2)$ over

- disjoint sets of function symbols $\mathcal{F}$ of $\mathcal{L}_1$ and $\mathcal{G}$ of $\mathcal{L}_2$,

- disjoint sets of relation symbols $\mathcal{R}$ of $\mathcal{L}_1$ and $\mathcal{Q}$ of $\mathcal{L}_2$, and

- the same sets of set and object variables

is defined as follows: $\mathcal{L}_1 \otimes \mathcal{L}_2$ is the ADL based on

- the union $\mathcal{F} \cup \mathcal{G}$ of the function symbols of $\mathcal{L}_1$ and $\mathcal{L}_2$, and

- the union $\mathcal{R} \cup \mathcal{Q}$ of the relation symbols of $\mathcal{L}_1$ and $\mathcal{L}_2$,

and $\mathcal{M}_1 \otimes \mathcal{M}_2$ is defined as

$$\{ \left\langle W, \mathcal{F}^{\mathfrak{W}} \cup \mathcal{G}^{\mathfrak{W}}, \mathcal{R}^{\mathfrak{W}} \cup \mathcal{Q}^{\mathfrak{W}} \right\rangle \mid \left\langle W, \mathcal{F}^{\mathfrak{W}}, \mathcal{R}^{\mathfrak{W}} \right\rangle \in \mathcal{M}_1 \text{ and } \left\langle W, \mathcal{G}^{\mathfrak{W}}, \mathcal{Q}^{\mathfrak{W}} \right\rangle \in \mathcal{M}_2 \}.$$

As an example, consider the ADSs $S_1$ and $S_2$ corresponding to the DLs $\mathcal{ALCF}$ and $\mathcal{ALC}^{+,\circ,\sqcup}$ introduced in Section 2. We concentrate on the function symbols provided by their fusion. In the following, we assume that the set of role names employed by $\mathcal{ALCF}$ and $\mathcal{ALC}^{+,\circ,\sqcup}$ are disjoint.

- The ADS $S_1$ is based on the following function symbols: (i) unary functions symbol $f_{\forall R}$ and $f_{\exists R}$ for every role name $R$ of $\mathcal{ALCF}$, (ii) nullary functions symbols corresponding to the same-as constructor for every pair of chains of functional roles of $\mathcal{ALCF}$.

- The ADS $S_2$ is based on the following function symbols: (iii) unary functions symbol $f_{\forall Q}$ and $f_{\exists Q}$ for every role description $Q$ built from role names of $\mathcal{ALC}^{+,\circ,\sqcup}$ using union, composition, and transitive closure.

Since we assumed that the set of role names employed by $\mathcal{ALCF}$ and $\mathcal{ALC}^{+,\circ,\sqcup}$ are disjoint, these sets of function symbols are also disjoint. The union of these sets provides us both with the symbols for the same-as constructor and with the symbols for value and existential restrictions on role descriptions involving union, composition, and transitive closure. However, the role descriptions contain only role names from $\mathcal{ALC}^{+,\circ,\sqcup}$, and thus none of the functional roles of $\mathcal{ALCF}$ occurs in such descriptions. Thus, the fusion of $\mathcal{ALCF}$ and $\mathcal{ALC}^{+,\circ,\sqcup}$ yields a strict fragment of their union $\mathcal{ALCF}^{+,\circ,\sqcup}$.

### 4.1 Relativized satisfiability

We prove a transfer result for decidability of the relativized satisfiability problem, show that this also yields a corresponding transfer result for the relativized *term* satisfiability problem, and investigate how these transfer results can be extended to ADSs that correspond to DLs providing for the universal role.

### 4.1.1 The transfer result

This section is concerned with establishing the following transfer theorem:

**Theorem 17.** *Let $S_1$ and $S_2$ be local ADSs, and suppose that the relativized satisfiability problems for $S_1$ and $S_2$ are decidable. Then the relativized satisfiability problem for $S_1 \otimes S_2$ is also decidable.*

The idea underlying the proof of this theorem is to translate a given set of assertions $\Gamma$ of $S_1 \otimes S_2$ into a set of assertions $\Gamma_1$ of $S_1$ and a set of assertions $\Gamma_2$ of $S_2$ such that $\Gamma$ is satisfiable in $S_1 \otimes S_2$ iff $\Gamma_1$ is satisfiable in $S_1$ and $\Gamma_2$ is satisfiable in $S_2$. The first (naive) idea for how to obtain the set $\Gamma_i$ ($i = 1, 2$) is to replace in $\Gamma$ alien terms (i.e., subterms starting with function symbols not belonging to $S_i$) by new set variables (the surrogate variables introduced below). With this approach, satisfiability of $\Gamma$ would in fact imply satisfiability of the sets $\Gamma_i$, but the converse would not be true. The difficulty arises when trying to combine the models of $\Gamma_1$ and $\Gamma_2$ into one for $\Gamma$. To ensure that the two models can indeed be combined, the sets $\Gamma_i$ must contain additional assertions that make sure that the surrogate variables in one model and the corresponding alien subterms in the other model are interpreted in a "compatible" way. To be more precise, there are (finitely many) different ways of adding such assertions, and one must try which of them (if any) leads to a satisfiable pair $\Gamma_1$ and $\Gamma_2$.

For the proof of Theorem 17, we fix two local ADSs $S_i = (\mathcal{L}_i, \mathcal{M}_i)$, $i \in \{1, 2\}$, in which $\mathcal{L}_1$ is based on the set of function symbols $\mathcal{F}$ and relation symbols $\mathcal{R}$, and $\mathcal{L}_2$ is based on $\mathcal{G}$ and $\mathcal{Q}$. Let $\mathcal{L} = \mathcal{L}_1 \otimes \mathcal{L}_2$ and $\mathcal{M} = \mathcal{M}_1 \otimes \mathcal{M}_2$.

In what follows, we use the following notation: for a set of assertions $\Gamma$, denote by $term(\Gamma)$ and $obj(\Gamma)$ the set of terms and object names in $\Gamma$, respectively.

We start with explaining how alien subterms in the set $\Gamma$ can be replaced by new set variables. For each $\mathcal{L}$-term $t$ of the form $h(t_1, \ldots, t_n)$, $h \in \mathcal{F} \cup \mathcal{G}$, we reserve a new variable $x_t$, which will be called the *surrogate* of $t$. We assume that the set of surrogate variables is disjoint to the original sets of variables. As sketched above, the idea underlying the introduction of surrogate variables is that the decision procedure for $S_1$ ($S_2$) cannot deal with terms containing function symbols from $\mathcal{G}$ ($\mathcal{F}$). Thus, these "alien" function symbols must be replaced before applying the procedure. To be more precise, we replace the whole alien subterm starting with the alien function symbol by its surrogate. For example, if the unary symbol $f$ belongs to $\mathcal{F}$, and the unary symbol $g$ belongs to $\mathcal{G}$, then $f(g(f(x)))$ is a "mixed" $\mathcal{L}$-term. To obtain a term of $\mathcal{L}_1$, we replace the subterm $g(f(x))$ by its surrogate, which yields $f(x_{g(f(x))})$. Analogously, to obtain a term of $\mathcal{L}_2$, we replace the whole term by its surrogate, which yields $x_{f(g(f(x)))}$. We now define this replacement process more formally.

**Definition 18.** For an $\mathcal{L}$-term $t$ without surrogate variables, denote by $sur_1(t)$ the $\mathcal{L}_1$-term resulting from $t$ when all occurrences of terms $g(t_1, \ldots, t_n)$, $g \in \mathcal{G}$, that are not within the scope of some $g' \in \mathcal{G}$ are replaced by their surrogate variable $x_{g(t_1,\ldots,t_n)}$. For a set $\Theta$ of terms, put $sur_1(\Theta) := \{sur_1(t) \mid t \in \Theta\}$ and define $sur_2(t)$ as well as $sur_2(\Theta)$ accordingly.

Denote by $sub(\Theta)$ the set of subterms of terms in $\Theta$, and by $sub^1(\Theta)$ the variables occurring in $\Theta$ as well as the subterms of alien terms (i.e., terms starting with a symbol from $\mathcal{G}$) in $\Theta$. More formally, we can define

$$sub^1(\Theta) := sub\{t \mid x_t \in var(sur_1(\Theta))\} \cup var(\Theta).$$

Define $sub^2(\Theta)$ accordingly.

For example, let $f \in \mathcal{F}$ be unary and $g \in \mathcal{G}$ be binary. If $t = f(g(x, f(g(x, y))))$, then $sur_1(t) = f(x_{g(x,f(g(x,y)))})$. Note that the restriction "not within the scope of some $g' \in G$" is there to clarify that only the top-most alien subterms are to be replaced. For the term $t$ of this example, we have $sub^1(\{t\}) = \{g(x, f(g(x, y))), f(g(x, y)), g(x, y), x, y\}$.

Note that the Boolean operators occurring in terms are "shared" function symbols in the sense that they are alien to neither $\mathcal{L}_1$ nor $\mathcal{L}_2$. Thus, $sur_1(f(x) \wedge g(x, y)) = f(x) \wedge x_{g(x,y)}$ and $sur_2(f(x) \wedge g(x, y)) = x_{f(x)} \wedge g(x, y)$.

Of course, when replacing whole terms by variables, some information is lost. For example, consider the (inconsistent) assertion $(\exists R_1.((\leq 1 R_2) \sqcap (\geq 2 R_2)))(i)$ and assume that $R_1$ is a role of one component of a fusion, and $R_2$ a role of the other component. Translated into abstract description language syntax, the concept description $\exists R_1.((\leq 1 R_2) \sqcap (\geq 2 R_2))$ yields the term $t := f_{\exists R_1}(f_{(\leq 1 R_2)} \wedge f_{(\geq 2 R_2)})$, where $f_{\exists R_1}$ is a function symbol of $\mathcal{L}_1$ and the other two function symbols belong to $\mathcal{L}_2$. Now, $sur_1(t) = f_{\exists R_1}(x \wedge y)$, where $x$ is the surrogate for $f_{(\leq 1 R_2)}$ and $y$ is the surrogate for $f_{(\geq 2 R_2)}$. If the decision procedure for the first ADS only sees $f_{\exists R_1}(x \wedge y)$, it has no way to know that the conjunction of the alien subterms corresponding to $x$ and $y$ is unsatisfiable. In fact, for this procedure $x$ and $y$ are arbitrary set variables, and thus $x \wedge y$ is satisfiable. To avoid this problem, we introduce so-called consistency set consisting of "types", where a type says for each "relevant" formula whether the formula itself or its negation is supposed to hold. The sets $\Gamma_1$ and $\Gamma_2$ will then contain additional information that basically ensures that their models satisfy the same types. This will allow us to merge these models into one for $\Gamma$.

**Definition 19.** Given a finite set $\Theta$ of $\mathcal{L}$-terms, we define the *consistency set* $\mathcal{C}(\Theta)$ of $\Theta$ as $\mathcal{C}(\Theta) := \{t_c \mid c \subseteq \Theta\}$, where the *type* $t_c$ determined by $c \subseteq \Theta$ is defined as

$$t_c := \bigwedge\{\chi \mid \chi \in c\} \wedge \bigwedge\{\neg\chi \mid \chi \in \Theta \setminus c\}.$$

Given a finite set $\Gamma$ of assertions in $\mathcal{L}$, we define $sub^i(\Gamma) := sub^i(term(\Gamma))$. We abbreviate $\mathcal{C}^i(\Gamma) := \mathcal{C}(sub^i(\Gamma))$, for $i \in \{1, 2\}$.

In the example above, we have

$$sub^1(f_{\exists R_1}(f_{(\leq 1 R_2)} \wedge f_{(\geq 2 R_2)})) = \{f_{(\leq 1 R_2)}, f_{(\geq 2 R_2)}\},$$

and thus $\mathcal{C}^1(\{a_i : f_{\exists R_1}(f_{(\leq 1 R_2)} \wedge f_{(\geq 2 R_2)})\})$ consists of the 4 terms

$$\begin{array}{rcl}
f_{(\leq 1 R_2)} & \wedge & f_{(\geq 2 R_2)}, \\
f_{(\leq 1 R_2)} & \wedge & \neg f_{(\geq 2 R_2)}, \\
\neg f_{(\leq 1 R_2)} & \wedge & f_{(\geq 2 R_2)}, \text{ and} \\
\neg f_{(\leq 1 R_2)} & \wedge & \neg f_{(\geq 2 R_2)}.
\end{array}$$

Given a set of terms $\Theta$, an element $t_c$ of its consistency set $\mathcal{C}(\Theta)$ can indeed be considered as the "type" of an element $e$ of the domain of an ADM w.r.t. $\Theta$. Any such element $e$ belongs to the interpretations of some of the terms in $\Theta$, and to the complements of the interpretations of the other terms. Thus, if $c$ is the set of terms of $\Theta$ to which $e$ belongs, then $e$ also belongs to the interpretation of $t_c$ and it does not belong to the interpretation of any of the other terms in $\mathcal{C}(\Theta)$. In this case we say that $e$ *realizes* the type $t_c$.

We are now ready to formulate the theorem that reduces the relativized satisfiability problem in a fusion of two local ADSs to relativized satisfiability in the component ADSs. A proof of this theorem can be found in the appendix.

**Theorem 20.** *Let $S_i = (\mathcal{L}_i, \mathcal{M}_i)$, $i \in \{1, 2\}$, be two local ADSs in which $\mathcal{L}_1$ is based on the set of function symbols $\mathcal{F}$ and relation symbols $\mathcal{R}$, and $\mathcal{L}_2$ is based on $\mathcal{G}$ and $\mathcal{Q}$, and let $\mathcal{L} = \mathcal{L}_1 \otimes \mathcal{L}_2$ and $\mathcal{M} = \mathcal{M}_1 \otimes \mathcal{M}_2$. If $\Gamma$ is a finite set of assertions from $\mathcal{L}$, then the following are equivalent:*

*1. $\Gamma$ is satisfiable in $(\mathcal{L}, \mathcal{M})$.*

*2. There exist*

    *(a) a set $D \subseteq \mathcal{C}^1(\Gamma)$,*

    *(b) for every term $t \in D$ an object variable $a_t \notin obj(\Gamma)$,*

    *(c) for every $a \in obj(\Gamma)$ a term $t_a \in D$,*

  *such that the union $\Gamma_1$ of the following sets of assertions in $\mathcal{L}_1$ is satisfiable in $(\mathcal{L}_1, \mathcal{M}_1)$:*

    *(d) $\{a_t : sur_1(t) \mid t \in D\} \cup \{\top \sqsubseteq sur_1(\bigvee D)\}$,*

    *(e) $\{a : sur_1(t_a) \mid a \in obj(\Gamma)\}$,*

    *(f) $\{R(a, b) \mid R(a, b) \in \Gamma, R \in \mathcal{R}\}$,*

    *(g) $\{sur_1(t_1) \sqsubseteq sur_1(t_2) \mid t_1 \sqsubseteq t_2 \in \Gamma\} \cup \{a : sur_1(s) \mid (a : s) \in \Gamma\}$;*

  *and the union $\Gamma_2$ of the following sets of assertions in $\mathcal{L}_2$ is satisfiable in $(\mathcal{L}_2, \mathcal{M}_2)$:*

    *(h) $\{a_t : sur_2(t) \mid t \in D\} \cup \{\top \sqsubseteq sur_2(\bigvee D)\}$,*

    *(i) $\{a : sur_2(t_a) \mid a \in obj(\Gamma)\}$,*

    *(j) $\{Q(a, b) \mid Q(a, b) \in \Gamma, Q \in \mathcal{Q}\}$.*

Intuitively, (2a) "guesses" a set $D$ of types (i.e., elements of the consistency set). The idea is that these are exactly the types that are realized in the model of $\Gamma$ (to be constructed when showing $(2 \to 1)$ and given when showing $(1 \to 2)$). Condition (2b) introduces for every type in $D$ a name for an object realizing this type, and (2c) "guesses" for every object variable occurring in $\Gamma$ a type from $D$.

Regarding (2d) and (2h), one should note that the set of assertions $\{a_t : t \mid t \in D\} \cup \{\top \sqsubseteq \bigvee D\}$ states that every type in $D$ is realized (i.e., there is an object in the model having this type) and that every object has one of the types in $D$. The sets of assertions (2d) and (2h) are obtained from this set through surrogation to make it digestible by the decision procedures of the component logics.

The assertions in (2e) and (2i) state (again in surrogated versions) that the object interpreting the variable $a$ has type $t_a$. This ensures that, in the models of $\Gamma_1$ and $\Gamma_2$ (given when showing $(2 \to 1)$), the objects interpreting $a$ have the same type $t_a$ from $D$. Otherwise, these models could not be combined into a common one for $\Gamma$.

The sets (2f) and (2j) are obtained from $\Gamma$ by distributing its relationship assertions between $\Gamma_1$ and $\Gamma_2$, depending on the relation symbol used in the assertion.

The set (2g) contains (in surrogated version) the term assertions of the form $t_1 \sqsubseteq t_2$ and the membership assertions of the form $a : s$ of $\Gamma$.

Condition 2 is asymmetric in two respects. First, it guesses a subset of $\mathcal{C}^1(\Gamma)$ rather than a subset of $\mathcal{C}^2(\Gamma)$. Of course this is arbitrary, we could also have chosen index 2 instead of 1 here. Second, the set $\Gamma_2$ neither contains the assertions $\{ sur_2(t_1) \sqsubseteq sur_2(t_2) \mid t_1 \sqsubseteq t_2 \in \Gamma \}$ nor $\{ a : sur_2(s) \mid (a : s) \in \Gamma \}$. If we added these assertions, the theorem would still be true, but this would unnecessarily increase the amount of work to be done by the combined decision procedure. In fact, since the other assertions in $\Gamma_1$ and $\Gamma_2$ enforce a tight coordination between the models of $\Gamma_1$ and $\Gamma_2$, the fact that the membership assertions and term assertions of $\Gamma$ are satisfied in the models of $\Gamma_1$ implies that they are also satisfied in the models of $\Gamma_2$ (see the appendix for details).

To prove Theorem 17, we must show how Theorem 20 can be used to construct a decision procedure for relativized satisfiability in $S_1 \otimes S_2$ from such decision procedures for the component systems $S_1$ and $S_2$. For a given finite set of assertions $\Gamma$ of $S_1 \otimes S_2$, the set $\mathcal{C}^1(\Gamma)$ is also finite, and thus there are finitely many sets $D$ in (2a) and choices of types for object variables in (2c). Consequently, we can enumerate all of them and check whether one of these choices leads to satisfiable sets $\Gamma_1$ and $\Gamma_2$. By definition of the sets $\Gamma_i$ and of the functions $sur_i$, the assertions in $\Gamma_i$ are indeed assertions of $\mathcal{L}_i$, and thus the satisfiability algorithm for $(\mathcal{L}_i, \mathcal{M}_i)$ can be applied to $\Gamma_i$. This proves Theorem 17.

Regarding the complexity of the obtained decision procedure, the costly step is guessing the right set $D$. Since the cardinality of the set $sub^1(\Gamma)$ is linear in the size of $\Gamma$, the cardinality of $\mathcal{C}^1(\Gamma)$ is exponential in the size of $\Gamma$ (and each element of it has size quadratic in $\Gamma$). Thus, there are doubly exponentially many different subsets to be chosen from. Since the cardinality of the chosen set $D$ may be exponential in the size of $\Gamma$, also the size of $\Gamma_1$ and $\Gamma_2$ may be exponential in $\Gamma$ (because of the big disjunction over $D$). From this, the following corollary follows.

**Corollary 21.** *Let $S_1$ and $S_2$ be local ADSs, and suppose that the relativized satisfiability problems for $S_1$ and $S_2$ are decidable in* ExpTime *(*PSpace*). Then the relativized satisfiability problem for $S_1 \otimes S_2$ is decidable in* 2ExpTime *(*ExpSpace*).*

**Proof.** Assume that $\Gamma$ has size $n$. Then we must consider $2^{2^{p_1(n)}}$ (for some polynomial $p_1$) different sets $D$ in (2a). Each such set has size $2^{p_1(n)}$ and thus we have of $2^{2^{p_2(n)}}$ choices in (2c) (for some polynomial $p_2$). Overall, this still leaves us with doubly exponentially many choices. Now assume that the relativized satisfiability problems for $S_1$ and $S_2$ are decidable in ExpTime. Since each call of these procedures is applied to a set of assertions of exponential size, it may take double exponential time, say $2^{2^{p_3(n)}}$ and $2^{2^{p_4(n)}}$ (for polynomials $p_3$ and $p_4$). Overall, we thus have a time complexity of

$$2^{2^{p_1(n)}} \cdot 2^{2^{p_2(n)}} \cdot \left( 2^{2^{p_3(n)}} + 2^{2^{p_4(n)}} \right),$$

which can clearly be majorized by $2^{2^{p(n)}}$ for an appropriate polynomial $p$. This shows membership in 2ExpTime.

The argument regarding the space complexity is similar. Here one must additionally take into account that doubly exponentially many choices can be enumerated using an exponentially large counter. ❑

### 4.1.2 The relativized term satisfiability problem

The statement of Theorem 17 itself does not imply a transfer result for the relativized term satisfiability problem. The problem is that decidability of the relativized *term* satisfiability problem in $S_1$ and $S_2$ does not necessarily imply decidability of the relativized satisfiability problem in these ADSs, and thus the prerequisite for the theorem to apply is not satisfied. However, if we consider the statement of Theorem 20, then it is easy to see that this theorem also yields a transfer result for the relativized *term* satisfiability problem.

**Corollary 22.** *Let $S_1$ and $S_2$ be local ADSs, and suppose that the relativized term satisfiability problems for $S_1$ and $S_2$ are decidable. Then the relativized* term *satisfiability problem for $S_1 \otimes S_2$ is also decidable.*

**Proof.** Consider the satisfiability criterion in Theorem 20. If we are interested in relativized term satisfiability, then $\Gamma$ is of the form $\{a : t\} \cup \Gamma'$, where $\Gamma'$ is a set of term assertions. In this case, the sets of assertions $\Gamma_1$ and $\Gamma_2$ do not contain object assertions involving relations. Now, assume that $\Gamma_i$ is of the form $\{a_1 : t_1, \ldots, a_n : t_n\} \cup \Gamma'_i$, where $\Gamma'_i$ is a set of term assertions. Since two assertions of the form $b : s_1, b : s_2$ are equivalent to one assertion $b : s_1 \wedge s_2$, we may assume that the $a_i$ are distinct from each other. Since $S_i$ is local, it is easy to see that the following are equivalent:

1. $\{a_1 : t_1, \ldots, a_n : t_n\} \cup \Gamma'_i$ is satisfiable in $S_i$.

2. $\{a_j : t_j\} \cup \Gamma'_i$ is satisfiable in $S_i$ for all $j = 1, \ldots, n$.

Since $(1 \rightarrow 2)$ is trivial, it is enough to show $(2 \rightarrow 1)$. Given models $\mathfrak{W}_j \in \mathcal{M}_i$ of $\{a_j : t_j\} \cup \Gamma'_i$ $(j = 1, \ldots, n)$, their disjoint union also belongs to $\mathcal{M}_i$, and it is clearly a model of $\{a_1 : t_1, \ldots, a_n : t_n\} \cup \Gamma'_i$.

The second condition can now be checked by applying the term satisfiability test in $S_i$ $n$ times. ❑

### 4.1.3 Dealing with the universal role

As stated above (Proposition 15), ADSs corresponding to DLs with the universal role are not local, and thus Theorem 17 cannot be applied directly. Nevertheless, in some cases this theorem can also be used to obtain a decidability result for fusions of DLs with the universal role, provided that both of them provide for a universal role. (We will comment on the usefulness of this approach in more detail in Section 5.4).

**Definition 23.** Given an ADS $S = (\mathcal{L}, \mathcal{M})$, we denote by $S^U$ the ADS obtained from $S$ by

1. extending $\mathcal{L}$ with two function symbols $f_{\exists U_S}$ and $f_{\forall U_S}$, and

2. extending every ADM $\mathfrak{W} = \langle W, \mathcal{F}^{\mathfrak{W}}, \mathcal{R}^{\mathfrak{W}} \rangle \in \mathcal{M}$ with the unary functions $f_{\exists U_S}^{\mathfrak{W}}$ and $f_{\forall U_S}^{\mathfrak{W}}$, where

- $f_{\exists U_S}^{\mathfrak{W}}(X) = \emptyset$ if $X = \emptyset$, and $f_{\exists U_S}^{\mathfrak{W}}(X) = W$ otherwise;

- $f_{\forall U_S}^{\mathfrak{W}}(X) = W$ if $X = W$, and $f_{\forall U_S}^{\mathfrak{W}}(X) = \emptyset$ otherwise.

For ADSs $S$ corresponding to a DL $L$, the ADS $S^U$ corresponds to the extension of $L$ with the universal role, where the universal role can only be used within value and existential restrictions.[8] There is a close connection between the relativized satisfiability problem in $S$ and the satisfiability problem in $S^U$.

**Proposition 24.** *If $S$ is a local ADS, then the following conditions are equivalent:*

1. *the relativized (term) satisfiability problem for $S$ is decidable,*

2. *the (term) satisfiability problem for $S^U$ is decidable,*

3. *the relativized (term) satisfiability problem for $S^U$ is decidable.*

**Proof.** We restrict our attention to the term satisfiability problem since the equivalences for the satisfiability problem can be proved similarly.

The implication $(3 \to 2)$ is trivial, and $(2 \to 1)$ is easy to show. In fact, $t$ is satisfiable in $S$ relative to the term assertions $\{s_1 \sqsubseteq t_1, \ldots, s_n \sqsubseteq t_n\}$ iff $t \wedge f_{\forall U_S}.((\neg t_1 \vee s_1) \wedge \ldots \wedge (\neg t_n \vee s_n))$ is satisfiable in $S^U$.

To show $(1 \to 3)$, we assume that the relativized term satisfiability problem for $S$ is decidable. Let $S = (\mathcal{L}, \mathcal{M})$ and $S^U = (\mathcal{L}^U, \mathcal{M}^U)$. In the following, we use $f_U$ as an abbreviation for $f_{\forall U_S}$. Since we can replace equivalently in any term the function symbol $f_{\exists U_S}$ by $\neg f_U \neg$, we may assume without loss of generality that $f_{\exists U_S}$ does not occur in terms of $\mathcal{L}^U$.

Suppose a set $\Sigma = \{a : s\} \cup \Gamma$ from $\mathcal{L}^U$ is given, where $\Gamma$ is a set of term assertions. We want to decide whether $\Sigma$ is satisfiable in some model $\mathfrak{W} \in \mathcal{M}^U$. To this purpose, we transform $\Sigma$ into a set of assertions not containing $f_U$. The idea underlying this transformation is that, given a model $\mathfrak{W} \in \mathcal{M}^U$, we have $f_U(t)^{\mathfrak{W}} \in \{W, \emptyset\}$, depending on whether $t^{\mathfrak{W}} = W$ or not. Consequently, if we replace $f_U(t)$ accordingly by $\top$ or $\bot$, the evaluation of this term in $\mathfrak{W}$ does not change. However, in the satisfiability test we do not have the model $\mathfrak{W}$ (we are trying to decide whether one exists), and thus we must *guess* the right replacement.

A term $t$ from $\mathcal{L}^U$ is called a *U-term* iff it starts with $f_U$. The set of $U$-terms that occur (possibly as subterms) in $\Sigma$ is denoted by $\Sigma^U$. Set, inductively, for any function

---

8. Note that it is not necessary to add the universal role $U$ to the set of relation symbols since an assertion of the form $U(a, b)$ is trivially true. However, the use of the universal role within (qualified) number restrictions is not covered by this extension.

$\sigma : \Sigma^U \to \{\bot, \top\}$ and all subterms of terms in $\Sigma$:

$$
\begin{aligned}
x^\sigma &:= x, \\
(t_1 \wedge t_2)^\sigma &:= t_1^\sigma \wedge t_2^\sigma, \\
(t_1 \vee t_2)^\sigma &:= t_1^\sigma \vee t_2^\sigma, \\
(\neg t)^\sigma &:= \neg t^\sigma, \\
(f(t_1, \ldots, t_n))^\sigma &:= f(t_1^\sigma, \ldots, t_n^\sigma) \text{ for } f \neq f_U \text{ of arity } n, \\
(f_U(t))^\sigma &:= \sigma(f_U(t)).
\end{aligned}
$$

Thus, $t^\sigma$ is obtained from $t$ by replacing all occurrences of $U$-terms in $t$ by their image under $\sigma$, i.e., by $\bot$ or $\top$. Define, for any such function $\sigma$,

$$
\begin{aligned}
\Sigma^\sigma := \ & \{t_1^\sigma \sqsubseteq t_2^\sigma \mid t_1 \sqsubseteq t_2 \in \Gamma\} \cup \{a : s^\sigma\} \cup \\
& \{\top \sqsubseteq t^\sigma \mid f_U(t) \in \Sigma^U \text{ and } \sigma(f_U(t)) = \top\} \cup \\
& \{a_t : \neg t^\sigma \mid f_U(t) \in \Sigma^U \text{ and } \sigma(f_U(t)) = \bot\},
\end{aligned}
$$

where the $a_t$ are mutually distinct new object variables. Note that $\Sigma^\sigma$ does not contain the function symbol $f_U$, and thus it can be viewed as a set of assertions of $S$. In addition, though it contains more than one membership assertion, it does not contain assertions involving relation symbols. Consequently, the satisfiability of $\Sigma^\sigma$ in $S$ can be checked using the term satisfiability test for $S$ (see the proof of Corollary 22 above). Decidability of the relativized term satisfiability problem for $S^U$ then follows from the following claim:

**Claim.** $\Sigma$ is satisfiable in a member of $\mathcal{M}^U$ iff there exists a mapping $\sigma : \Sigma^U \to \{\bot, \top\}$ such that $\Sigma^\sigma$ is satisfiable in a member of $\mathcal{M}$.

To prove this claim, first suppose that $\Sigma$ is satisfied under an assignment $\mathcal{A}$ in a member $\mathfrak{W} = \langle W, \mathcal{F}^{\mathfrak{W}} \cup \{f_U^{\mathfrak{W}}\}, \mathcal{R}^{\mathfrak{W}} \rangle$ of $\mathcal{M}^U$. Define $\sigma$ by setting $\sigma(f_U(t)) = \top$ if $(f_U(t))^{\mathfrak{W},\mathcal{A}} = W$, and $\sigma(f_U(t)) = \bot$ otherwise. Obviously, this implies that $\Sigma^\sigma$ is satisfied under the assignment $\mathcal{A}$ in $\langle W, \mathcal{F}^{\mathfrak{W}}, \mathcal{R}^{\mathfrak{W}} \rangle$, which is a member of $\mathcal{M}$.

Conversely, suppose $\Sigma^\sigma$ is satisfiable for some mapping $\sigma$. Take a member $\mathfrak{W} = \langle W, \mathcal{F}^{\mathfrak{W}}, \mathcal{R}^{\mathfrak{W}} \rangle$ of $\mathcal{M}$ and an assignment $\mathcal{A}$ such that $\langle \mathfrak{W}, \mathcal{A} \rangle \models \Sigma^\sigma$. Set $\mathfrak{W}' := \langle W, \mathcal{F}^{\mathfrak{W}} \cup \{f_U^{\mathfrak{W}}\}, \mathcal{R}^{\mathfrak{W}} \rangle$, and prove, by induction, for all terms $t$ that occur in $\Sigma$:

$$(*) \qquad t^{\mathfrak{W}',\mathcal{A}} = (t^\sigma)^{\mathfrak{W},\mathcal{A}}.$$

The only critical case is the one where $t = f_U(s)$. First, assume that $\sigma(f_U(s)) = (f_U(s))^\sigma = \top$. Then $\Sigma^\sigma$ contains $\top \sqsubseteq s^\sigma$, and thus $W = (s^\sigma)^{\mathfrak{W},\mathcal{A}} = s^{\mathfrak{W},\mathcal{A}}$, where the second identity holds by induction. However, $s^{\mathfrak{W}',\mathcal{A}} = W$ implies $(f_U(s))^{\mathfrak{W}',\mathcal{A}} = W = \top^{\mathfrak{W},\mathcal{A}}$. The case where $\sigma(f_U(s)) = (f_U(s))^\sigma = \bot$ can be treated similarly. Here the term assertion $a_s : \neg s^\sigma$ ensures that $s^\sigma$ (and thus by induction $s$) is not interpreted as the whole domain. Consequently, applying $f_U$ to it yields the empty set.

Since $\langle \mathfrak{W}, \mathcal{A} \rangle \models \Sigma^\sigma$, the identity $(*)$ implies that $\langle \mathfrak{W}', \mathcal{A} \rangle \models \Sigma$. This completes the proof of the claim, and thus also of the proposition. ❏

For normal modal logics, the result stated in this proposition was already shown by Goranko and Passy (1992). The proof technique used there can, however, not be transfered to our more general situation since it strongly depends on the normality of the modal operators.

Using Proposition 24, we obtain the following corollary to our first transfer theorem.

**Corollary 25.** *Let $S_1$, $S_2$ be local ADSs and assume that, for $i \in \{1, 2\}$, the relativized (term) satisfiability problem for $S_i$ is decidable. Then the relativized (term) satisfiability problem for $S_1^U \otimes S_2^U$ is decidable.*

**Proof.** We know by Theorem 17 (Corollary 22) that the relativized (term) satisfiability problem for $S_1 \otimes S_2$ is decidable. Hence, Proposition 24 yields that the relativized (term) satisfiability problem for $(S_1 \otimes S_2)^U$ is decidable. But $S_1^U \otimes S_2^U$ is just a notational variant of $(S_1 \otimes S_2)^U$: the function symbols $f_{\exists U_{S_1}}$ and $f_{\exists U_{S_2}}$ can be replaced by $f_{\exists U_{S_1 \otimes S_2}}$ (and analogously for $f_{\forall U_{S_1 \otimes S_2}}$) since all three have identical semantics. ❑

## 4.2 Satisfiability

Note that Theorem 17 does not yield a transfer result for the unrelativized satisfiability problem. Of course, if the relativized satisfiability problems for $S_1$ and $S_2$ are decidable, then the theorem implies that the satisfiability problem for $S_1 \otimes S_2$ is also decidable (since it is a special case of the relativized satisfiability problem). However, to be able to apply the theorem to obtain decidability of the satisfiability problem in the fusion, the component ADSs must satisfy the stronger requirement that the relativized satisfiability problem is decidable. Indeed, the set $\Gamma_i$ in Theorem 20 contains a term assertion (namely $\top \sqsubseteq sur_i(\bigvee D)$) even if $\Gamma$ does not contain any term assertions.

There are cases where the relativized satisfiability problem is undecidable whereas the satisfiability problem is still decidable. For example, Theorem 17 cannot be applied for the fusion of $\mathcal{ALCF}$ and $\mathcal{ALC}^{+,\circ,\sqcup}$ since the relativized satisfiability problem for $\mathcal{ALCF}$ is already undecidable (Baader et al., 1993). However, the satisfiability problem is decidable for both DLs.

### 4.2.1 Covering normal terms

Before we can formulate a transfer result for the satisfiability problem, we need to introduce an additional notion, which generalizes the notion of a normal modal logic.

**Definition 26 (Covering normal terms).** Let $(\mathcal{L}, \mathcal{M})$ be an ADS and $f$ be a function symbol of $\mathcal{L}$ of arity $n$. The term $t_f(x)$ (with one variable $x$) is a *covering normal term* for $f$ iff the following holds for all $\mathfrak{W} \in \mathcal{M}$:

- $t_f^{\mathfrak{W}}(W) = W$

- for all $X, Y \subseteq W$, $t_f^{\mathfrak{W}}(X \cap Y) = t_f^{\mathfrak{W}}(X) \cap t_f^{\mathfrak{W}}(Y)$, and

- for all $X, X_1, \ldots, Y_n \subseteq W$: $X \cap X_i = X \cap Y_i$ for $1 \le i \le n$ implies

$$t_f^{\mathfrak{W}}(X) \cap f^{\mathfrak{W}}(X_1, \ldots, X_n) = t_f^{\mathfrak{W}}(X) \cap f^{\mathfrak{W}}(Y_1, \ldots, Y_n).$$

An ADS $(\mathcal{L}, \mathcal{M})$ is said to *have covering normal terms* iff one can effectively determine a covering normal term $t_f$ for every function symbol $f$ of $\mathcal{L}$.

Intuitively, the first two conditions state that the covering normal term behaves like a value restriction (or box operator). Consider the term $f_{\forall R}(x)$, where $f_{\forall R}$ is the function symbol corresponding to the value restriction constructor for the role $R$. Then $f_{\forall R}(x)$ obviously satisfies the first two requirements for covering normal terms. Note that the second condition implies that the function induced by $t_f$ is monotonic, i.e., $X \subseteq Y$ implies $t_f^{\mathfrak{W}}(X) \subseteq t_f^{\mathfrak{W}}(Y)$. The third condition specifies the connection between the covering normal term and the function symbol it covers. With respect to elements of $t_f^{\mathfrak{W}}(X)$, the values of the functions $f^{\mathfrak{W}}(X_1, \ldots, X_n)$ and $f^{\mathfrak{W}}(Y_1, \ldots, Y_n)$ agree provided that their arguments agree on $X$. It is easy to see that $f_{\forall R}(x)$ is a covering normal term for the function symbols corresponding to the value, existential, and (qualified) number restrictions on the role $R$ (see Proposition 35 below).

Given covering normal terms $t_f$ for the function symbols $f$ of a finite set of function symbols $E$, one can construct a term $t_E$ that is a covering normal term for all the elements of $E$.

**Lemma 27.** *Suppose the ADS $(\mathcal{L}, \mathcal{M})$ has covering normal terms and $\mathcal{L}$ is based on a set of function symbols $F$. Denote by $t_f$ the covering normal term for the function symbol $f$, for all $f \in F$. Then, for every finite set $E \subseteq F$ of function symbols, the term*

$$t_E(x) := \bigwedge_{f \in E} t_f(x)$$

*is a covering normal term for all $f \in E$.*

### 4.2.2 Correspondence to normal modal logics

The following result shows that any ADS in which every function symbol is normal has covering normal terms. Hence, the notion of covering normal terms generalizes the notion of normality in modal logics.

**Proposition 28.** *Let $(\mathcal{L}, \mathcal{M})$ be an ADS, and assume that $f$ is a normal function symbol in $(\mathcal{L}, \mathcal{M})$. Then*

$$t_f(x) := f(x, \bot, \ldots, \bot) \wedge f(\bot, x, \ldots, \bot) \wedge \cdots \wedge f(\bot, \ldots, \bot, x)$$

*is a covering normal term for $f$. In particular, if $f$ is nullary (unary), then $t_f(x) = \top$ ($t_f(x) = f(x)$) is a covering normal term for $f$.*

**Proof.** The first two conditions in the definition of covering normal terms immediately follow from the definition of normal function symbols. Thus, we concentrate on the third condition. Assume, for simplicity, that $f$ is binary. Suppose $\mathfrak{W} \in \mathcal{M}$ and $X, X_1, X_2, Y_1, Y_2 \subseteq W$ with $X \cap X_i = X \cap Y_i$ for $i = 1, 2$, and set $F := f^{\mathfrak{W}}$. Then $F(X \cap X_1, X \cap X_2) = F(X \cap Y_1, X \cap Y_2)$. Since $F$ is normal, we know that

$$
\begin{aligned}
F(X \cap X_1, X \cap X_2) &= F(X, X) \cap F(X, X_2) \cap F(X_1, X) \cap F(X_1, X_2), \\
F(X \cap Y_1, X \cap Y_2) &= F(X, X) \cap F(X, Y_2) \cap F(Y_1, X) \cap F(Y_1, Y_2),
\end{aligned}
$$

and thus

$$F(X, X) \cap F(X, X_2) \cap F(X_1, X) \cap F(X_1, X_2) =$$
$$F(X, X) \cap F(X, Y_2) \cap F(Y_1, X) \cap F(Y_1, Y_2).$$

Since, by normality of $F$,

$$F(X, X) \cap F(X, X_2) \cap F(X_1, X) \quad \supseteq \quad t_f^{\mathfrak{M}}(X),$$
$$F(X, X) \cap F(X, Y_2) \cap F(Y_1, X) \quad \supseteq \quad t_f^{\mathfrak{M}}(X),$$

this implies $t_f^{\mathfrak{M}}(X) \cap F(X_1, X_2) = t_f^{\mathfrak{M}}(X) \cap F(Y_1, Y_2)$. ❑

### 4.2.3 THE TRANSFER RESULT

Using covering normal terms, we can now formulate the second transfer theorem, which is concerned with the transfer of decidability of (non-relativized) satisfiability.

**Theorem 29.** *Let $S_1$ and $S_2$ be local ADSs having covering normal terms, and suppose that the satisfiability problems for $S_1$ and $S_2$ are decidable. Then the satisfiability problem for $S_1 \otimes S_2$ is also decidable.*

As in the proof of Theorem 17, we fix two local ADSs $S_i = (\mathcal{L}_i, \mathcal{M}_i)$, $i \in \{1, 2\}$, in which $\mathcal{L}_1$ is based on the set of function symbols $\mathcal{F}$ and relation symbols $\mathcal{R}$, and $\mathcal{L}_2$ is based on $\mathcal{G}$ and $\mathcal{Q}$. Let $\mathcal{L} = \mathcal{L}_1 \otimes \mathcal{L}_2$ and $\mathcal{M} = \mathcal{M}_1 \otimes \mathcal{M}_2$.

The proof of Theorem 29 follows the same general ideas as the proof of Theorem 17. There are, however, notable differences in the way satisfiability in $S_1 \otimes S_2$ is reduced to satisfiability in $S_1$ and $S_2$. In Theorem 20 we had to "guess" a set $D$ of types, and then based on this set and some additional guesses, a pair of satisfiability problems $\Gamma_1$ and $\Gamma_2$ in $S_1$ and $S_2$, respectively, was generated. In the proof of Theorem 29, we do not need to guess $D$. Instead, we can compute the right set. However, this computation requires us to solve additional satisfiability problems in the fusion $S_1 \otimes S_2$. Nevertheless, this yields a reduction since the alternation depth (i.e., number of alternations between function symbols of $S_1$ and $S_2$) decreases when going from the input set $\Gamma$ to these additional mixed satisfiability problems.

Before we can describe this reduction in more detail, we must introduce some new notation. In the case of relativized satisfiability, term assertions of the form $\top \sqsubseteq sur_i(\bigvee D)$ were used to assert that all elements of the domain belong to $sur_i(\bigvee D)$. Now, we use covering normal terms to "propagate" $sur_i(\bigvee D)$ into terms up to a certain depth. For a set of function symbols $E$, define the $E$-depth $d_E(t)$ of a term $t$ inductively:

$$
\begin{aligned}
d_E(x_i) &= 0 \\
d_E(\neg t) &= d_E(t) \\
d_E(t_1 \vee t_2) = d_E(t_1 \wedge t_2) &= max\{d_E(t_1), d_E(t_2)\} \\
d_E(f(t_1, \ldots, t_n)) &= max\{d_E(t_1), \ldots, d_E(t_n)\} + 1 \ \text{ if } \ f \in E \\
d_E(f(t_1, \ldots, t_n)) &= max\{d_E(t_1), \ldots, d_E(t_n)\} \ \text{ if } \ f \notin E
\end{aligned}
$$

If $\Gamma$ is a finite set of assertions, then

$$d_E(\Gamma) := max\{d_E(t) \mid t \in term(\Gamma)\}.$$

Put, for a term $t(x)$ with one variable $x$, $t^0(x) := x$, $t^{m+1}(x) := t(t^m(x))$, $t^{\leq 0}(x) := x$, and $t^{\leq m+1}(x) := t^{m+1}(x) \wedge t^{\leq m}(x)$.

We are now in the position to formulate the result that reduces satisfiability in the fusion of two local ADSs with covering normal terms to satisfiability in the component ADSs.

**Theorem 30.** *Let $S_i = (\mathcal{L}_i, \mathcal{M}_i)$, $i \in \{1, 2\}$, be two local ADSs having covering normal terms in which $\mathcal{L}_1$ is based on the set of function symbols $\mathcal{F}$ and relation symbols $\mathcal{R}$, and $\mathcal{L}_2$ is based on $\mathcal{G}$ and $\mathcal{Q}$, and let $\mathcal{L} = \mathcal{L}_1 \otimes \mathcal{L}_2$ and $\mathcal{M} = \mathcal{M}_1 \otimes \mathcal{M}_2$. Let $\Gamma$ be a finite set of object assertions from $\mathcal{L}$. Put $m := d_{\mathcal{F}}(\Gamma)$, $r := d_{\mathcal{G}}(\Gamma)$, and let $c(x)$ $(d(x))$ be a covering normal term for all function symbols in $\Gamma$ that are in $\mathcal{F}$ $(\mathcal{G})$.*

*For $i \in \{1, 2\}$, denote by $\Sigma_i$ the set of all $s \in \mathcal{C}^i(\Gamma)$ such that the term $s$ is satisfiable in $(\mathcal{L}, \mathcal{M})$. Then the following three conditions are equivalent:*

1. *$\Gamma$ is satisfiable in $(\mathcal{L}, \mathcal{M})$.*

2. *There exist*

   - *for every $t \in \Sigma_1$ an object variable $a_t \notin obj(\Gamma)$*
   - *for every $a \in obj(\Gamma)$ a term $t_a \in \Sigma_1$*

   *such that the union $\Gamma_1$ of the following sets of object assertions is satisfiable in $(\mathcal{L}_1, \mathcal{M}_1)$:*

   - *$\{a_t : sur_1(t \wedge c^{\leq m}(sur_1(\bigvee \Sigma_1))) \mid t \in \Sigma_1\}$,*
   - *$\{a : sur_1(t_a \wedge c^{\leq m}(sur_1(\bigvee \Sigma_1))) \mid a \in obj(\Gamma)\}$,*
   - *$\{R(a, b) \mid R(a, b) \in \Gamma, R \in \mathcal{R}\}$,*
   - *$\{a : sur_1(s) \mid (a : s) \in \Gamma\}$;*

   *and the union $\Gamma_2$ of the following sets of object assertions is satisfiable in $(\mathcal{L}_2, \mathcal{M}_2)$:*

   - *$\{a_t : sur_2(t \wedge d^{\leq r}(sur_2(\bigvee \Sigma_1))) \mid t \in \Sigma_1\}$,*
   - *$\{a : sur_2(t_a \wedge d^{\leq r}(sur_2(\bigvee \Sigma_1))) \mid a \in obj(\Gamma)\}$,*
   - *$\{Q(a, b) \mid Q(a, b) \in \Gamma, Q \in \mathcal{Q}\}$.*

3. *The same condition as in (2) above, with $\Sigma_1$ replaced by $\Sigma_2$.*

The sets $\Gamma_i$ in the above theorem are very similar to the ones in Theorem 20. The main difference is that the term assertion $\top \sqsubseteq sur_i(\bigvee D)$ is no longer there. Instead, the disjunction $sur_i(\bigvee \Sigma_1)$ is directly "inserted" into the terms using the covering normals terms. As already mentioned above, another difference is that the set $D$, which had to be guessed in Theorem 20, is replaced by the set $\Sigma_1$ in (2) and $\Sigma_2$ in (3). Actually, guessing the set $D$ is no longer possible in this case. In the proof of Theorem 30 we need to know that $\top \sqsubseteq sur_i(\bigvee D)$ is satisfiable in $S_i$ (i.e., holds in at least one model in $\mathcal{M}_i$). But we have no way to check this effectively since we do not have an algorithm for relativized satisfiability in $S_i$. Taking the set $\Sigma_i$ ensures that this property is satisfied (see the proof in the appendix for details).

By definition, $\Sigma_i$ is the set of all $s \in \mathcal{C}^i(\Gamma)$ such that the term $s$ is satisfiable in $(\mathcal{L}, \mathcal{M})$. Recall that the term $s$ is satisfiable iff $\{a : s\}$ is satisfiable in $(\mathcal{L}, \mathcal{M})$ for an arbitrary object variable $a$. Since the elements of $\mathcal{C}^i(\Gamma)$ are still mixed terms (i.e., terms of the fusion), computing the set $\Sigma_i$ actually needs a recursive call to the decision procedure for satisfiability in $(\mathcal{L}, \mathcal{M})$. This recursion is well-founded since the alternation depth decreases.

**Definition 31.** For a term $s$ of $\mathcal{L}$, denote by $a_1(s)$ and $a_2(s)$ the 1-alternation and the 2-alternation depth of $s$, respectively. That is to say, $a_1(s)$ is the length of the longest sequence of the form $(g_1, f_2, g_3, \dots)$ such that

$$g_1(\dots(f_2\dots(g_3\dots)))$$

with $g_j \in \mathcal{G}$ and $f_j \in \mathcal{F}$ appears in $s$. The 2-alternation depth $a_2(s)$ is defined by exchanging the roles of $\mathcal{F}$ and $\mathcal{G}$. Put $a(s) := a_1(s) + a_2(s)$, and call this the *alternation depth*. For a finite set $\Theta$ of terms, $a(\Theta)$ is the maximum of all $a(s)$ with $s \in \Theta$.

Thus, $a_1(s)$ counts the maximal number of changes between symbols from the first and the second ADS, starting with the first symbol from $S_2$ (i.e., the first symbol from $S_2$ counts as a change, even if it does not occur inside the scope of a symbol from $S_2$). The 2-alternation depth is defined accordingly. The alternation depth sums up the 1- and the 2-alternation depth.

**Lemma 32.** *If $a(term(\Gamma)) > 0$, then $a(\mathcal{C}^1(\Gamma)) < a(term(\Gamma))$ or $a(\mathcal{C}^2(\Gamma)) < a(term(\Gamma))$.*

**Proof.** We show that, if $a(term(\Gamma)) > 0$, then we have $a(sub^1(\Gamma)) < a(term(\Gamma))$ or $a(sub^2(\Gamma)) < a(term(\Gamma))$, which, by definition of $\mathcal{C}^i$, clearly implies the lemma. First note that, by definition of $sub^i$, we have

$$a_i(sub^j(\Gamma)) \leq a_i(term(\Gamma)) \text{ for all } i, j. \tag{$*$}$$

We now make a case distinction as follows:

1. $a_1(term(\Gamma)) \geq a_2(term(\Gamma))$. We want to show that $a_1(sub^2(\Gamma)) < a_1(term(\Gamma))$, since, by $(*)$, this implies $a(sub^2(\Gamma)) < a(term(\Gamma))$. Assume to the contrary that $a_1(sub^2(\Gamma)) \geq a_1(term(\Gamma))$. Then $(*)$ implies $a_1(sub^2(\Gamma)) = a_1(term(\Gamma))$. Hence, there exists a term $s \in sub^2(\Gamma)$ and a sequence $(g_1, f_2, g_3, \dots)$ of function symbols $g_i \in \mathcal{G}, f_i \in \mathcal{F}$ of length $a_1(term(\Gamma))$ such that $g_1(\dots(f_2\dots(g_3\dots)))$ occurs in $s$. By definition of $sub^2$, this implies the existence of a term $t \in term(\Gamma)$ and a function symbol $f \in \mathcal{F}$ such that $f(\dots g_1(\dots(f_2\dots(g_3\dots))))$ occurs in $t$. Since the length of $(g_1, f_2, g_3, \dots)$ is $a_1(term(\Gamma))$, this obviously yields $a_2(term(\Gamma)) > a_1(term(\Gamma))$ which is a contradiction.

2. $a_1(term(\Gamma)) \leq a_2(term(\Gamma))$. Similar to the previous case: just exchange the roles of $a_1$ and $a_2$, $\mathcal{F}$ and $\mathcal{G}$, and $sub^1$ and $sub^2$.

$\qquad\qquad\qquad\qquad\qquad\qquad\qquad\qquad\qquad\qquad\qquad\qquad\qquad\qquad\qquad\qquad\qquad\qquad\quad$ ❏

To prove Theorem 29, we must show how Theorem 30 can be used to construct a decision procedure for satisfiability in $S_1 \otimes S_2$ from such decision procedures for the component systems $S_1$ and $S_2$. Let us first consider the problem of computing the sets $\Sigma_1$ and $\Sigma_2$. If $a((term(\Gamma)) = 0$, then $\Gamma$ consists of Boolean combinations of set variables. In this case, $\mathcal{C}^i(\Gamma)$ consists of set variables, and $\Sigma_i, i = 1, 2$, can be computed using Boolean reasoning. If $a(term(\Gamma)) > 0$, then Lemma 32 states that there is an $i \in \{1, 2\}$ such that $a(\mathcal{C}^i(\Gamma)) < a(term(\Gamma))$. By induction we can thus assume that $\Sigma_i$ can effectively be computed. Consequently, it remains to check Condition $(i+1)$ of Theorem 30 for $i \in \{1, 2\}$. Since $\Sigma_i$ is finite, we can guess for every object variable $a$ occurring in $\Gamma$ a type $t_a$ in $\Sigma_i$. The sets $\Gamma_1$ and $\Gamma_2$ obtained this way are indeed sets of assertions of $\mathcal{L}_1$ and $\mathcal{L}_2$, respectively. Thus, their satisfiability can effectively be checked using the decision procedures for $S_1$ and $S_2$. This proves Theorem 29.

The argument used above also shows why in Theorem 30 it was not sufficient to state equivalence of (1) and (2) (as in Theorem 20). In fact, the induction argument used above does not necessarily always apply to the computation of $\Sigma_1$. In some cases, the alternation depth may not decreases for $\Sigma_1$, but only for $\Sigma_2$. It should be noted that Theorem 20 could also have been formulated in this symmetric way. We have not done this since it was not necessary for proving Theorem 17.

Regarding the complexity of the combined decision procedure, we must in principle also consider the complexity of computing covering normal terms and the size of these terms. In the examples from DL, these terms are just value restrictions, and thus their size and the complexity of computing them is linear. Here, we assume a polynomial bound on both. Under this assumption, we obtain the same complexity results as for the case of relativized satisfiability. In fact, the complexity of testing Condition (2) and (3) of Theorem 30 agrees with the complexity of testing Condition (2) of Theorem 20: it adds one exponential to the complexity of the decision procedure for the single ADSs. In order to compute $\Sigma_i$, we need exponentially many recursive calls to the procedure. Since the recursion depth is linear in the size of $\Gamma$, we end up with at most exponentially many tests of Condition (2) and (3).

**Corollary 33.** *Let $S_1$ and $S_2$ be local ADSs having covering normal terms, and assume that these covering normal terms can be computed in polynomial time. If the satisfiability problems for $S_1$ and $S_2$ are decidable in* ExpTime *(*PSpace*), then the satisfiability problem for $S_1 \otimes S_2$ is decidable in* 2ExpTime *(*ExpSpace*).*

With the same argument as in the case of relativized satisfiability, we can extend the transfer result also to *term* satisfiability.

**Corollary 34.** *Let $S_1$ and $S_2$ be local ADSs having covering normal terms, and suppose that the term satisfiability problems for $S_1$ and $S_2$ are decidable. Then the* term *satisfiability problem for $S_1 \otimes S_2$ is also decidable.*

## 5. Fusions of description logics

Given two DLs $L_1$ and $L_2$, their fusion is defined as follows. We translate them into the corresponding ADSs $S_1$ and $S_2$, and then build the fusion $S_1 \otimes S_2$. The fusion $L_1 \otimes L_2$ of $L_1$ and $L_2$ is the DL that corresponds to $S_1 \otimes S_2$. Since the definition of the fusion of ADSs requires their sets of function symbols to be disjoint, we must ensure that the ADSs corresponding to $L_1$ and $L_2$ are built over disjoint sets of function symbols. For the DLs introduced in Section 2, this can be achieved by assuming that the sets of role names of $L_1$ and $L_2$ are disjoint and the sets of nominals of $L_1$ and $L_2$ are disjoint. The DL $L_1 \otimes L_2$ then allows the use of the concept and role constructors of both DLs, but in a restricted way. Role descriptions are either role descriptions of $L_1$ or of $L_2$. There are no role descriptions involving constructors or names of both DLs. Concept descriptions may contain concept constructors of both DLs; however, a constructor of $L_i$ may only use a role description of $L_i$ $(i = 1, 2)$.

Let us illustrate these restrictions by two simple examples. The fusion $\mathcal{ALC}^+ \otimes \mathcal{ALC}^{-1}$ of the two DLs $\mathcal{ALC}^+$ and $\mathcal{ALC}^{-1}$ is the fragment of $\mathcal{ALC}^{+,-1}$ whose set of role names is partitioned into two sets $N_{R_1}$ and $N_{R_2}$ such that

- the transitive closure operator may only be applied to names from $N_{R_1}$;

- the inverse operator may only be applied to names from $N_{R_2}$.

For example, if $A$ is a concept name, $R \in N_{R_1}$ and $Q \in N_{R_2}$, then $\exists R^+.A \sqcap \forall Q^{-1}.\neg A$ is a concept description of $\mathcal{ALC}^+ \otimes \mathcal{ALC}^{-1}$, but $\exists R^+.A \sqcap \forall R^{-1}.\neg A$ and $\exists (Q^{-1})^+.A$ are not. Note that, although the two source DLs have disjoint sets of role names, in $\mathcal{ALC}^+ \otimes \mathcal{ALC}^{-1}$ role names from both sets may be used inside existential and value restrictions since these concept constructors are available in both DLs.

The fusion $\mathcal{ALCQ} \otimes \mathcal{ALC}_{R^+}$ of the two DLs $\mathcal{ALCQ}$ and $\mathcal{ALC}_{R^+}$ is the fragment of $\mathcal{ALCQ}_{R^+}$ whose set of role names $N_R$ (with transitive roles $N_{R^+} \subseteq N_R$) is partitioned into two sets $N_{R_1}$ and $N_{R_2}$ with $N_{R^+} \subseteq N_{R_2}$ such that, inside qualifying number restrictions, only role names from $N_{R_1}$ may be used. In particular, this means that transitive roles cannot occur within qualified number restrictions.

In the following, we give examples that illustrate the usefulness of the transfer results proved in the previous section. First, we will give an example for the case of satisfiability and then for relativized satisfiability. Subsequently, we will consider a more complex example involving so-called concrete domains. Here, our general transfer result can be used to prove a decidability result that has only recently been proved by designing a specialized algorithm for the fusion. Finally, we will give an example that demonstrates that the restriction to local ADSs is really necessary.

## 5.1 Decidability transfer for satisfiability

In this subsection, we will give an example for an application of Theorem 29 where the decidability result could not be obtained using Theorem 17.

Theorem 29 requires the ADSs to have covering normal terms. This is, however, satisfied by all the DLs that yield local ADSs.

**Proposition 35.** *Let $L$ be one of the DLs introduced in Section 2, and let the corresponding ADS $S = (\mathcal{L}, \mathcal{M})$ be local. Then $S$ has covering normal terms, and these terms can be computed in linear time.*

**Proof.** For all function symbols $f$ in $\mathcal{L}$, the term $t_f$ has the form $f_{\forall R}(x)$ for some role description $R$. The semantics of value restrictions implies that terms of this form satisfy the first two properties of Definition 26. This completes the proof for all function symbols $f$ of arity 0 since for these the third condition of Definition 26 is trivially satisfied. Thus, for nullary function symbols, $f_{\forall R}(x)$ for an arbitrary role name $R$ does the job.

It remains to show that, for every unary function symbol $f \in \{f_{\exists R}, f_{\forall R}, f_{\dot{\geq} nR}, f_{\dot{\leq} nR}\}$, the term $f_{\forall R}(x)$ also satisfies the third property. This is an immediate consequence of the fact that, for these function symbols $f$, we have $f_{\forall R}^{\mathfrak{M}}(X) \cap f^{\mathfrak{M}}(Y) = f_{\forall R}^{\mathfrak{M}}(X) \cap f^{\mathfrak{M}}(X \cap Y)$ for all models $\mathfrak{M} \in \mathcal{M}$ and $X, Y \subseteq W$. ❏

In the following, we consider the two description logics $\mathcal{ALCF}$ and $\mathcal{ALC}^{+,\circ,\sqcup}$. Hollunder and Nutt (1990) show that satisfiability of $\mathcal{ALCF}$-concept descriptions is decidable. The same is true for consistency of $\mathcal{ALCF}$-ABoxes (Lutz, 1999). Note, however, that relativized satisfiability of $\mathcal{ALCF}$-concept descriptions and thus also relativized ABox consistency in $\mathcal{ALCF}$ is undecidable (Baader et al., 1993). For $\mathcal{ALC}^{+,\circ,\sqcup}$, decidability of satisfiability is shown by Baader (1991) and Schild (1991).[9] Decidability of ABox consistency in $\mathcal{ALC}^{+,\circ,\sqcup}$ is shown in Chapter 7 of (De Giacomo, 1995).

The unrestricted combination $\mathcal{ALCF}^{+,\circ,\sqcup}$ of the two DLs is undecidable. To be more precise, satisfiability of $\mathcal{ALCF}^{+,\circ,\sqcup}$-concept descriptions (and thus also consistency of $\mathcal{ALCF}^{+,\circ,\sqcup}$-ABoxes) is undecidable. This follows from the undecidability of relativized satisfiability of $\mathcal{ALCF}$-concept descriptions and the fact that the role operators in $\mathcal{ALCF}^{+,\circ,\sqcup}$ can be used to internalize TBoxes (Schild, 1991; Baader et al., 1993). In contrast to the undecidability of $\mathcal{ALCF}^{+,\circ,\sqcup}$, Theorem 29 immediately implies that satisfiability of concept descriptions in the fusion of $\mathcal{ALCF}$ and $\mathcal{ALC}^{+,\circ,\sqcup}$ is decidable.

**Theorem 36.** *Satisfiability of concept descriptions and consistency of ABoxes is decidable in $\mathcal{ALCF} \otimes \mathcal{ALC}^{+,\circ,\sqcup}$, whereas satisfiability of $\mathcal{ALCF}^{+,\circ,\sqcup}$-concept descriptions is already undecidable.*

Taking the fusion thus yields a decidable combination of two DLs whose unrestricted combination is undecidable. The price one has to pay is that the fusion offers less expressivity than the unrestricted combination. The concept $f_1 \!\downarrow\! f_2 \sqcap \forall f_1^+.C$ is an example of a concept description of $\mathcal{ALCF}^{+,\circ,\sqcup}$ that is not allowed in the fusion $\mathcal{ALCF} \otimes \mathcal{ALC}^{+,\circ,\sqcup}$.

## 5.2 Decidability transfer for relativized satisfiability

As an example for the application of Corollary 22 (and thus of Theorem 17), we consider the DL $\mathcal{ALC}_f^{+,\circ,\sqcap,\sqcup}$. For this DL, satisfiability of concept descriptions is undecidable. However, an expressive fragment with a decidable relativized satisfiability problem can be obtained by building the fusion of the two sublanguages $\mathcal{ALC}_f^{+,\circ,\sqcup}$ and $\mathcal{ALC}^{+,\circ,\sqcup,\sqcap}$.

**Theorem 37.** *Satisfiability of $\mathcal{ALC}_f^{+,\circ,\sqcap,\sqcup}$-concept descriptions is undecidable.*

Undecidability can be shown by a reduction of the domino problem (Berger, 1966; Knuth, 1973) (see, e.g., Baader & Sattler, 1999, for undecidability proofs of DLs using such a reduction). The main tasks to solve in such a reduction is that one can express the $\mathbb{N} \times \mathbb{N}$ grid and that one can access all points on the grid. One square of the grid can be expressed by a description of the form $\exists(x \circ y \sqcap y \circ x).\top$, where $x, y$ are features. In fact, this description expresses that the "points" belonging to it have both an $x \circ y$ and a $y \circ x$ successor, and that these two successors coincide. Accessing all point on the grid can then be achieved by using the role description $(x \sqcup y)^+$.

---

9. Note that $\mathcal{ALC}^{+,\circ,\sqcup}$ is a notational variant of test-free propositional dynamic logic (PDL) (Fischer & Ladner, 1979).

Note that this undecidability result is also closely related to the known undecidability of IDPDL, i.e., deterministic propositional dynamic logic with intersection (Harel, 1984). However, the undecidability proof for IDPDL by Harel (1984) uses the test construct, which is not available in $\mathcal{ALC}_f^{+,\circ,\sqcap,\sqcup}$.

Next, we show that relativized satisfiability in two rather expressive sublanguages of $\mathcal{ALC}_f^{+,\circ,\sqcap,\sqcup}$ is decidable.

**Theorem 38.** *Relativized satisfiability of concept descriptions is decidable in $\mathcal{ALC}_f^{+,\circ,\sqcup}$ and $\mathcal{ALC}^{+,\circ,\sqcup,\sqcap}$.*

*Proof sketch.* In both cases, TBoxes can be internalized as described by Schild (1991) and Baader et al. (1993). Thus, it is sufficient to show decidability of (unrelativized) satisfiability.

For $\mathcal{ALC}_f^{+,\circ,\sqcup}$, this follows from decidability of DPDL (Ben-Ari, Halpern, & Pnueli, 1982), the known correspondence between PDL and $\mathcal{ALC}^{+,\circ,\sqcup}$ (Schild, 1991), and the fact that non-functional roles can be simulated by functional ones in the presence of composition and transitive closure (Parikh, 1980).

For $\mathcal{ALC}^{+,\circ,\sqcup,\sqcap}$, decidability of satisfiability follows from IPDL, i.e., PDL with intersection (Danecki, 1984). ❏

Given this theorem, Corollary 22 now yields the following decidability result.

**Corollary 39.** *Relativized satisfiability of concept descriptions is decidable in the fusion $\mathcal{ALC}_f^{+,\circ,\sqcup} \otimes \mathcal{ALC}^{+,\circ,\sqcup,\sqcap}$.*

## 5.3 A "concrete" example

Description logics with concrete domains were introduced by Baader and Hanschke (1991) in order to allow for the reference to concrete objects like numbers, time intervals, spatial regions, etc. when defining concepts. To be more precise, Baader and Hanschke (1991) define the extension $\mathcal{ALC}(\mathcal{D})$ of $\mathcal{ALC}$, where $\mathcal{D}$ is a concrete domain (see below). Under suitable assumptions on $\mathcal{D}$, they show that satisfiability in $\mathcal{ALC}(\mathcal{D})$ is decidable. One of the main problems with this extension of DLs is that relativized satisfiability (and satisfiability in DLs where TBoxes can be internalized) is usually undecidable (Baader & Hanschke, 1992) (though there are exceptions, see Lutz, 2001). For this reason, Haarslev et al. (2001) introduce a restricted way of extending DLs by concrete domains, and show that the corresponding extension of $\mathcal{ALCN}_{\mathcal{HR}^+}$ has a decidable relativized satisfiability problem.[10] In the following, we show that this result can also be obtained as an easy consequence of our Theorem 17. Moreover, $\mathcal{ALCN}_{\mathcal{HR}^+}$ can be replaced by an arbitrary local DL with a decidable relativized satisfiability problem.

---

10. To be more precise, they even show that relativized ABox consistency is decidable in their restricted extension of $\mathcal{ALCN}_{\mathcal{HR}^+}$ by concrete domains. Here, we restrict ourself to satisfiability of concepts since the ABoxes introduced by Haarslev et al. (2001) also allow for the use of concrete individuals and for predicate assertions on these individuals, which is not covered by the object assertions for ADSs introduced in the present paper.

**Definition 40 (Concrete Domain).** A *concrete domain* $\mathcal{D}$ is a pair $(\Delta_{\mathcal{D}}, \Phi_{\mathcal{D}})$, where $\Delta_{\mathcal{D}}$ is a nonempty set called the domain, and $\Phi_{\mathcal{D}}$ is a set of predicate names. Each predicate name $P \in \Phi_{\mathcal{D}}$ is associated with an arity $n$ and an $n$-ary predicate $P^{\mathcal{D}} \subseteq \Delta_{\mathcal{D}}^n$. A concrete domain $\mathcal{D}$ is called *admissible* iff (1) the set of its predicate names is closed under negation and contains a name $\top_{\mathcal{D}}$ for $\Delta_{\mathcal{D}}$, and (2) the satisfiability problem for finite conjunctions of predicates is decidable.

Given a concrete domain $\mathcal{D}$ and one of the predicates $P \in \Phi_{\mathcal{D}}$ (of arity $n$), one can define a new concept constructor $\exists f_1, \ldots, f_n.P$ (predicate restriction), where $f_1, \ldots, f_n$ are concrete features.[11] In contrast to the abstract features considered until now, concrete features are interpreted by partial functions from the abstract domain $\Delta^{\mathcal{I}}$ into the concrete domain $\Delta_{\mathcal{D}}$. We consider the basic DL that allows for Boolean operators and these new concept constructors only.

**Definition 41 ($\mathcal{B}(\mathcal{D})$).** Let $N_C$ be a set of concept names and $N_{F_c}$ be a set of names for concrete features disjoint from $N_C$, and let $\mathcal{D}$ be an admissible concrete domain. Concepts descriptions of $\mathcal{B}(\mathcal{D})$ are Boolean combinations of concept names and *predicate restrictions*, i.e., expressions of the form $\exists f_1, \ldots, f_n.P$ where $P$ is an $n$-ary predicate in $\Phi_{\mathcal{D}}$ and $f_1, \ldots, f_n \in N_{F_c}$.

The semantics of $\mathcal{B}(\mathcal{D})$ is defined as follows. We consider an interpretation $\mathcal{I}$, which has a nonempty domain $\Delta^{\mathcal{I}}$, and interprets concept names as subsets of $\Delta^{\mathcal{I}}$ and concrete features as partial functions from $\Delta^{\mathcal{I}}$ into $\Delta_{\mathcal{D}}$. The Boolean operators are interpreted as usual, and

$$(\exists f_1, \ldots, f_n.P)^{\mathcal{I}} = \{a \in \Delta^{\mathcal{I}} \mid \exists x_1, \ldots, x_n \in \Delta_{\mathcal{D}}.$$
$$f_i^{\mathcal{I}}(a) = x_i \text{ for all } 1 \leq i \leq n \text{ and } (x_1, \ldots, x_n) \in P^{\mathcal{D}}\}.$$

Note that concept descriptions are interpreted as subsets of $\Delta^{\mathcal{I}}$ and not of $\Delta^{\mathcal{I}} \cup \Delta_{\mathcal{D}}$. Thus, if we go to the ADS corresponding to $\mathcal{B}(\mathcal{D})$, the concrete domain is not an explicit part of the corresponding ADMs. It is only used to define the interpretation of the function symbols corresponding to predicate restrictions. The predicate restriction constructor is translated into a function symbol $f_{\exists f_1, \ldots, f_n.P}$ and, for an ADM $\mathfrak{W}$ corresponding to a frame $\mathfrak{F}$, $f_{\exists f_1, \ldots, f_n.P}^{\mathfrak{W}}$ is defined as $(\exists f_1, \ldots, f_n.P)^{\mathcal{I}_{\emptyset}}$, where $\mathcal{I}_{\emptyset}$ is the interpretation based on $\mathfrak{F}$ that maps all concept names to the empty set.

**Theorem 42.** *Let $\mathcal{D}$ be an admissible concrete domain. Then, $\mathcal{B}(\mathcal{D})$ is local and the relativized satisfiability problem for $\mathcal{B}(\mathcal{D})$-concept descriptions is decidable.*

**Proof.** Given the family $(\mathfrak{W}_i)_{i \in I}$ of ADMs $\mathfrak{W}_i$ corresponding to the frames $\mathfrak{F}_i$ over pairwise disjoint domains $\Delta^{\mathfrak{F}_i}$ $(i \in I)$, we first build the union $\mathfrak{F}$ of the frames: the domain of $\mathfrak{F}$ is $\bigcup_{i \in I} \Delta^{\mathfrak{F}_i}$ and it interprets the concrete features in the obvious way, i.e., $f^{\mathfrak{F}}(x) := f^{\mathfrak{F}_i}(x)$ if $x \in \Delta^{\mathfrak{F}_i}$. Let $\mathfrak{W}$ be the ADM induced by $\mathfrak{F}$. To prove that $\mathfrak{W}$ is in fact the disjoint union of $(\mathfrak{W}_i)_{i \in I}$, it remains to show that $f_{\exists f_1,\ldots,f_n.P}^{\mathfrak{W}} = \bigcup_{i \in I} f_{\exists f_1,\ldots,f_n.P}^{\mathfrak{W}_i}$. This is an easy consequence of the semantics of the predicate restriction constructor, the interpretation of the concrete features in $\mathfrak{F}$, and the fact that the domains $\Delta^{\mathfrak{F}_i}$ are pairwise disjoint.

---

11. Note that the general framework introduced by Baader and Hanschke (1991) allows for feature chains in predicate restrictions. Considering only feature chains of length one is the main restriction introduced by Haarslev et al. (2001).

Decidability of the unrelativized satisfiability problem is an immediate consequence of the decidability results for $\mathcal{ALC}(\mathcal{D})$ given by Baader and Hanschke (1991). Since $\mathcal{B}(\mathcal{D})$ is a very simple DL that does not contain any concept constructors requiring the generation of abstract individuals, it is easy to see that a $\mathcal{B}(\mathcal{D})$-concept description $C_0$ is satisfiable relative to the TBox $C_1 \sqsubseteq D_1, \ldots, C_n \sqsubseteq D_n$ iff it is satisfiable in a one-element interpretation. But then the TBox can be internalized in a very simple way: $C_0$ is satisfiable relative to the TBox $C_1 \sqsubseteq D_1, \ldots, C_n \sqsubseteq D_n$ iff $C_0 \sqcap (\neg C_1 \sqcup D_1) \sqcap \ldots \sqcap (\neg C_n \sqcup D_n)$ is satisfiable. ❏

Given this theorem, Corollary 22 now yields the following transfer result, which shows that concrete domains with the restricted form of predicate restrictions introduced above can be integrated into any local DL with a decidable relativized satisfiability problem without losing decidability.

**Corollary 43.** *Let $\mathcal{D}$ be an admissible concrete domain and $L$ be a local DL for which relativized satisfiability of concept descriptions is decidable. Then, relativized satisfiability of concept descriptions in $\mathcal{B}(\mathcal{D}) \otimes L$ is also decidable.*

### 5.4 Non-local DLs

By Proposition 15, DLs allowing for nominals, the universal role, or role negation are not local. It follows that the decidability transfer theorems are not applicable to fusions of such DLs. In the following, we try to clarify the reasons for this restricted applicability of the theorems.

First, we show that there are DLs with decidable satisfiability problem such that their fusion has an undecidable satisfiability problem. The culprit in this case is the universal role (or role negation).

**Theorem 44.** *Satisfiability of concept descriptions is decidable in $\mathcal{ALC}^U$ and $\mathcal{ALCF}$, but undecidable in their fusion $\mathcal{ALC}^U \otimes \mathcal{ALCF}$.*

**Proof.** Decidability of $\mathcal{ALCF}$ was shown by Hollunder and Nutt (1990) and of $\mathcal{ALC}^U$ by Baader et al. (1990) and Goranko and Passy (1992). Undecidability of $\mathcal{ALC}^U \otimes \mathcal{ALCF}$ (which is identical to $\mathcal{ALCF}^U$) follows from the results by Baader et al. (1993) and the fact that the universal role can be used to simulate TBoxes (see Proposition 24). ❏

Note that role negation can be used to simulate the universal role: just replace $\forall U.C$ by $\forall R.C \sqcap \forall \overline{R}.C$ and $\exists U.C$ by $\exists R.C \sqcup \exists \overline{R}.C$. In addition, decidability of $\mathcal{ALC}^{\overline{\phantom{.}}}$ is known to be decidable (Lutz & Sattler, 2000). Consequently, the theorem also holds if we replace $\mathcal{ALC}^U$ by $\mathcal{ALC}^{\overline{\phantom{.}}}$.

It should be noted that the example given in the above theorem depends on the fact that one of the two DLs allows for the universal role and the other becomes undecidable if the universal role is added. In fact, Corollary 25 shows that decidability does transfer if both DLs already provide for the universal role.

Concerning nominals, we do not have a counterexample to the transfer of decidability in their presence. However, we think that it is very unlikely that there can be a general transfer result in this case. In fact, note that for each DL $L$ without nominals introduced in Section 2, its fusion with $\mathcal{ALCO}$ is identical to $L$ extended with nominals. Since (relativized) satisfiability in $\mathcal{ALCO}$ is decidable, a general transfer result in this case would imply that this extension is decidable provided that $L$ is decidable. Consequently, this would yield a general transfer result for adding nominals.

## 6. Conclusion

Regarding related work, the work that is most closely related to the one presented here is (Wolter, 1998). There, analogs of our Theorems 20 and 30 are proved for normal modal logics within an algebraic framework. The present results extend the ones from Wolter (1998) in two directions. First, we have added object assertions, and thus can also prove transfer results for ABox reasoning. Second, we can show transfer results for satisfiability in non-normal modal logics as long as we have covering normal terms. This allows us to handle non-normal concept constructors like qualified number restrictions (graded modalities) in our framework.

We also think that the introduction of abstract description systems (ADSs) is a contribution in its own right. ADSs abstract from the internal structure of concept constructors and thus allow us to treat a vast range of such constructors in a uniform way. Nevertheless, the model theoretic semantics provided by ADSs is less abstract than the algebraic semantics employed by Wolter (1998). It is closer to the usual semantics of DLs, and thus easier to comprehend for people used to this semantics. The results in this paper show that ADSs in fact yield a good level of abstraction for proving general results on description logics. Recently, the same notion has been used for proving general results about so-called E-connections of representation formalisms like description logics, modal spatial logics, and temporal logics (Kutz, Wolter, & Zakharyaschev, 2001). In contrast to fusions, in an E-connection the two domains are not merged but connected by means of relations.

Regarding complexity, our transfer results yield only upper bounds. Basically, they show that the complexity of the algorithm for the fusion is at most one exponent higher than of the ones for the components. We believe that the complexity of satisfiability in the fusion of ADSs can indeed be exponentially higher than the complexity of satisfiability in the component ADSs. However, we do not yet have matching lower bounds, i.e., we know of no example where this exponential increase in the complexity really happens.

Note that Spaan's results (1993) on the transfer of NP and PSPACE decidability from the component modal logics to their fusion are restricted to normal modal logics, and that they make additional assumptions on the algorithms used to solve the satisfiability problem in the component logics. Nevertheless, for many PSPACE-complete description logics it is easy to see that their fusion is also PSPACE-complete. In this sense, the general techniques for reasoning in the fusion of descriptions logics developed in this paper give only a rough complexity estimate.

# Appendix A. Proofs

In this appendix, we give detailed proofs of criteria for (relativized) satisfiability in the fusion of local ADSs. Recall that, from these criteria, the transfer theorems for decidability easily follow. We have deferred the proofs of these theorems to the appendix since they are rather technical.

## A.1 Proof of Theorem 20

Before we can prove this theorem, we need a technical lemma. In the proof of Theorem 20, we are going to merge models $\mathfrak{W}_1 \in \mathcal{M}_1$ and $\mathfrak{W}_2 \in \mathcal{M}_2$ by means of a bijective function $b$ from the domain $W_1$ of $\mathfrak{W}_1$ onto the domain $W_2$ of $\mathfrak{W}_2$ in such a way that the surrogates $sur_i(t)$, $t \in \mathcal{C}^1(\Gamma)$, are respected by $b$ in the sense that

$$w \in sur_1(t)^{\mathfrak{W}_1, \mathcal{A}^1} \Leftrightarrow b(w) \in sur_2(t)^{\mathfrak{W}_2, \mathcal{A}^2}$$

for all $w \in W_1$ and $t \in \mathcal{C}^1(\Gamma)$. The existence of such a bijection is equivalent to the condition that the cardinalities $|sur_1(t)^{\mathfrak{W}_1, \mathcal{A}^1}|$ of $sur_1(t)^{\mathfrak{W}_1, \mathcal{A}^1}$ and $|sur_2(t)^{\mathfrak{W}_2, \mathcal{A}^2}|$ of $sur_2(t)^{\mathfrak{W}_2, \mathcal{A}^2}$ coincide for all $t \in \mathcal{C}^1(\Gamma)$: if $t \neq t'$ for $t, t' \in \mathcal{C}^1(\Gamma)$, then $t$ contains a conjunct which is (equivalent to) the negation of a conjunct of $t'$; hence, for all such $t, t'$, we have $sur_i(t)^{\mathfrak{W}_i, \mathcal{A}^i} \cap sur_i(t')^{\mathfrak{W}_i, \mathcal{A}^i} = \emptyset$ for $i \in \{1, 2\}$, which clearly yields the above equivalence. The following lemma will be used to choose models in such a way that this cardinality condition is satisfied. (We refer the reader to, e.g., Grätzer, 1979 for information about cardinals.)

**Lemma 45.** *Let $(\mathcal{L}, \mathcal{M})$ be a local ADS and $\Gamma$ a set of assertions satisfiable in $(\mathcal{L}, \mathcal{M})$. Then there exists a cardinal $\kappa$ such that, for all cardinals $\kappa' \geq \kappa$, there exists a model $\mathfrak{W} = \langle W, \mathcal{F}^{\mathfrak{W}}, \mathcal{R}^{\mathfrak{W}} \rangle \in \mathcal{M}$ with $|W| = \kappa'$ and an assignment $\mathcal{A}$ with $\langle \mathfrak{W}, \mathcal{A} \rangle \models \Gamma$ and $|s^{\mathfrak{W}, \mathcal{A}}| \in \{0, \kappa'\}$ for all terms $s$.*

**Proof.** By assumption, there exists an ADM $\mathfrak{W}_0 = \langle W_0, \mathcal{F}^{\mathfrak{W}_0}, \mathcal{R}^{\mathfrak{W}_0} \rangle \in \mathcal{M}$ and an assignment $\mathcal{B} = \langle \mathcal{B}_1, \mathcal{B}_2 \rangle$ in it such that $\langle \mathfrak{W}_0, \mathcal{B} \rangle \models \Gamma$. Let $\kappa = \max\{\aleph_0, |W_0|\}$. We show that $\kappa$ is as required. Let $\kappa' \geq \kappa$. Take $\kappa'$ disjoint isomorphic copies $\langle \mathfrak{W}_\rho, \mathcal{B}_1^\rho \rangle$, $\mathfrak{W}_\rho = \langle W_\rho, \mathcal{F}^{\mathfrak{W}_\rho}, \mathcal{R}^{\mathfrak{W}_\rho} \rangle$, $\rho < \kappa'$, of $\langle \mathfrak{W}_0, \mathcal{B}_1 \rangle$. (The first member of the list coincides with $\mathfrak{W}_0$.) Let $\mathfrak{W} = \langle W, \mathcal{F}^{\mathfrak{W}}, \mathcal{R}^{\mathfrak{W}} \rangle$ be the disjoint union of the $\mathfrak{W}_\rho$, $\rho < \kappa'$, and define $\langle \mathfrak{W}, \mathcal{A} = \langle \mathcal{A}_1, \mathcal{A}_2 \rangle \rangle$ by putting $\mathcal{A}_2(a) = \mathcal{B}_2(a)$, for all $a \in \mathcal{X}$, and

$$\mathcal{A}_1(x) = \bigcup_{\rho < \kappa'} \mathcal{B}_1^\rho(x),$$

for all $x \in V$. Note that all object variables are interpreted in $W_0$. It follows from the definitions of term semantics and disjoint unions that

$$s^{\mathfrak{W}, \mathcal{A}} = \bigcup_{\rho < \kappa'} s^{\mathfrak{W}_\rho, \mathcal{B}_\rho}, \tag{$*$}$$

for all terms $s$. Hence $|W| = \kappa'$ and $\langle \mathfrak{W}, \mathcal{A} \rangle \models \Gamma$. It remains to show that $|s^{\mathfrak{W}, \mathcal{A}}| \in \{0, \kappa'\}$ for every term $s$. Suppose $|s^{\mathfrak{W}, \mathcal{A}}| \neq 0$. Then, by $(*)$, $\kappa' \leq |s^{\mathfrak{W}, \mathcal{A}}| \leq \kappa \times \kappa' = \kappa'$, which means $\kappa' = |s^{\mathfrak{W}, \mathcal{A}}|$. ❑

As noted above, the disjointness of the sets $sur_i(t)^{\mathfrak{W}_i, \mathcal{A}^i}$ and $sur_i(t')^{\mathfrak{W}_i, \mathcal{A}^i}$ (for $t \neq t'$) is required in order to ensure the existence of the bijection $b$. More precisely, in order to merge models $\mathfrak{W}_1, \mathfrak{W}_2$, the sets $sur_i(t)^{\mathfrak{W}_i, \mathcal{A}^i}$ for $t$ member of some "relevant" subset of $\mathcal{C}^1(\Gamma)$ must form a partition of $\mathfrak{W}_i$'s domain that satisfies a certain cardinality condition. This is formalized by the following definition:

**Definition 46.** Let $\kappa$ be a cardinal. A set $\{X_1, \ldots, X_n\}$ is called a $\kappa$-partition of a set $W$ iff

1. $|X_i| = \kappa$, for all $1 \leq i \leq n$,

2. $X_i \cap X_j = \emptyset$ whenever $i \neq j$, and

3. $W = \bigcup_{1 \leq i \leq n} X_i$.

$\{X_1, \ldots, X_n\}$ is a $\kappa$-partition of an ADM $\mathfrak{W}$ with domain $W$ iff it is a $\kappa$-partition of $W$.

In the proof, we will enforce that Properties 1 and 3 hold by appropriate constructions, while Property 2 holds by definition of $\mathcal{C}^1(\Gamma)$.

Before proving Theorem 20, we repeat its formulation.

**Theorem 20.** *Let $S_i = (\mathcal{L}_i, \mathcal{M}_i)$, $i \in \{1, 2\}$, be two local ADSs in which $\mathcal{L}_1$ is based on the set of function symbols $\mathcal{F}$ and relation symbols $\mathcal{R}$, and $\mathcal{L}_2$ is based on $\mathcal{G}$ and $\mathcal{Q}$, and let $\mathcal{L} = \mathcal{L}_1 \otimes \mathcal{L}_2$ and $\mathcal{M} = \mathcal{M}_1 \otimes \mathcal{M}_2$. If $\Gamma$ is a finite set of assertions from $\mathcal{L}$, then the following are equivalent:*

1. *$\Gamma$ is satisfiable in $(\mathcal{L}, \mathcal{M})$.*

2. *There exist*

   (a) *a set $D \subseteq \mathcal{C}^1(\Gamma)$,*

   (b) *for every term $t \in D$ an object variable $a_t \notin obj(\Gamma)$,*

   (c) *for every $a \in obj(\Gamma)$ a term $t_a \in D$,*

   *such that the union $\Gamma_1$ of the following sets of assertions in $\mathcal{L}_1$ is satisfiable in $(\mathcal{L}_1, \mathcal{M}_1)$:*

   (d) *$\{a_t : sur_1(t) \mid t \in D\} \cup \{\top \sqsubseteq sur_1(\bigvee D)\}$,*

   (e) *$\{a : sur_1(t_a) \mid a \in obj(\Gamma)\}$,*

   (f) *$\{R(a, b) \mid R(a, b) \in \Gamma, R \in \mathcal{R}\}$,*

   (g) *$\{sur_1(t_1) \sqsubseteq sur_1(t_2) \mid t_1 \sqsubseteq t_2 \in \Gamma\} \cup \{a : sur_1(s) \mid (a : s) \in \Gamma\}$;*

   *and the union $\Gamma_2$ of the following sets of assertions in $\mathcal{L}_2$ is satisfiable in $(\mathcal{L}_2, \mathcal{M}_2)$:*

   (h) *$\{a_t : sur_2(t) \mid t \in D\} \cup \{\top \sqsubseteq sur_2(\bigvee D)\}$,*

   (i) *$\{a : sur_2(t_a) \mid a \in obj(\Gamma)\}$,*

   (j) *$\{Q(a, b) \mid Q(a, b) \in \Gamma, Q \in \mathcal{Q}\}$.*

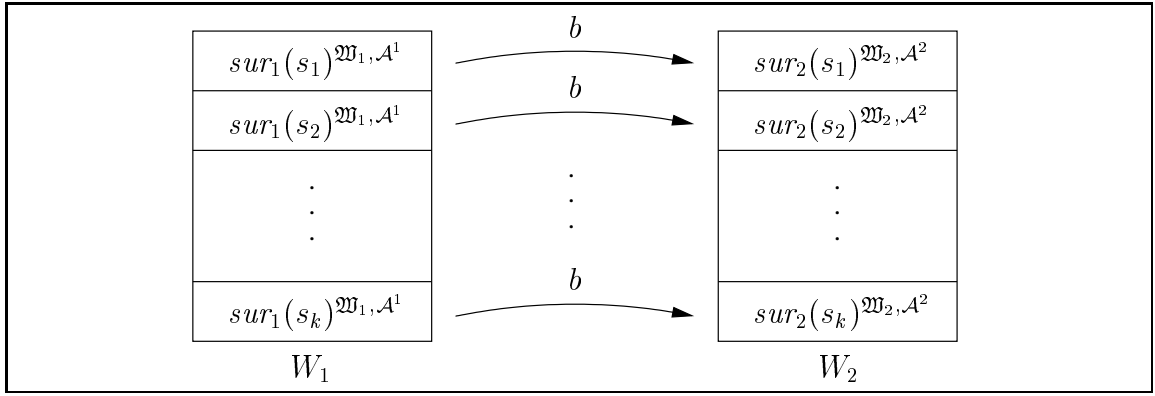

Figure 3: The mapping $b$.

**Proof.** We start with the direction from (2) to (1). Take a set $D \subseteq \mathcal{C}^1(\Gamma)$ satisfying the properties listed in the theorem. Take cardinals $\kappa_i$, $i \in \{1, 2\}$ as in Lemma 45 for $(\mathcal{L}_i, \mathcal{M}_i)$, put $\kappa = max\{\kappa_1, \kappa_2\}$, and take $\langle \mathfrak{W}_1, \mathcal{A}^1 = \langle \mathcal{A}^1_1, \mathcal{A}^1_2 \rangle \rangle$ and $\langle \mathfrak{W}_2, \mathcal{A}^2 = \langle \mathcal{A}^2_1, \mathcal{A}^2_2 \rangle \rangle$ with $\mathfrak{W}_i \in \mathcal{M}_i$ such that $\langle \mathfrak{W}_i, \mathcal{A}^i \rangle \models \Gamma_i$ for $i \in \{1, 2\}$. By Lemma 45, for $i \in \{1, 2\}$ we can assume $|W_i| = \kappa$ and, $|sur_i(s)^{\mathfrak{W}_i, \mathcal{A}^i}| \in \{0, \kappa\}$ for all $s \in D$.

The sets $\{sur_i(s)^{\mathfrak{W}_i, \mathcal{A}^i} : s \in D\}$ are $\kappa$-partitions of $W_i$ for $i \in \{0, 1\}$ since (i) for each $s \in D$, we have $(a_s : sur_i(s)) \in \Gamma_i$, (ii) $\langle \mathfrak{W}_i, \mathcal{A}^i \rangle \models \top \sqsubseteq sur_i(\bigvee D)$, and (iii) $s, s' \in D$ and $s \neq s'$ implies $sur_i(s)^{\mathfrak{W}_i, \mathcal{A}^i} \cap sur_i(s')^{\mathfrak{W}_i, \mathcal{A}^i}$ by definition of $D$ and $\mathcal{C}^1$. Moreover, $obj(\Gamma_1) = obj(\Gamma_2)$ and, for all $a \in obj(\Gamma_1)$ and $s \in D$, we have $\mathcal{A}^1_2(a) \in sur_1(s)^{\mathfrak{W}_1, \mathcal{A}^1}$ iff $\mathcal{A}^2_2(a) \in sur_2(s)^{\mathfrak{W}_2, \mathcal{A}^2}$.

Together with the fact that $\mathcal{A}^1_2$ and $\mathcal{A}^2_2$ are injective, this implies the existence of a bijection $b$ from $W_1$ onto $W_2$ such that

$$\{b(w) : w \in sur_1(t)^{\mathfrak{W}_1, \mathcal{A}^1}\} = sur_2(t)^{\mathfrak{W}_2, \mathcal{A}^2},$$

for all $t \in D$, and

$$b(\mathcal{A}^1_2(a)) = \mathcal{A}^2_2(a),$$

for all $a \in obj(\Gamma_1)$. Figure 3, in which it is assumed that $D = \{s_1, \ldots, s_k\}$, illustrates the mapping $b$.

Define a model $\mathfrak{W} = \langle W, (\mathcal{F} \cup \mathcal{G})^{\mathfrak{W}}, (\mathcal{R} \cup \mathcal{Q})^{\mathfrak{W}} \rangle \in \mathcal{M}$ by putting

- $W = W_1$,

- $f^{\mathfrak{W}} = f^{\mathfrak{W}_1}$, for $f \in \mathcal{F}$,

- for all $g \in \mathcal{G}$ of arity $n$ and all $Z_1, \ldots, Z_n \subseteq W$,

$$g^{\mathfrak{W}}(Z_1, \ldots, Z_n) = b^{-1}(g^{\mathfrak{W}_2}(b(Z_1), \ldots, b(Z_n))),$$

   where $b(Z) = \{b(z) : z \in Z\}$,

- $R^{\mathfrak{W}} = R^{\mathfrak{W}_1}$, for all $R \in \mathcal{R}$,

- $Q^{\mathfrak{W}}(x, y)$ iff $Q^{\mathfrak{W}_2}(b(x), b(y))$, for all $Q \in \mathcal{Q}$.

Since $\mathcal{M}_2$ is closed under isomorphic copies, it is not hard to see that $\mathfrak{W} \in \mathcal{M}_1 \otimes \mathcal{M}_2$. Let $\mathcal{A} = \mathcal{A}^1$. To prove the implication from (2) to (1) of the theorem it remains to show that $\langle \mathfrak{W}, \mathcal{A} \rangle \models \Gamma$. To this end it suffices to prove the following claim:

**Claim.** For all terms $t \in sub^1(\Gamma)$, we have

$$t^{\mathfrak{W}, \mathcal{A}} = sur_1(t)^{\mathfrak{W}_1, \mathcal{A}^1} = b^{-1}(sur_2(t)^{\mathfrak{W}_2, \mathcal{A}^2}).$$

Before we prove this claim, let us show that it implies $\langle \mathfrak{W}, \mathcal{A} \rangle \models \Gamma$. First note that, from the claim, we obtain

$$t^{\mathfrak{W}, \mathcal{A}} = sur_1(t)^{\mathfrak{W}_1, \mathcal{A}^1} \text{ for all } t \in term(\Gamma). \tag{1}$$

This may be proved by induction on the construction of $t \in term(\Gamma)$ from terms in $sub^1(\Gamma)$ using the booleans and function symbols from $\mathcal{L}_1$, only. The basis of induction (i.e., the equality for members of $sub^1(\Gamma)$) is stated in the claim and the induction step is straightforward.

We now show that $\langle \mathfrak{W}, \mathcal{A} \rangle \models \Gamma$ is a consequence of (1). Suppose $R(a, b) \in \Gamma$. Then $R(a, b) \in \Gamma_1$ and thus $\langle \mathfrak{W}, \mathcal{A} \rangle \models R(a, b)$. Similarly, $Q(a, b) \in \Gamma$ implies $Q(a, b) \in \Gamma_2$ and $\langle \mathfrak{W}, \mathcal{A} \rangle \models Q(a, b)$. Suppose $(a : t) \in \Gamma$. Then $(a : sur_1(t)) \in \Gamma_1$ and so $\mathcal{A}_2^1(a) \in sur_1(t)^{\mathfrak{W}_1, \mathcal{A}^1}$ which implies, by (1), $\mathcal{A}_2^1(a) \in t^{\mathfrak{W}, \mathcal{A}}$. Hence $\langle \mathfrak{W}, \mathcal{A} \rangle \models (a : t)$. If $t_1 \sqsubseteq t_2 \in \Gamma$, then $sur_1(t_1) \sqsubseteq sur_1(t_2) \in \Gamma_1$ and so, by (1), $t_1^{\mathfrak{W}, \mathcal{A}} \subseteq t_2^{\mathfrak{W}, \mathcal{A}}$. Hence $\langle \mathfrak{W}, \mathcal{A} \rangle \models t_1 \sqsubseteq t_2$.

We come to the proof of the claim. It is proved by induction on the structure of $t$. Due to the following equalities holding for all $t \in sub^1(\Gamma)$, it suffices to show that $t^{\mathfrak{W}, \mathcal{A}} = sur_1(t)^{\mathfrak{W}_1, \mathcal{A}^1}$.

$$\begin{aligned}
sur_1(t)^{\mathfrak{W}_1, \mathcal{A}^1} &= \bigcup \{ sur_1(s)^{\mathfrak{W}_1, \mathcal{A}^1} : s \in D, t \text{ is a conjunct of } s \} \\
&= \bigcup \{ b^{-1}(sur_2(s)^{\mathfrak{W}_2, \mathcal{A}^2}) : s \in D, t \text{ is a conjunct of } s \} \\
&= b^{-1}(sur_2(t)^{\mathfrak{W}_2, \mathcal{A}^2})
\end{aligned}$$

The first equality holds since $sur_1(\bigvee D)^{\mathfrak{W}_1, \mathcal{A}^1} = W_1$ and, for all $s \in D$, either $t$ or $\neg t$ is a conjunct of $s$. The second equality is true by definition of $b$ and the validity of the third equality can be seen analogously to the validity of the first one by considering that $sur_2(\bigvee D)^{\mathfrak{W}_2, \mathcal{A}^2} = W_2$.

Hence let us show $t^{\mathfrak{W}, \mathcal{A}} = sur_1(t)^{\mathfrak{W}_1, \mathcal{A}^1}$. For the induction start, let $t$ be a variable. The equation $t^{\mathfrak{W}, \mathcal{A}} = sur_1(t)^{\mathfrak{W}_1, \mathcal{A}^1}$ is an immediate consequence of the fact that $\mathcal{A} = \mathcal{A}^1$. For the induction step, we distinguish several cases:

- $t = \neg t_1$. By induction hypothesis, $t_1^{\mathfrak{W}, \mathcal{A}} = sur_1(t_1)^{\mathfrak{W}_1, \mathcal{A}_1}$. Hence, $t^{\mathfrak{W}, \mathcal{A}} = W \setminus t_1^{\mathfrak{W}, \mathcal{A}} = W \setminus sur_1(t_1)^{\mathfrak{W}_1, \mathcal{A}^1} = sur_1(t)^{\mathfrak{W}_1, \mathcal{A}^1}$ (since $W = W_1$).

- $t = t_1 \wedge t_2$. By induction hypothesis, $t_i^{\mathfrak{W}, \mathcal{A}} = sur_1(t_i)^{\mathfrak{W}_1, \mathcal{A}_1}$ for $i \in \{1, 2\}$. Hence, $t^{\mathfrak{W}, \mathcal{A}} = t_1^{\mathfrak{W}, \mathcal{A}} \cap t_2^{\mathfrak{W}, \mathcal{A}} = sur_1(t_1)^{\mathfrak{W}_1, \mathcal{A}^1} \cap sur_1(t_2)^{\mathfrak{W}_1, \mathcal{A}^1} = sur_1(t)^{\mathfrak{W}_1, \mathcal{A}^1}$.

- $t = t_1 \vee t_2$. Similar to the above case.

- $t = f(t_1, \ldots, t_n)$. By induction hypothesis, $t_i^{\mathfrak{W}, \mathcal{A}} = sur_1(t_i)^{\mathfrak{W}_1, \mathcal{A}^1}$ for $1 \leq i \leq n$. Hence, $t^{\mathfrak{W}, \mathcal{A}} = f^{\mathfrak{W}}(t_1^{\mathfrak{W}, \mathcal{A}}, \ldots, t_n^{\mathfrak{W}, \mathcal{A}}) = f^{\mathfrak{W}}(sur_1(t_1)^{\mathfrak{W}_1, \mathcal{A}^1}, \ldots, sur_1(t_n)^{\mathfrak{W}_1, \mathcal{A}^1}) = sur_1(t)^{\mathfrak{W}_1, \mathcal{A}^1}$ (since $f^{\mathfrak{W}} = f^{\mathfrak{W}_1}$).

- $t = g(t_1, \ldots, t_n)$. In this case, $t^{\mathfrak{W}, \mathcal{A}} = b^{-1}(g^{\mathfrak{W}_2}(b(t_1^{\mathfrak{W}, \mathcal{A}}), \ldots, b(t_n^{\mathfrak{W}, \mathcal{A}})))$. Since, by the above equalities, $sur_1(t)^{\mathfrak{W}_1, \mathcal{A}^1} = b^{-1}(sur_2(t)^{\mathfrak{W}_2, \mathcal{A}^2})$, it remains to show that $sur_2(t)^{\mathfrak{W}_2, \mathcal{A}^2} = g^{\mathfrak{W}_2}(b(t_1^{\mathfrak{W}, \mathcal{A}}), \ldots, b(t_n^{\mathfrak{W}, \mathcal{A}}))$. Since we have $sur_2(t)^{\mathfrak{W}_2, \mathcal{A}^2} = g^{\mathfrak{W}_2}(sur_2(t_1)^{\mathfrak{W}_2, \mathcal{A}^2}, \ldots, sur_2(t_n)^{\mathfrak{W}_2, \mathcal{A}^2})$, this amounts to showing that $b(t_i^{\mathfrak{W}, \mathcal{A}}) = sur_2(t_i)^{\mathfrak{W}_2, \mathcal{A}^2}$ for $1 \leq i \leq n$. This, however, follows by induction hypothesis together with the above equations.

This concludes the proof of the direction from (2) to (1).

It remains to prove the direction from (1) to (2). Suppose $\langle \mathfrak{W}, \mathcal{A} \rangle \models \Gamma$, for some $\mathfrak{W} \in \mathcal{M}$ and $\mathcal{A} = \langle \mathcal{A}_1, \mathcal{A}_2 \rangle$. Put

$$D = \{s \in \mathcal{C}^1(\Gamma) : s^{\mathfrak{W}, \mathcal{A}} \neq \emptyset\}.$$

Note that the fusion of local ADLs is a local ADL again. Hence $(\mathcal{L}, \mathcal{M})$ is local and we may assume, by Lemma 45, that the sets $s^{\mathfrak{W}, \mathcal{A}}$ are infinite.

Take a new object name $a_s \notin obj(\Gamma)$ for every $s \in D$ and let, for $a \in obj(\Gamma)$,

$$t_a = \bigwedge\{t \in sub^1(\Gamma) : \mathcal{A}_2(a) \in t^{\mathfrak{W}, \mathcal{A}}\} \wedge \bigwedge\{\neg t : t \in sub^1(\Gamma), \mathcal{A}_2(a) \notin t^{\mathfrak{W}, \mathcal{A}}\}.$$

We prove that set of assertions $\Gamma_1$ based on $D$, $t_a$, $a \in obj(\Gamma)$, and $a_s$, $s \in D$, is satisfiable in $(\mathcal{L}_1, \mathcal{M}_1)$.

Let $\mathcal{F}^{\mathfrak{W}}$ denote the restriction of $(\mathcal{F} \cup \mathcal{G})^{\mathfrak{W}}$ to the symbols in $\mathcal{F}$. Similarly, $\mathcal{R}^{\mathfrak{W}}$ is the restriction of $(\mathcal{R} \cup \mathcal{Q})^{\mathfrak{W}}$ to the symbols in $\mathcal{R}$. Set $\mathfrak{W}_1 = \langle W, \mathcal{F}^{\mathfrak{W}}, \mathcal{R}^{\mathfrak{W}} \rangle \in \mathcal{M}_1$, $\mathcal{A}^1 = \langle \mathcal{A}_1^1, \mathcal{A}_2^1 \rangle$, where

$$\mathcal{A}_1^1 = \mathcal{A}_1 \cup \{x_t \mapsto t^{\mathfrak{W}, \mathcal{A}} : t = g(t_1, \ldots, t_k) \in sub^1(\Gamma)\},$$

$\mathcal{A}_2^1(a) = \mathcal{A}_2(a)$, for $a \in obj(\Gamma)$, and $\mathcal{A}_2^1(a_s) \in s^{\mathfrak{W}, \mathcal{A}}$, for all $s \in D$. Note that we can choose an injective function $\mathcal{A}_2^1$ because the $s^{\mathfrak{W}, \mathcal{A}}$ are infinite. We show by induction that

$$sur_1(t)^{\mathfrak{W}_1, \mathcal{A}_1} = t^{\mathfrak{W}, \mathcal{A}} \text{ for all } t \in term(\Gamma). \tag{2}$$

Let $t = x$ be a variable. Then $x$ is not a surrogate, and so $\mathcal{A}_1^1(x) = \mathcal{A}_1(x)$. For the induction step, we distinguish several cases:

- The inductive steps for $t = \neg t_1$, $t = t_1 \wedge t_2$, $t = t_1 \vee t_2$, and $t = f(t_1, \ldots, t_n)$, $f \in \mathcal{F}$, are identical to the corresponding cases in the proof of Equation 1, which occurs in the direction that (2) implies (1) above.

- $t = g(t_1, \ldots, t_n)$, where $g \in \mathcal{G}$. Then $sur_1(t) = x_t$. Hence $\mathcal{A}_1^1(x_t) = t^{\mathfrak{W}, \mathcal{A}}$ and the equation is proved.

From Equation 2, we obtain $\langle \mathfrak{W}_1, \mathcal{A}^1 \rangle \models \Gamma_1$: we prove $\langle \mathfrak{W}_1, \mathcal{A}^1 \rangle \models R(a, b)$ whenever $R(a, b) \in \Gamma_1$ and $\langle \mathfrak{W}_1, \mathcal{A}^1 \rangle \models sur_1(t_1) \sqsubseteq sur_1(t_2)$ whenever $sur_1(t_1) \sqsubseteq sur_1(t_2) \in \Gamma_1$. The remaining formulas from $\Gamma_1$ are left to the reader. Suppose $R(a, b) \in \Gamma_1$. Then $R(a, b) \in \Gamma$ and so $\langle \mathfrak{W}, \mathcal{A} \rangle \models R(a, b)$. Hence $\langle \mathfrak{W}_1, \mathcal{A}^1 \rangle \models R(a, b)$. Suppose $sur_1(t_1) \sqsubseteq sur_1(t_2) \in \Gamma_1$.

Then $t_1 \sqsubseteq t_2 \in \Gamma$. Hence $\langle \mathfrak{W}, \mathcal{A} \rangle \models t_1 \sqsubseteq t_2$ which means $t_1^{\mathfrak{W}, \mathcal{A}} \subseteq t_2^{\mathfrak{W}, \mathcal{A}}$. By Equation 2, $sur_1(t_1)^{\mathfrak{W}_1, \mathcal{A}^1} \subseteq sur_1(t_2)^{\mathfrak{W}_1, \mathcal{A}^1}$ which means $\langle \mathfrak{W}_1, \mathcal{A}^1 \rangle \models sur_1(t_1) \sqsubseteq sur_1(t_2)$.

The construction of a model in $\mathcal{M}_2$ satisfying $\Gamma_2$ is similar and left to the reader.

❏

## A.2 Proof of Theorem 30

As in the proof of Theorem 17, we fix two local ADSs $S_i = (\mathcal{L}_i, \mathcal{M}_i)$, $i \in \{1, 2\}$, in which $\mathcal{L}_1$ is based on the set of function symbols $\mathcal{F}$ and relation symbols $\mathcal{R}$, and $\mathcal{L}_2$ is based on $\mathcal{G}$ and $\mathcal{Q}$. Let $\mathcal{L} = \mathcal{L}_1 \otimes \mathcal{L}_2$ and $\mathcal{M} = \mathcal{M}_1 \otimes \mathcal{M}_2$. We assume that $S_1$ and $S_2$ have covering normal terms.

Similarly to what was done in the previous section, we will merge models by means of bijections which map points in sets $sur_1(t)^{\mathfrak{W}_1, \mathcal{A}^1}$ to points in the corresponding sets $sur_2(t)^{\mathfrak{W}_2, \mathcal{A}^2}$. For a finite set of object assertions $\Gamma$ of $\mathcal{L}$, let $\Sigma_i(\Gamma)$ denote the set of all $s \in \mathcal{C}^i(\Gamma)$ such that the term $s$ is satisfiable in $(\mathcal{L}, \mathcal{M})$ (for $i \in \{1, 2\}$). To ensure that the merging of models succeeds, we must enforce that the elements of $\Sigma_1(\Gamma)$ and $\Sigma_2(\Gamma)$ form $\kappa$-partitions (for some appropriate $\kappa$) of the models to be merged. For $\Sigma_1(\Gamma)$, this is captured by the following lemma. Explicitly stating a dual of this lemma for $\Sigma_2(\Gamma)$ is omitted for brevity.

**Lemma 47.** *Let $\Gamma$ be a finite set of object assertions of $\mathcal{L}$, $\kappa$ a cardinal satisfying the conditions of Lemma 45 for $(\mathcal{L}, \mathcal{M})$ and $\Gamma$, and $\Sigma_1 = \Sigma_1(\Gamma)$. If $\kappa' \geq \kappa$, then*

1. *there exists a model $\mathfrak{W} \in \mathcal{M}_1$ and an assignment $\mathcal{A}$ such that*

$$\{sur_1(s)^{\mathfrak{W}, \mathcal{A}} \mid s \in \Sigma_1\}$$

   *is a $\kappa'$-partition of $\mathfrak{W}$; and*

2. *there exists a model $\mathfrak{W} \in \mathcal{M}_2$ and an assignment $\mathcal{A}$ such that*

$$\{sur_2(s)^{\mathfrak{W}, \mathcal{A}} \mid s \in \Sigma_1\}$$

   *is a $\kappa'$-partition of $\mathfrak{W}$.*

**Proof.** 1. By definition of $\Sigma_1$, for each $s \in \Sigma_1$, we find a model $\mathfrak{W}_s \in \mathcal{M}$ and an assignment $\mathcal{A}_s$ such that $s^{\mathfrak{W}_s, \mathcal{A}_s} \neq \emptyset$. Since the fusion of two local ADSs is again local, the set of models $\mathcal{M}$ is closed under disjoint unions. Hence, there exists a model $\mathfrak{W}_{\Sigma_1}$ and an assignment $\mathcal{A}_{\Sigma_1}$ such that $s^{\mathfrak{W}_{\Sigma_1}, \mathcal{A}_{\Sigma_1}} \neq \emptyset$ for all $s \in \Sigma_1$. It follows that the set $\Gamma_1 := \{a_s : s \mid s \in \Sigma_1\}$ is satisfiable in $(\mathcal{L}, \mathcal{M})$. By Lemma 45, there thus exists a model $\mathfrak{W}' = \left\langle W', (\mathcal{F} \cup \mathcal{G})^{\mathfrak{W}'}, (\mathcal{R} \cup \mathcal{Q})^{\mathfrak{W}'} \right\rangle \in \mathcal{M}$ and an assignment $\mathcal{A}'$ such that $\mathfrak{W}', \mathcal{A}' \models \Gamma_1$ and $\{s^{\mathfrak{W}', \mathcal{A}'} \mid s \in \Sigma_1\}$ is a $\kappa'$-partition of $W'$. Now let $\mathfrak{W}$ denote the restriction of $\mathfrak{W}'$ to $\mathcal{L}_1$ and define

$$\mathcal{A}_1 = \mathcal{A}_1' \cup \{x_t \mapsto t^{\mathfrak{W}', \mathcal{A}'} \mid t = g(t_1, \ldots, t_k) \in sub^1(\Gamma)\}.$$

Then $\langle \mathfrak{W}, \mathcal{A} \rangle$ is as required. To prove this note that $sur_1(t)^{\mathfrak{W}, \mathcal{A}} = t^{\mathfrak{W}', \mathcal{A}'}$ for all $t \in term(\Gamma)$.

2. is similar and left to the reader. ❏

We repeat the formulation of the theorem to be proved.

**Theorem 30.** *Let $S_i = (\mathcal{L}_i, \mathcal{M}_i)$, $i \in \{1, 2\}$, be two local ADSs having covering normal terms in which $\mathcal{L}_1$ is based on the set of function symbols $\mathcal{F}$ and relation symbols $\mathcal{R}$, and $\mathcal{L}_2$ is based on $\mathcal{G}$ and $\mathcal{Q}$, and let $\mathcal{L} = \mathcal{L}_1 \otimes \mathcal{L}_2$ and $\mathcal{M} = \mathcal{M}_1 \otimes \mathcal{M}_2$. Let $\Gamma$ be a finite set of object assertions from $\mathcal{L}$. Put $m := d_F(\Gamma)$, $r := d_G(\Gamma)$, and let $c(x)$ $(d(x))$ be a covering normal term for all function symbols in $\Gamma$ that are in $F$ $(G)$.*

*For $i \in \{1, 2\}$, denote by $\Sigma_i$ the set of all $s \in C^i(\Gamma)$ such that the term $s$ is satisfiable in $(\mathcal{L}, \mathcal{M})$. Then the following three conditions are equivalent:*

1. *$\Gamma$ is satisfiable in $(\mathcal{L}, \mathcal{M})$.*

2. *There exist*

   - *for every $t \in \Sigma_1$ an object variable $a_t \notin obj(\Gamma)$*
   - *for every $a \in obj(\Gamma)$ a term $t_a \in \Sigma_1$*

   *such that the union $\Gamma_1$ of the following sets of object assertions is satisfiable in $(\mathcal{L}_1, \mathcal{M}_1)$:*

   - *$\{a_t : sur_1(t \wedge c^{\leq m}(sur_1(\bigvee \Sigma_1))) \mid t \in \Sigma_1\}$,*
   - *$\{a : sur_1(t_a \wedge c^{\leq m}(sur_1(\bigvee \Sigma_1))) \mid a \in obj(\Gamma)\}$,*
   - *$\{R(a, b) \mid R(a, b) \in \Gamma, R \in \mathcal{R}\}$,*
   - *$\{a : sur_1(s) \mid (a : s) \in \Gamma\}$;*

   *and the union $\Gamma_2$ of the following sets of object assertions is satisfiable in $(\mathcal{L}_2, \mathcal{M}_2)$:*

   - *$\{a_t : sur_2(t \wedge d^{\leq r}(sur_2(\bigvee \Sigma_1))) \mid t \in \Sigma_1\}$,*
   - *$\{a : sur_2(t_a \wedge d^{\leq r}(sur_2(\bigvee \Sigma_1))) \mid a \in obj(\Gamma)\}$,*
   - *$\{Q(a, b) \mid Q(a, b) \in \Gamma, Q \in \mathcal{Q}\}$.*

3. *The same condition as in (2) above, with $\Sigma_1$ replaced by $\Sigma_2$.*

We start the proof with the direction from (1) to (2) and (1) to (3). The proofs are dual to each other, so we only give a proof for (1) $\Rightarrow$ (2). Suppose $\langle \mathfrak{W}, \mathcal{A} \rangle \models \Gamma$, where $\mathfrak{W} = \langle W, (\mathcal{F} \cup \mathcal{G})^{\mathfrak{W}}, (\mathcal{R} \cup \mathcal{Q})^{\mathfrak{W}} \rangle$. By Lemma 45, we can assume that that, for every $t \in \Sigma_1$, $|t^{\mathfrak{W}, \mathcal{A}}|$ is infinite. Take a new object name $a_s \notin obj(\Gamma)$ for every $s \in \Sigma_1$ and let, for $a \in obj(\Gamma)$,

$$t_a = \bigwedge \{t \in sub^1(\Gamma) : \mathcal{A}_2(a) \in t^{\mathfrak{W}, \mathcal{A}}\} \wedge \bigwedge \{\neg t : t \in sub^1(\Gamma), \mathcal{A}_2(a) \notin t^{\mathfrak{W}, \mathcal{A}}\}.$$

We prove that the set $\Gamma_1$ of assertions based on $t_a$, $a \in obj(\Gamma)$, and $a_s$, $s \in \Sigma_1$, is satisfiable in $(\mathcal{L}_1, \mathcal{M}_1)$ (the proof is rather similar to the proof of the direction from (1) to (2) in the proof of Theorem 20). Let $\mathcal{F}^{\mathfrak{W}}$ (resp. $\mathcal{G}^{\mathfrak{W}}$) denote the restriction of $(\mathcal{F} \cup \mathcal{G})^{\mathfrak{W}}$ to the symbols in $\mathcal{F}$ (resp. $\mathcal{G}$). Similarly, $\mathcal{R}^{\mathfrak{W}}$ and $\mathcal{Q}^{\mathfrak{W}}$ are the restrictions of $(\mathcal{R} \cup \mathcal{Q})^{\mathfrak{W}}$ to the symbols in $\mathcal{R}$ and $\mathcal{Q}$, respectively. Set $\mathfrak{W}_1 = \langle W, \mathcal{F}^{\mathfrak{W}}, \mathcal{R}^{\mathfrak{W}} \rangle \in \mathcal{M}_1$, $\mathcal{A}^1 = \langle \mathcal{A}_1^1, \mathcal{A}_2^1 \rangle$, where

$$\mathcal{A}_1^1 = \mathcal{A}_1 \cup \{x_t \mapsto t^{\mathfrak{W}, \mathcal{A}} \mid t = g(t_1, \ldots, t_k) \in sub^1(\Gamma)\},$$

$\mathcal{A}_2^1(a) = \mathcal{A}_2(a)$, for $a \in obj(\Gamma)$, and $\mathcal{A}_2^1(a_t) \in t^{\mathfrak{W},\mathcal{A}}$, for all $t \in \Sigma_1$ (we can choose an injective function for $\mathcal{A}_2^1$ since the sets $t^{\mathfrak{W},\mathcal{A}}$ are infinite).

As in the corresponding part of the proof of Theorem 20, it can show by induction that

$$sur_1(t)^{\mathfrak{W}_1,\mathcal{A}_1} = t^{\mathfrak{W},\mathcal{A}} \text{ for all } t \in term(\Gamma).$$

Let us see now why $\langle \mathfrak{W}_1, \mathcal{A}^1 \rangle \models \Gamma_1$ follows from this equation. For $R(a,b) \in \Gamma_1$ we have $R(a,b) \in \Gamma$ and so $\langle \mathfrak{W}, \mathcal{A} \rangle \models R(a,b)$. Hence $\langle \mathfrak{W}_1, \mathcal{A}^1 \rangle \models R(a,b)$. We have $\langle \mathfrak{W}, \mathcal{A} \rangle \models (\bigvee \Sigma_1) = \top$ (by the definition of $\Sigma_1$). Hence $\langle \mathfrak{W}_1, \mathcal{A}^1 \rangle \models sur_1(\bigvee \Sigma_1) = \top$ and so, by the definition of $c^{\leq m}$, $\langle \mathfrak{W}_1, \mathcal{A}^1 \rangle \models (c^{\leq m}(sur_1(\bigvee \Sigma_1))) = \top$. It remains to observe that $\mathcal{A}_2^1(a) \in sur_1(t_a)^{\mathfrak{W}_1,\mathcal{A}_1}$ for all $a \in obj(\Gamma)$, $\mathcal{A}_2^1(a) \in sur_1(s)^{\mathfrak{W}_1,\mathcal{A}_1}$ whenever $(a:s) \in \Gamma$, and $\mathcal{A}_2^1(a_t) \in sur_1(t)^{\mathfrak{W}_1,\mathcal{A}_1}$ for all $t \in \Sigma_1$.

The construction of a model in $\mathcal{M}_2$ satisfying $\Gamma_2$ is similar and left to the reader.

It remains to show the implications $(2) \Rightarrow (1)$ and $(3) \Rightarrow (1)$. They are similar, so we concentrate on the first. In the proof of Theorem 20 it was possible to construct the required model for $\Gamma$ by merging models for $\Gamma_1$ and $\Gamma_2$. The situation is different here. It is not possible to merge models for $\Gamma_1$ and $\Gamma_2$ in one step, since we do not know whether they satisfy $sur_1(\bigvee \Sigma_1) = \top$ and $sur_2(\bigvee \Sigma_1) = \top$, respectively. We only know that they satisfy the approximations $a : sur_1(s) \wedge c^{\leq m}(sur_1(\bigvee \Sigma_1))$ and $a : sur_2(s) \wedge d^{\leq r}(sur_2(\bigvee \Sigma_1))$, respectively, for $a : s \in \Gamma$. To merge models of this type we have to distinguish various pieces of the models and have to add new pieces as well. To define those pieces we need a technical claim. As in the proof of Theorem 17, take cardinals $\kappa_i$, $i \in \{1,2\}$ as in Lemma 45 for $(\mathcal{L}_i, \mathcal{M}_i)$ and put $\kappa = max\{\kappa_1, \kappa_2\}$.

**Claim 1.** Suppose (2) holds.

(a) There exist $\mathfrak{W}_1 = \langle W_1, \mathcal{F}^{\mathfrak{W}}, \mathcal{R}^{\mathfrak{W}} \rangle \in \mathcal{M}_1$, an assignment $\mathcal{A} = \langle \mathcal{A}_1, \mathcal{A}_2 \rangle$ into $\mathfrak{W}_1$, and a sequence $X_0, \ldots, X_m$ of subsets of $W_1$ such that

[a1] $\mathcal{A}_2(a) \in X_m$, for all $a \in obj(\Gamma_1)$,

[a2] $\langle \mathfrak{W}_1, \mathcal{A} \rangle \models \Gamma_1$,

[a3] $X_{n+1} \subseteq X_n \cap c^{\mathfrak{W}_1}(X_n)$, for all $0 \leq n < m$,

[a4] The set $\{sur_1(s)^{\mathfrak{W}_1,\mathcal{A}} \cap X_m : s \in \Sigma_1\}$ is a $\kappa$-partition of $X_m$,

[a5] The sets

$$\{sur_1(s)^{\mathfrak{W}_1,\mathcal{A}} \cap (X_n - X_{n+1}) : s \in \Sigma_1\}$$

are $\kappa$-partitions of $X_n - X_{n+1}$, for $0 \leq n < m$.

[a6] $|W_1 - X_0| = \kappa$.

(b) There exist $\mathfrak{W}_2 = \langle W_2, \mathcal{G}^{\mathfrak{W}}, \mathcal{Q}^{\mathfrak{W}} \rangle \in \mathcal{M}_2$, an assignment $\mathcal{B} = \langle \mathcal{B}_1, \mathcal{B}_2 \rangle$, and a sequence $Y_0, \ldots, Y_r$ of subsets of $W_2$ such that

[b1] $\mathcal{B}_2(a) \in Y_r$, for all $a \in obj(\Gamma_1)$,

[b2] $\langle \mathfrak{W}_2, \mathcal{B} \rangle \models \Gamma_2$,

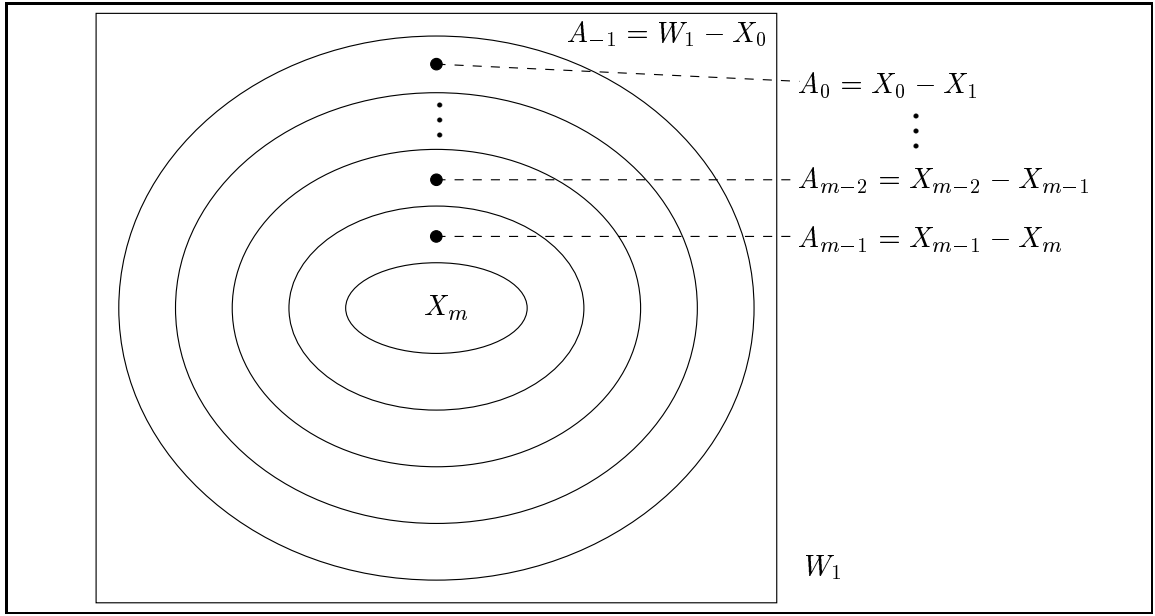

Figure 4: The sets $X_i$.

[b3] $Y_{n+1} \subseteq Y_n \cap d^{\mathfrak{W}_2}(Y_n)$, for all $0 \le n < r$,

[b4] The set $\{sur_2(s)^{\mathfrak{M},\mathcal{A}} \cap Y_r : s \in \Sigma_1\}$ is a $\kappa$-partition of $Y_r$,

[b5] The sets

$$\{sur_2(s)^{\mathfrak{M},\mathcal{A}} \cap (Y_n - Y_{n+1}) : s \in \Sigma_1\}$$

are $\kappa$-partitions of $Y_n - Y_{n+1}$, for $0 \le n < r$.

[b6] $|W_2 - Y_0| = \kappa$.

Figure 4 illustrates the relation between the sets $X_i$. (We set $A_i = X_i - X_{i+1}$ for $0 \le i < m$ and $A_{-1} = W_1 - X_0$.) Intuitively, $X_m$ is the set of points for which we know that points in $W_1 - sur_1(\bigvee \Sigma_1)^{\mathfrak{W}_1}$ are "very far away". For $X_{m-1}$ they are possibly less "far away", for $X_{m-2}$ possibly even "less far", and so on for $X_i$, $i < m-1$. Finally, for members of $A_{-1}$ it is not even known whether they are in $sur_1(\bigvee \Sigma_1)^{\mathfrak{W}_1,\mathcal{A}}$ or not. Note that all object names are interpreted in $X_m$. We now come to the formal construction of the sets $X_i$.

*Proof of Claim 1.* We prove (a). Part (b) is proved similarly and left to the reader. By assumption and Lemma 45, we find an ADM $\mathfrak{W}_a = \langle W_a, F^{\mathfrak{W}_a}, R^{\mathfrak{W}_a} \rangle \in \mathcal{M}_1$ with $|W_a| = \kappa$ and an assignment $\mathcal{A}^a = \langle \mathcal{A}_1^a, \mathcal{A}_2^a \rangle$ such that $\langle \mathfrak{W}_a, \mathcal{A}^a \rangle \models \Gamma_1$.

Let

$$Z_n = (c^{\le n}(sur_1(\bigvee \Sigma_1)))^{\mathfrak{W}_a,\mathcal{A}_a},$$

for $0 \le n \le m$. By Lemma 47 (1) we can take for every $n$ with $-1 \le n \le m$ an ADM $\mathfrak{W}_n = \langle W_n, \mathcal{F}^{\mathfrak{W}_n}, \mathcal{R}^{\mathfrak{W}_n} \rangle \in \mathcal{M}_1$ and assignments $\mathcal{A}^n$ such that

$$\{sur_1(s)^{\mathfrak{W}_n,\mathcal{A}^n} : s \in \Sigma_1\}$$

are $\kappa$-partitions of $\mathfrak{W}_n$.

Take the disjoint union $\mathfrak{W}$ (with $\mathfrak{W} = \langle W, \mathcal{F}^{\mathfrak{W}}, \mathcal{R}^{\mathfrak{W}} \rangle$) of the $\mathfrak{W}_n$, $-1 \leq n \leq m$, and $\mathfrak{W}_a$. Define $\mathcal{A} = \langle \mathcal{A}_1, \mathcal{A}_2 \rangle$ in $\mathfrak{W}$ by putting

$$\mathcal{A}_1(x) = \mathcal{A}_1^a(x) \cup \bigcup_{-1 \leq i \leq m} \mathcal{A}_1^i(x),$$

for all set variables $x$ and $\mathcal{A}_2(b) = \mathcal{A}_2^a(b)$, for all object variables $b$. Let, for $0 \leq n \leq m$,

$$X_n = Z_n \cup \bigcup_{n \leq i \leq m} W_i.$$

We show that $\langle \mathfrak{W}, \mathcal{A} \rangle$ and the sets $X_n$, $0 \leq n \leq m$, are as required.

[a1] We have $\langle \mathfrak{W}_a, \mathcal{A}^a \rangle \models \Gamma_1$ and so $\mathcal{A}_2(b) = \mathcal{A}_2^a(b) \in Z_m$ for all $b \in obj(\Gamma_1)$. Hence $\mathcal{A}_2(b) \in X_m = Z_m \cup W_m$ for all $b \in obj(\Gamma_1)$.

[a2] By the definition of disjoint unions and because $\langle \mathfrak{W}_a, \mathcal{A}^a \rangle \models \Gamma_1$.

[a3] Firstly, we have, by the definition of $c^{\leq n}t$ and since $c^{\mathfrak{W}}$ is monotone (it distributes over intersections),

$$Z_{n+1} \subseteq Z_n \cap c^{\mathfrak{W}}(Z_n) \subseteq X_n \cap c^{\mathfrak{W}}(X_n). \tag{3}$$

Secondly, by the definition of disjoint unions, the first property of covering normal terms, and since $c^{\mathfrak{W}}$ is monotone

$$\bigcup_{n+1 \leq i \leq m} W_i \subseteq \bigcup_{n \leq i \leq m} W_i \subseteq \bigcup_{n \leq i \leq m} W_i \cap c^{\mathfrak{W}}(\bigcup_{n \leq i \leq m} W_i) \subseteq X_n \cap c^{\mathfrak{W}} X_n. \tag{4}$$

From (3) and (4) we obtain

$$X_{n+1} = Z_{n+1} \cup \bigcup_{n+1 \leq i \leq m} W_i \subseteq X_n \cap c^{\mathfrak{W}} X_n. \tag{5}$$

[a4] We show that the three properties from Definition 46 are satisfied. Since

$$\{sur_1(s)^{\mathfrak{W}_m, \mathcal{A}_m} : s \in \Sigma_1\}$$

is a $\kappa$-partition of $W_m$, we have $|sur_1(s)^{\mathfrak{W}_m, \mathcal{A}_m}| = \kappa$ for all $s \in \Sigma_1$. This implies Property 1 since $sur_1(s)^{\mathfrak{W}, \mathcal{A}} \cap W_m = sur_1(s)^{\mathfrak{W}_m, \mathcal{A}_m}$, $W_m \subseteq X_m$, and $|X_m| \leq \kappa$.

Property 2 is an immediate consequence of the definition of $\Sigma_1$. As for Property 3, we show that, for all $w \in X_m$, we have $w \in s^{\mathfrak{W}, \mathcal{A}}$ for an $s \in \Sigma_1$. Fix a $w \in X_m$. We distinguish two cases: firstly, assume $w \in W_m$. Then, by the fact that $\{sur_1(s)^{\mathfrak{W}_m, \mathcal{A}_m} : s \in \Sigma_1\}$ is a $\kappa$-partition of $W_m$, it is clear that there exists an $s \in \Sigma_1$ as required. Secondly, assume $w \in Z_m = (c^{\leq m}(sur_1(\bigvee \Sigma_1)))^{\mathfrak{W}_a, \mathcal{A}_a}$. By definition of $c^{\leq m} t$, we have $w \in (sur_1(\bigvee \Sigma_1))^{\mathfrak{W}_a, \mathcal{A}_a}$ and so again $w \in sur_1(s)^{\mathfrak{W}, \mathcal{A}}$ for some $s \in \Sigma_1$.

[a5] The proof is similar to that of Property [a4].

[a6] By definition.

This finishes the proof of Claim 1.

Suppose now that we have

$$\mathfrak{W}_1 = \left\langle W_1, \mathcal{F}^{\mathfrak{W}_1}, \mathcal{R}^{\mathfrak{W}_1} \right\rangle, \ \mathcal{A}, \ X_m, \ldots, X_0 \text{ and } \mathfrak{W}_2 = \left\langle W_2, \mathcal{G}^{\mathfrak{W}_2}, \mathcal{Q}^{\mathfrak{W}_2} \right\rangle, \ \mathcal{B}, \ Y_r, \ldots, Y_0$$

satisfying the properties listed in Claim 1. We may assume that

$$(W_1 - X_m) \cap (W_2 - Y_r) = \emptyset.$$

Using an appropriate bijection $b$ from $X_m$ onto $Y_r$ we may also assume that $X_m = Y_r$, $\mathcal{A}_2(a) = \mathcal{B}_2(a)$ for all object variables $a \in obj(\Gamma_1)$, and

$$sur_1(s)^{\mathfrak{W}_1, \mathcal{A}} \cap X_m = sur_2(s)^{\mathfrak{W}_2, \mathcal{B}} \cap X_m \text{ for all } s \in \Sigma_1. \tag{6}$$

This follows from the fact that all object variables are mapped by $\mathcal{A}_2$ and $\mathcal{B}_2$ into $X_m$ and $Y_r$ ([a1], [b1]), respectively, the injectivity of the mappings $\mathcal{A}_2$ and $\mathcal{B}_2$, and the conditions [a4] and [b4] which state that $\{sur_1(s)^{\mathfrak{W}_1, \mathcal{A}} \cap X_m : s \in \Sigma_1\}$ and $\{sur_2(s)^{\mathfrak{W}_2, \mathcal{B}} \cap Y_r : s \in \Sigma_1\}$ both form $\kappa$-partitions of $X_m = Y_r$. Some abbreviations are useful: set

- $A_i = X_i - X_{i+1}$, for $0 \le i < m$,

- $B_i = Y_i - Y_{i+1}$, for $0 \le i < r$,

- $A_{-1} = W_1 - X_0$, $B_{-1} = W_2 - Y_0$.

So far we have merged the $X_m$-part of $\mathfrak{W}_1$ with the $Y_r$-part of $\mathfrak{W}_2$. It remains to take care of the sets $A_i$, $-1 \le i < m$, and $B_i$, $-1 \le i < r$: the sets $A_i$ will be merged with new models $\mathfrak{W}^i \in \mathcal{M}_2$ and the sets $B_i$ will be merged with new models $\mathfrak{V}^i$ from $\mathcal{M}_1$. Thus, the final model will be obtained by merging the disjoint union of $\mathfrak{W}_1$ and $\mathfrak{W}^i$, $-1 \le i < m$ with the disjoint union of $\mathfrak{W}_2$ and $\mathfrak{V}^i$, $-1 \le i < r$. Figure 5 illustrates this merging. In the figure, we assume that $\Sigma_1 = \{s_1, \ldots, s_k\}$.

Of course, when merging $A_i$, $i \ge 0$, with a new model $\mathfrak{W}^i$ we have to respect the partition

$$\{sur_1(t)^{\mathfrak{W}_1, \mathcal{A}} \cap A_i \mid t \in \Sigma_1\}$$

of $A_i$. And when merging $B_i$, $i \ge 0$, with a new model $\mathfrak{V}^i$ we have to respect the partition

$$\{sur_1(t)^{\mathfrak{W}_1, \mathcal{B}} \cap B_i \mid t \in \Sigma_1\}$$

of $B_i$. Note that for $A_{-1}$ and $B_{-1}$ there is no partition to take care of. We now proceed with the formal construction. We find models $\mathfrak{W}^i = \left\langle A_i, \mathcal{G}^{\mathfrak{W}^i}, \mathcal{Q}^{\mathfrak{W}^i} \right\rangle \in \mathcal{M}_2$ with assignments $\mathcal{B}^i = \langle \mathcal{B}_1^i, \mathcal{B}_2^i \rangle$, $-1 \le i \le m - 1$, such that, for $0 \le i \le m - 1$,

$$sur_2(s)^{\mathfrak{W}^i, \mathcal{B}^i} = sur_1(s)^{\mathfrak{W}_1, \mathcal{A}} \cap A_i \text{ for all } s \in \Sigma_1. \tag{7}$$

This follows from [a5], [a6], and Lemma 47 (2).

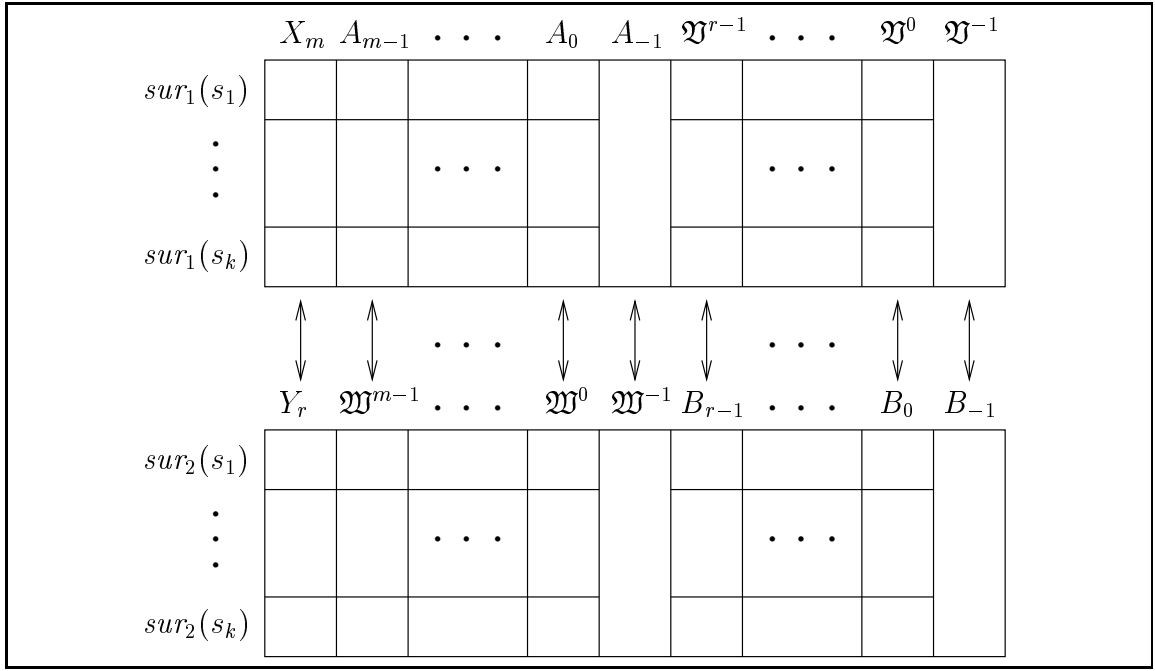

Figure 5: The bijection.

We find, now using [b5], [b6], and Lemma 47 (1), models $\mathfrak{V}^i = \left\langle B_i, \mathcal{F}^{\mathfrak{V}^i}, \mathcal{R}^{\mathfrak{V}^i} \right\rangle \in \mathcal{M}_1$ with assignments $\mathcal{A}^i = \left\langle \mathcal{A}_1^i, \mathcal{A}_2^i \right\rangle$, $-1 \leq i \leq r - 1$, such that, for $0 \leq i \leq r - 1$,

$$sur_1(s)^{\mathfrak{V}^i, \mathcal{A}^i} = sur_2(s)^{\mathfrak{W}_2, \mathcal{B}} \cap B_i \text{ for all } s \in \Sigma_1. \tag{8}$$

Let

$$\mathfrak{W}_1' = \left\langle W_1 \cup (W_2 - Y_r), \mathcal{F}^{\mathfrak{W}_1'}, \mathcal{R}^{\mathfrak{W}_1'} \right\rangle \in \mathcal{M}_1$$

be the disjoint union of the $\mathfrak{V}^i$, $-1 \leq i < r$, and $\mathfrak{W}_1$, and let

$$\mathfrak{W}_2' = \left\langle W_2 \cup (W_1 - X_m), \mathcal{G}^{\mathfrak{W}_2'}, \mathcal{Q}^{\mathfrak{W}_2'} \right\rangle \in \mathcal{M}_2$$

be the disjoint union of the $\mathfrak{W}^i$, $-1 \leq i < m$, and $\mathfrak{W}_2$. We assume $X_m = Y_r$ and so the domain of both ADMs is $W_1 \cup W_2$.

Define a model $\mathfrak{W} = \left\langle W, (\mathcal{F} \cup \mathcal{G})^{\mathfrak{W}}, (\mathcal{R} \cup \mathcal{Q})^{\mathfrak{W}} \right\rangle \in \mathcal{M}$ based on $W = W_1 \cup W_2$ by putting

- $\mathcal{R}^{\mathfrak{W}} = \mathcal{R}^{\mathfrak{W}_1'}$,

- $\mathcal{F}^{\mathfrak{W}} = \mathcal{F}^{\mathfrak{W}_1'}$,

- $\mathcal{Q}^{\mathfrak{W}} = \mathcal{Q}^{\mathfrak{W}_2'}$,

- $\mathcal{G}^{\mathfrak{W}} = \mathcal{G}^{\mathfrak{W}_2'}$.

Define an assignment $\mathcal{C} = \langle \mathcal{C}_1, \mathcal{C}_2 \rangle$ in $\mathfrak{W}$ by putting

- $\mathcal{C}_2(a) = \mathcal{A}_2(a) (= \mathcal{B}_2(a))$, for all $a \in obj(\Gamma_1)$.

- $\mathcal{C}_1(x) = \mathcal{A}_1(x) \cup \bigcup_{-1 \leq i < r} \mathcal{A}_1^i(x)$, for all set variables $x$ in $term(\Gamma)$.

  Notice that $\mathcal{C}_1(x) = \mathcal{B}_1(x) \cup \bigcup_{-1 \leq i < m} \mathcal{B}_1^i(x)$, for all set variables $x \in term(\Gamma)$.

- $\mathcal{C}_1(x_t) = \mathcal{A}_1(x_t) \cup \bigcup_{-1 \leq i < r} \mathcal{A}_1^i(x_t)$, for all $t = g(t_1, \ldots, t_k) \in sub^1(\Gamma)$.

- $\mathcal{C}_1(x_t) = \mathcal{B}_1(x_t) \cup \bigcup_{-1 \leq i < m} \mathcal{B}_1^i(x_t)$, for all $t = f(t_1, \ldots, t_k) \in sub^1(\Gamma)$.

We will show that $\langle \mathfrak{W}, \mathcal{C} \rangle \models \Gamma$. Firstly, however, we make a list of the relevant properties of $\langle \mathfrak{W}, \mathcal{C} \rangle$:

**Claim 2.**

[c1] $\mathcal{C}_2(a) \in X_m = Y_r$, for all $a \in obj(\Gamma)$;

[c2] $\langle \mathfrak{W}, \mathcal{C} \rangle \models \Gamma_1 \cup \Gamma_2$;

[c3] $sur_1(t)^{\mathfrak{W}, \mathcal{C}} \cap (X_0 \cup Y_0) = sur_2(t)^{\mathfrak{W}, \mathcal{C}} \cap (X_0 \cup Y_0)$, for all $t \in \Sigma_1$;

[c4] $sur_1(s)^{\mathfrak{W}, \mathcal{C}} \cap (X_0 \cup Y_0) = sur_2(s)^{\mathfrak{W}, \mathcal{C}} \cap (X_0 \cup Y_0)$, for all $s \in sub^1(\Gamma)$;

[c5] $X_{n+1} \subseteq X_n \cap c^{\mathfrak{W}}(X_n)$, for all $0 \leq n < m$;

[c6] $Y_{n+1} \subseteq Y_n \cap d^{\mathfrak{W}}(Y_n)$, for all $0 \leq n < r$;

[c7] for all $g \in \mathcal{G}$ of arity $l$, $0 \leq n < m$, and all $C_1, \ldots, C_l \subseteq W$:

$$g^{\mathfrak{W}}(C_1, \ldots, C_l) \cap X_n = g^{\mathfrak{W}}(C_1 \cap X_n, \ldots, C_l \cap X_n) \cap X_n;$$

[c8] for all $f \in \mathcal{F}$ of arity $l$, $0 \leq n < r$, and all $C_1, \ldots, C_l \subseteq W$:

$$f^{\mathfrak{W}}(C_1, \ldots, C_l) \cap Y_n = f^{\mathfrak{W}}(C_1 \cap Y_n, \ldots, C_l \cap Y_n) \cap Y_n.$$

*Proof of Claim 2.* [c1] follows from [a1] and [b1] and the construction of $\langle \mathfrak{W}, \mathcal{C} \rangle$. [c2] follows from [a2] and [b2]. [c3] follows from the construction of $\langle \mathfrak{W}, \mathcal{C} \rangle$ and equations (6), (7), and (8). [c4] follows from [c3]. [c5] and [c6] follow from [a3] and [b3], respectively. It remains to prove [c7] and [c8]. But [c7] follows from the fact that $\langle \mathfrak{W}, G^{\mathfrak{W}} \rangle$ is the disjoint union of structures based on $X_n$ and $W - X_n$, for $0 \leq n < m$, and [c8] is dual to [c7]. Claim 2 is proved.

We now show $\langle \mathfrak{W}, \mathcal{C} \rangle \models \Gamma$. To this end we first show the following:

**Claim 3.** For all $k_1, k_2$ with $0 \leq k_1 \leq m$ and $0 \leq k_2 \leq r$ and all $s \in sub^1(\Gamma)$ with $d_F(s) \leq k_1$ and $d_G(s) \leq k_2$ we have, for $Z \in \{X_{k_1}, Y_{k_2}\}$,

$$Z \cap s^{\mathfrak{M}, \mathcal{C}} = Z \cap sur_1(s)^{\mathfrak{M}, \mathcal{C}} = Z \cap sur_2(s)^{\mathfrak{M}, \mathcal{C}}.$$

*Proof of Claim 3.* By [c4] it suffices to prove the first equation. The proof is by induction on the cardinal $k_1 + k_2$. The induction base $k_1 = k_2 = 0$ follows from $sur_1(s) = sur_2(s)$ for $d_F(s) = d_G(s) = 0$.

Suppose the claim is proved for all $X_k, Y_{k'}$ with $k \leq m$, $k' \leq r$ and $k + k' < k_1 + k_2$. We prove the claim for $X_{k_1}, Y_{k_2}$. The proof is by induction on the construction of terms $s$ with $d_F(s) \leq k_1$ and $d_G(s) \leq k_2$. The boolean cases are trivial.

Suppose $s = f(s_1, \ldots, s_l)$ with $d_F(s) \leq k_1$ and $d_G(s) \leq k_2$. We have to show the following two statements:

(i) $X_{k_1} \cap s^{\mathfrak{W}, \mathcal{C}} = X_{k_1} \cap sur_1(s)^{\mathfrak{W}, \mathcal{C}}$.

(ii) $Y_{k_2} \cap s^{\mathfrak{W}, \mathcal{C}} = Y_{k_2} \cap sur_1(s)^{\mathfrak{M}, \mathcal{C}}$.

Consider (i) first. The induction hypothesis yields

$$X_{k_1-1} \cap s_i^{\mathfrak{W}, \mathcal{C}} = X_{k_1-1} \cap sur_1(s_i)^{\mathfrak{W}, \mathcal{C}}$$

for $1 \leq i \leq l$. We have

$$
\begin{aligned}
X_{k_1-1} \cap c^{\mathfrak{W}}(X_{k_1-1}) \cap s^{\mathfrak{W}, \mathcal{C}} &= X_{k_1-1} \cap c^{\mathfrak{W}}(X_{k_1-1}) \cap f^{\mathfrak{W}}(s_1^{\mathfrak{W}, \mathcal{C}}, \ldots, s_l^{\mathfrak{W}, \mathcal{C}}) \\
&= X_{k_1-1} \cap c^{\mathfrak{W}}(X_{k_1-1}) \cap f^{\mathfrak{W}}(sur_1(s_1)^{\mathfrak{W}, \mathcal{C}}, \ldots, sur_1(s_l)^{\mathfrak{W}, \mathcal{C}}) \\
&= X_{k_1-1} \cap c^{\mathfrak{W}}(X_{k_1-1}) \cap sur_1(s)^{\mathfrak{W}, \mathcal{C}}.
\end{aligned}
$$

The second equation is an immediate consequence of the third property of covering normal terms as given in Definition 26. Now the equation follows from [c5], i.e. $X_{k_1} \subseteq X_{k_1-1} \cap c^{\mathfrak{W}}(X_{k_1-1})$. (i) is proved.

(ii) Suppose first that $k_2 = r$. Then $Y_{k_2} = X_m$ and the claim can be proved as above since $X_m \subseteq X_{k_1}$ and, by induction hypothesis, $X_{k_1-1} \cap s_i^{\mathfrak{W}, \mathcal{C}} = X_{k_1-1} \cap sur_1(s_i)^{\mathfrak{W}, \mathcal{C}}$, for $1 \leq i \leq l$.

Assume now that $k_2 < r$. By induction hypothesis,

$$Y_{k_2} \cap s_i^{\mathfrak{W}, \mathcal{C}} = Y_{k_2} \cap sur_2(s_i)^{\mathfrak{W}, \mathcal{C}},$$

for $1 \leq i \leq l$. Hence

$$f^{\mathfrak{W}}(Y_{k_2} \cap s_1^{\mathfrak{W}, \mathcal{C}}, \ldots, Y_{k_2} \cap s_l^{\mathfrak{W}, \mathcal{C}}) = f^{\mathfrak{W}}(Y_{k_2} \cap sur_2(s_1)^{\mathfrak{W}, \mathcal{C}}, \ldots, Y_{k_2} \cap sur_2(s_l)^{\mathfrak{W}, \mathcal{C}}).$$

We intersect both sides of the equation with $Y_{k_2}$ and derive with the help of [c8]:

$$Y_{k_2} \cap f^{\mathfrak{W}}(s_1^{\mathfrak{W}, \mathcal{C}}, \ldots, s_l^{\mathfrak{W}, \mathcal{C}}) = Y_{k_2} \cap f^{\mathfrak{W}}(sur_2(s_1)^{\mathfrak{W}, \mathcal{C}}, \ldots, sur_2(s_l)^{\mathfrak{W}, \mathcal{C}}).$$

This means $Y_{k_2} \cap s^{\mathfrak{W}, \mathcal{C}} = Y_{k_2} \cap sur_2(s)^{\mathfrak{W}, \mathcal{C}}$, and the equation follows. The statements are proved.

The case $s = g(s_1, \ldots, s_l)$ is dual and left to the reader. We have proved claim 3.

By induction (c.f. in the proof of Theorem 20 the proof of (1) from the corresponding claim), we obtain from Claim 3:

$$X_m \cap s^{\mathfrak{W}, \mathcal{C}} = X_m \cap sur_1(s)^{\mathfrak{M}, \mathcal{C}} \text{ for all } s \in term(\Gamma). \tag{9}$$

Let us see how $\langle \mathfrak{W}, \mathcal{A} \rangle \models \Gamma$ follows from (9). We distinguish three cases: Suppose $R(a, b) \in \Gamma$. Then $R(a, b) \in \Gamma_1$ and therefore $\langle \mathfrak{W}, \mathcal{C} \rangle \models R(a, b)$. Similarly, $Q(a, b) \in \Gamma$ implies $Q(a, b) \in \Gamma_2$ and $\langle \mathfrak{W}, \mathcal{C} \rangle \models Q(a, b)$. Suppose $(a : t) \in \Gamma$. Then $(a : sur_1(t)) \in \Gamma_1$ and so, by [c2], $\mathcal{C}_2(a) \in sur_1(t)^{\mathfrak{W}, \mathcal{C}}$ which implies, by (9), $\mathcal{C}_2(a) \in t^{\mathfrak{W}, \mathcal{C}}$. Hence $\langle \mathfrak{W}, \mathcal{C} \rangle \models (a : t)$. This finishes the proof of Theorem 30.